\documentclass{article}

\usepackage{subcaption}
\usepackage{booktabs} % for professional tables
\usepackage{multirow}
\usepackage{hyperref}
\usepackage{xcolor}
\usepackage{authblk}
\usepackage{subcaption}
\usepackage{amsmath,amssymb,bm}
\usepackage{mathtools}
\usepackage{amsthm}
\usepackage[numbers]{natbib} 

\def\gX{{\mathcal{X}}}
\def\gT{{\mathcal{T}}}
\def\gO{{\mathcal{O}}}
\def\sA{{\mathbb{A}}}
\def\sC{{\mathbb{C}}}

\DeclareMathOperator{\cost}{\mathsf{cost}}
\DeclareMathOperator{\rrc}{\mathsf{rc}}
\DeclareMathOperator{\size}{\mathsf{size}}
\DeclareMathOperator{\eff}{\mathsf{eff}}
\DeclareMathOperator{\avcost}{\mathsf{avc}}
\DeclareMathOperator*{\argmin}{arg min}

\DeclareMathOperator{\Xa}{\gX_\mathsf{{aff}}}

\newtheorem{theorem}{Theorem}
\newtheorem{problem}[theorem]{Problem}

\usepackage[capitalize,noabbrev]{cleveref}
\usepackage{graphicx}

\usepackage{algorithm,algorithmic}

\begin{document}

\title{GLANCE: Global Actions in a Nutshell for Counterfactual Explainability}

% \begin{authorlist}
\author[1,$\dagger, *$]{Loukas Kavouras}
\author[2,1,$\dagger,*$]{Eleni Psaroudaki}
\author[1]{Konstantinos Tsopelas}
\author[3]{Dimitrios Rontogiannis}
\author[1]{Nikolaos Theologitis}
\author[4,5]{Dimitris Sacharidis}
\author[1]{Giorgos Giannopoulos}
\author[7]{Dimitrios Tomaras}
\author[6]{Kleopatra Markou}
\author[6]{Dimitrios Gunopulos}
\author[2,8]{Dimitris Fotakis}
\author[1,6]{Ioannis Emiris}

% \end{authorlist}

\affil[1]{Institute for the Management of Information Systems, Athena Research Center, Greece}
\affil[2]{Department of Electrical and Computer Engineering, National Technical University of Athens, Greece}
\affil[3]{Max Planck Institute for Software Systems, Kaiserslautern, Germany}
\affil[4]{Universit{é} Libre de Bruxelles, Belgium}
\affil[5]{FARI Institute, Belgium}
\affil[6]{Department of Informatics and Telecommunications, National and Kapodistrian University of Athens, Greece}
\affil[7]{Department of Informatics, Athens University of Economics and Business, Greece}
\affil[8]{Archimedes, Athena Research Center, Greece}
\affil[$\dagger$]{Corresponding authors: Loukas Kavouras - \href{mailto:kavouras@athenarc.gr}{kavouras@athenarc.gr} , Eleni Psaroudaki - \href{mailtp:epsaroudaki@mail.ntua.gr}{epsaroudaki@mail.ntua.gr} }
\affil[$*$]{Equal Contribution}
% \correspondingauthor{Firstname2 Lastname2}{first2.last2@www.uk}
% \thanks{This work has been funded by the European Union’s Horizon Europe research and innovation programme under Grant Agreement No. 101070568 (AutoFair).}

\maketitle

\begin{abstract}
The widespread deployment of machine learning systems in critical real-world decision-making applications has highlighted the urgent need for counterfactual explainability methods that operate effectively. Global counterfactual explanations, expressed as actions to offer recourse, aim to provide succinct explanations and insights applicable to large population subgroups. High effectiveness, measured by the fraction of the population that is provided recourse, ensures that the actions benefit as many individuals as possible. Keeping the cost of actions low ensures the proposed recourse actions remain practical and actionable. Limiting the number of actions that provide global counterfactuals is essential to maximizing interpretability. The primary challenge, therefore, is to balance these trade-offs--maximizing effectiveness, minimizing cost, while maintaining a small number of actions. We introduce \texttt{GLANCE}, a versatile and adaptive algorithm that employs a novel agglomerative approach, jointly considering both the feature space and the space of counterfactual actions, thereby accounting for the distribution of points in a way that aligns with the model's structure.  This design enables the careful balancing of the trade-offs among the three key objectives, with the size objective functioning as a tunable parameter to keep the actions few and easy to interpret. Our extensive experimental evaluation demonstrates that  \texttt{GLANCE} consistently shows greater robustness and performance compared to existing methods across various datasets and models.
\end{abstract}

% \input{intro}
% \input{related}
% \input{problem}
% \input{clustering}
% \input{explainable}
% \input{experimental}
% \input{conclusion}

%     % \link{Datasets}{https://aaai.org/example/datasets}

\section{Introduction}
% \frenchspacing
Machine learning models are increasingly deployed in critical domains such as loan approvals, hiring, and healthcare. This widespread adoption intensifies the need for transparency and interpretability in model decisions,
requiring users to understand how their input features influence the outcomes and how they might change them to achieve 
favorable outcomes, known as recourse \citep{miller2019explanation}. Counterfactual explanations have gathered extensive attention for their suitability for achieving algorithmic recourse \citep{karimi2020survey}, their interpretability \citep{wachter2017counterfactual}, actionability \citep{ustun2019actionable}, utility in fairness audits \citep{certifai,kavouras2023fairness}, etc. A counterfactual action, or simply an \emph{action}, defines the specific feature changes that convert an unfavorable decision into a favorable one.

Traditionally, counterfactual explanations refer to \emph{local explainability} tied to a specific negatively affected instance. However, many real-world scenarios require \emph{global counterfactual explainability}, which provides shared, population-level explanations.
% that apply across the affected population. 
While collecting 
% a collection of 
all local counterfactuals could technically cover all affected individuals,  this approach undermines interpretability, a core requirement of 
% which is central to 
global explainability. 

Building on prior research \citep{rawal2020beyond,kanamori2022counterfactual}, we define Global Counterfactual Explanations (GCEs) as a small set of \emph{global actions} designed to provide effective recourse for the affected population.
Any global counterfactual solution must meet three objectives:
(1) be composed of a small number of actions to ensure interpretability (small size), 
(2) minimize the cost of implementing those actions (low cost), and 
(3) offer recourse to as many affected individuals as possible (high effectiveness). 
\begin{figure}[t]
\centering
\begin{subfigure}[t]{0.54\columnwidth}
\centering
\includegraphics[height=4.5cm]{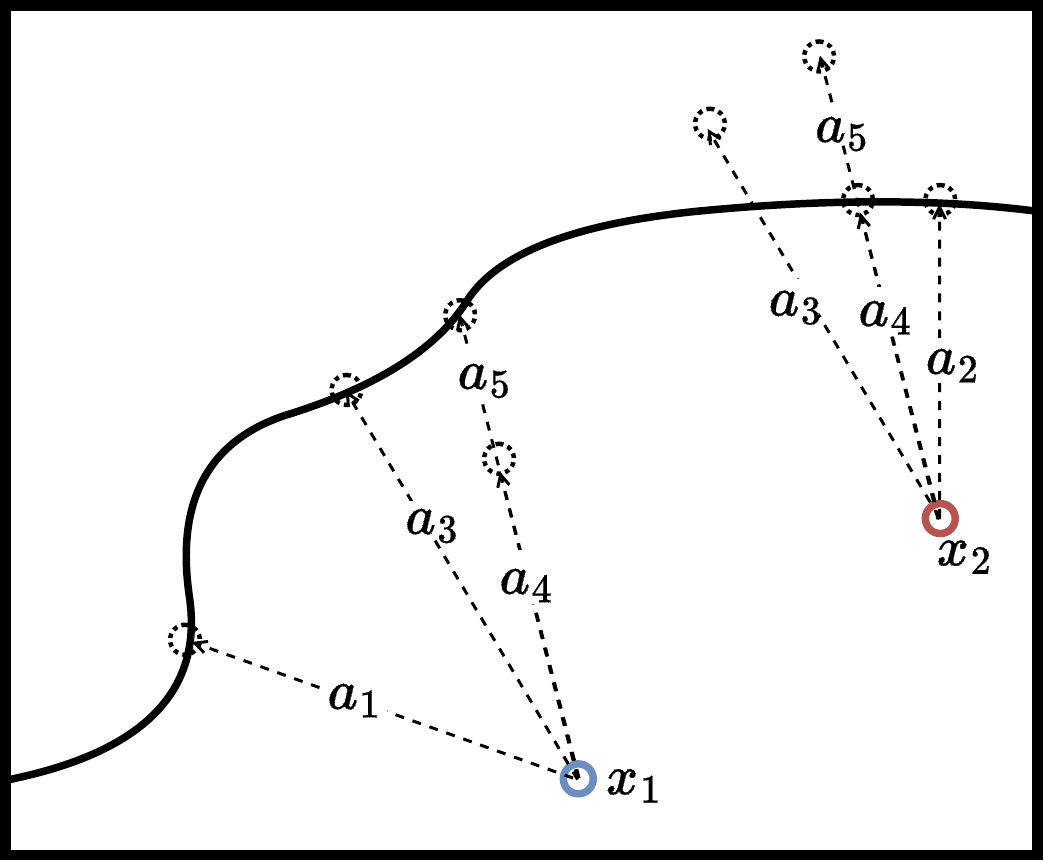}
\caption{Feature space}
\label{fig:feat_space}
\end{subfigure}%
\begin{subfigure}[t]{0.44\columnwidth}
\centering
\includegraphics[height=4.5cm]{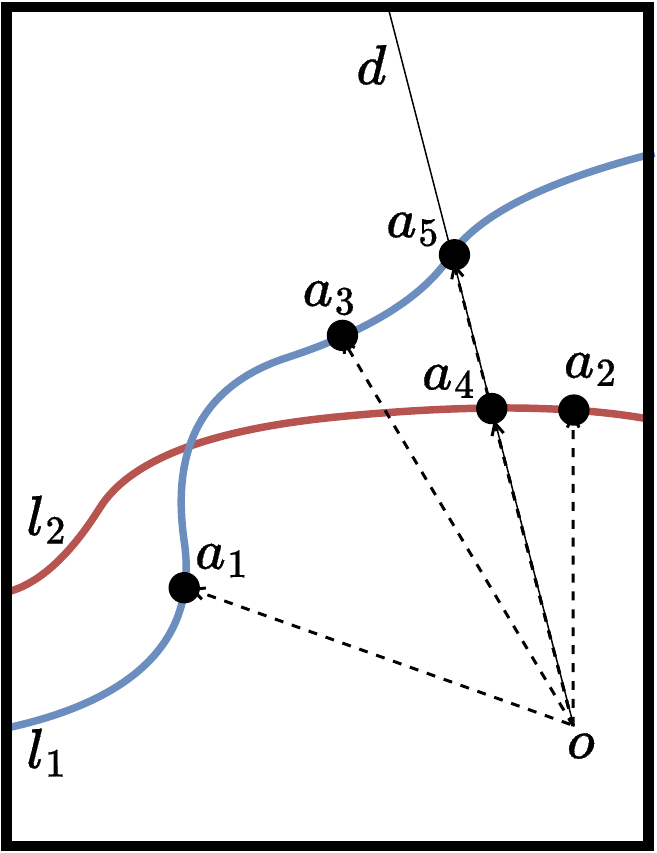}
\caption{Action space}
\label{fig:act_space}
\end{subfigure}
\caption{A toy example depicting two negative instances $x_1, x_2$, and five actions. 
(a) The feature space; the line is the decision boundary. (b) The action space; $l_1, l_2$ depict the decision boundary from the perspective of $x_1, x_2$, respectively.}
\label{fig:example}
\end{figure}

As noted by \citet{branke2008multiobjective}, the relationships between multiple optimization objectives are often complex, and aggregating them into a single objective, even though common in practice \citep{rawal2020beyond}, can be problematic since they are typically non-commensurable.
Framing GCEs as multi-objective optimization allows us to explore the inherent \emph{trade-offs} between effectiveness and cost, especially when the solution size is constrained.

To understand this optimization problem, assume a 2-d numerical feature space, and consider the two affected instances $x_1, x_2$ depicted in Fig.~\ref{fig:feat_space}. Assume that the recourse cost equals the distance to the decision boundary, drawn as a line in the figure. Observe that $a_1$ (resp.\ $a_2$) is the local action providing recourse for $x_1$ (resp.\ $x_2$) at minimum cost. 

Further, consider the \emph{action space} depicted in Fig.~\ref{fig:act_space}, where every action is represented as a point (or equivalently a vector relative to the center $o$ of the coordinate system). The blue $l_1$ and red $l_2$ lines represent the decision boundary seen from the perspectives of $x_1$ and $x_2$, respectively. The blue line $l_1$ separates the actions that provide recourse for $x_1$ (any action on the outside, away from $o$) from those that do not. Action $a_1$ lies on $l_1$, and is the closest point to $o$, and thus the min-cost local action for $x_1$.
Similarly, the red line $l_2$ concerns $x_2$ and contains its min-cost action $a_2$.

Consider now the problem of finding a {\it single action} GCE. To provide recourse for both $x_1$ and $x_2$, we look for an action that lies outside both lines in Fig.~\ref{fig:act_space}. Among all such actions, $a_3$ has the minimum cost and thus is the optimal global action that maximizes effectiveness.  
If we trade off effectiveness for cost, $a_2$ is the optimal global action that minimizes cost, but has 50\% effectiveness (brings recourse to $x_2$ but not to $x_1$).

Finding GCEs gives rise to a different optimization problem than its local counterpart, and requires a nuanced exploration of the action space.
Even if optimal local actions are generated for each instance and a small subset is chosen as the global actions, this may still result in a suboptimal set of GCEs.
In our example, the optimal global action $a_3$ is not an optimal local action for either $x_1$ or $x_2$; in fact, $a_3$ can be viewed as a compromise between $a_1$ and $a_2$, the locally optimal actions. 

\paragraph{Key Contributions}
First, we formally introduce Global Counterfactual Explanations (GCEs) as small, interpretable sets of actions that provide recourse to large population subgroups. A user study confirms that smaller action sets yield more intuitive and practical explanations.

Second, we formulate the problem of finding GCEs as a multi-objective optimization task, balancing effectiveness and cost under a size constraint. We compare our formulation to prior approaches, such as that of \citet{pmlr-v202-ley23a}, and highlight its interpretability advantages. We prove that a restricted version of the problem is NP-hard, motivating the need for efficient algorithms.

Third, we propose \texttt{GLANCE}, a novel algorithm that clusters individuals in both feature and action space to generate diverse and coherent counterfactuals. Our results show that proper use of clustering can lead to higher quality GCEs, contradicting the claims of \citet{kanamori2022counterfactual}. 

Finally, extensive experiments across models and datasets show that \texttt{GLANCE} \textit{Pareto dominates} baselines, i.e., achieves higher effectiveness and lower cost, in 57\% of cases (and is dominated in only 1\%). A user study further supports its practical appeal even when it is not formally dominant.

\section{Related Work}

\begin{figure}[t!]
    \centering
    \begin{subfigure}[t]{0.34\columnwidth}
    \centering
    \includegraphics[height=4.5cm]{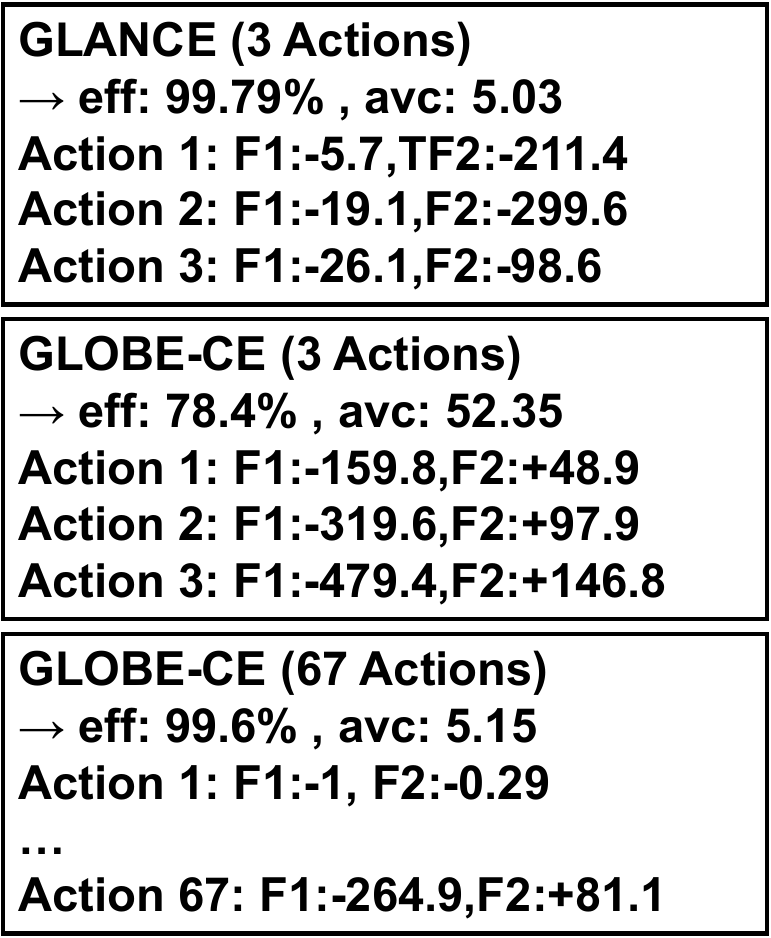}
        \caption{Actions}
    \end{subfigure}
     \begin{subfigure}[t]{0.64\columnwidth}
        \centering
        \includegraphics[height=4.5cm]{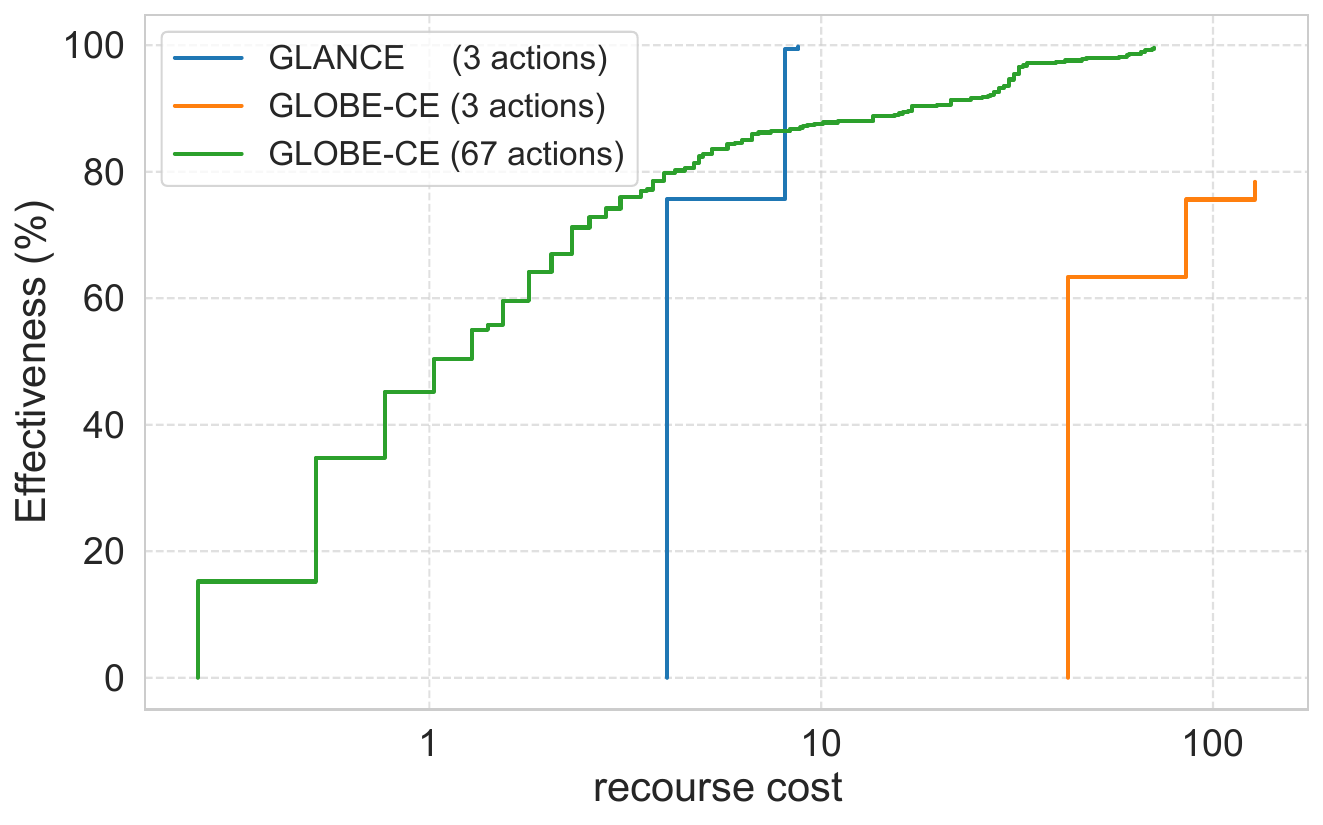}
        \caption{Effectiveness vs Recourse cost}
    \end{subfigure}
 \caption{
    Comparison of GCEs on the COMPAS dataset using an \texttt{XGBoost} model.
    (a) GCEs generated by \texttt{GLANCE} (s=3) and \texttt{GLOBE-CE} (s=3 \& default s=67).  
    (b) Effectiveness–recourse cost curves showing the share of individuals (y-axis) achieving recourse below a given cost (x-axis); the area under the curve reflects the average recourse cost. Costs follow Sec.~5.}
    % \caption{ Comparison of GCEs on the COMPAS dataset using an \texttt{XGBoost} model. (a) GCEs produced from \texttt{GLANCE} ($s=3$), by \texttt{GLOBE-CE} ($s=3$), and by \texttt{GLOBE-CE} (default, 67 actions). (b) Effectiveness–recourse cost curve: the percentage of individuals (y-axis) achieving recourse with actions below a given cost (x-axis); average recourse cost for the population is proportional to the area under the curveCosts computed as described in Section~5.}
\label{fig:actions_example}
\end{figure}

\paragraph{Counterfactual Explanations}
There has been a plethora of work focusing on counterfactual explanations \citep[see e.g.,][and references therein]{guidotti2024counterfactual,verma2024counterfactual}. An overview of methods of algorithmic recourse, which provides explanations and recommendations to individuals impacted by automated decision-making systems, is presented by \citet{karimi2020survey}. These methods can be either model-agnostic \citep[e.g.,][]{wachter2017counterfactual} or model-specific \citep[e.g., for trees as in][]{carreira2021counterfactual}, may focus on specific properties such as diversity \citep{mothilal2020explaining}, feasibility \citep{ustun2019actionable} or robustness \citep{stepka2025robust_counterfactuals}, and may follow different methodological paradigms such as optimization-based approaches \citep{dandl2020multi,karimi2021algorithmic} or instance-based approaches \cite{delaney2021instance,brughmans2024nice,smyth2020good}, target local or global explanations.
While local counterfactuals are well defined, global ones present challenges as they must provide recourse for all individuals within a specific group in the same manner while maintaining explainability and staying true to the notion of local counterfactuals, which focuses on minimal changes. Achieving this balance globally presents notable difficulties.

\paragraph{Global Counterfactuals}
\citet{rawal2020beyond} introduced \texttt{AReS}, a framework for global counterfactual explanations that jointly optimizes recourse correctness, coverage, and cost, providing an interpretable summary of recourses, expressed in a two-level rule set. 
However, the \texttt{AReS} framework may fail to cover the entire population.  \citet{ley2022global} later improved the computational efficiency with \texttt{Fast AReS}.
In another direction, \citet{kanamori2022counterfactual} introduced \texttt{CET}, which partitions the space and assigns an action to each part
% in a transparent, consistent way, though its computational complexity limits scalability.
transparently and consistently. Although effective, its computational complexity limits scalability.
\citet{warren2023explaining}  developed \texttt{Group-CF}, which generates counterfactuals that seek to maximize effectiveness, though it can result in higher costs.

\paragraph{Actions vs Directions}
\citet{pmlr-v202-ley23a} proposed \texttt{GLOBE-CE}, where global counterfactual explainability is defined differently, as a small set of action \emph{directions} along which individuals can ``move'' to achieve recourse.
\texttt{GLOBE-CE} attacks a different optimization problem 
% than the one we study here. 
from ours.
For the toy example in Fig.~\ref{fig:example}, the single direction that minimizes the total recourse cost for $x_1$ and $x_2$ is the direction $d$ depicted in Fig.~\ref{fig:act_space}. This direction contains actions $a_4$ and $a_5$ that bring recourse to $x_2$ and $x_1$, respectively, with minimum cost \emph{along} $d$. 
% It is important to note that 
However, neither $a_4$ nor $a_5$ is optimal as a GCE (or for local explainability). Even the set $\{a_4, a_5\}$ is a suboptimal GCE--the set $\{a_1, a_2\}$ dominates it with equal size and effectiveness but lower total cost. In general, (1) 
% directly 
translating \texttt{GLOBE-CE} outputs into GCE leads to numerous micro-actions (two in our example, but potentially as many as the individuals' number), which reduces interpretability, (2) choosing a few actions along the optimal directions may result in suboptimal GCEs (any single action along $d$ is dominated by $a$ in Fig.~\ref{fig:example}), and (3) since directions lack clear endpoints, they fail to specify the required magnitude of change, creating uncertainty for individuals seeking recourse and limiting real-world applicability. 

Therefore, \texttt{GLOBE-CE} relaxes the CGE formulation, 
% allowing them to ignore the size.
ignoring size.
Fig.~\ref{fig:actions_example} compares the output of \texttt{GLANCE} to that of  \texttt{GLOBE-CE}. By default, \texttt{GLOBE-CE} (green line) produces 67 micro-actions. With such a wide range of actions available, most individuals can achieve recourse through low-cost actions, resulting in high effectiveness and low average cost--
at the expense of interpretability.
% . However, this comes at the cost of interpretability. 
When restricted to 
% output only 
three actions (orange line), \texttt{GLOBE-CE} experiences a sharp decline in effectiveness and a substantial increase in average cost. 
In contrast, \texttt{GLANCE} (blue line), thanks to its effective exploration of the action space, produces high-quality GCEs (with high effectiveness and low average cost) without sacrificing interpretability (just three actions).

\paragraph{Other} Other works include the ones of \citet{carrizosa2024generating,carrizosa2024mathematical}, who use mixed-integer quadratic models for group-level explanations, and \citet{koo2020inverse}, who employ Lagrangian methods. Research has also extended to generating global counterfactuals for graphs \citep{huang2023global} and auditing subgroup fairness \citep{kavouras2023fairness,fragkathoulas2025facegroup}.

\section{Problem Formulation and Hardness }\label{sec:problem}

We consider a black box binary classifier $h:\gX \rightarrow \{-1,1\}$, where the positive outcome is favorable and the negative outcome is unfavorable. We will focus on the set $\Xa \subseteq \gX$ of adversely affected individuals, i.e., those who receive the unfavorable outcome.
We denote as $\sA$ the set of all possible actions (which is potentially infinite), where an action $a \in \sA$ is a set of changes to feature values, e.g., 
% $a = \{\text{country} \to \text{US},\text{education-num} \to +2\}$
$a$ $=$ $\{$ country $\to$ US, education-num $\to$ $+2\}$, which, when applied to an instance $x \in \Xa$, results in a counterfactual instance $x' = a(x)$. Every action $a$ has a cost, denoted as $\cost(a,x)$, and is effective for an instance $x$ if $h(a(x))$ $=$ $h(x')$ $=$ $1$. Let $\sC \subseteq \sA$. The recourse cost $\rrc(\sC,x)$ of an instance $x$ is the minimum cost incurred from an effective action in $\sC$:
\begin{equation}
\label{eq:rc}
\rrc(\sC,x)=\min \{ \cost(a,x) | a \in \sC : h(a(x)) = 1 \}
    \end{equation}

Let $X_{\sC}=\{x \in \Xa | h(a(x)) = 1, a \in \sC\}$ be the set of instances that flip their prediction using one of the actions in $\sC$. Then the effectiveness, also known as coverage \citep{ley2022global},
% footnote{Another way to formalize effectiveness using \citet{rawal2020beyond} notation is as the  $1 - \frac{\text{incorrectrecourse}}{\text{coverage}}$
% . They aim to jointly minimize the number of instances in $\Xa$ that do not achieve the desired class while maximizing the number $\Xa$. We formally use the effectiveness as the combination of these two metrics. 
% \inote{tofix: or coverage\citep{pmlr-v202-ley23a}. However, coverage terminology was used with multiple definitions\citep{rawal2020beyond,pmlr-v202-ley23a,huang2023global} and thus decided not to use it. }
% } 
of $\sC$ for the affected instances $\Xa$ is defined as the percentage of $\Xa$ that managed to flip their prediction using one of the actions in $\sC$:
\[
\eff(\sC,\Xa) = \frac{|X_{\sC}|}{|\Xa|},
\]
The cost of $\sC$ in $\Xa$ is defined as the average recourse cost of the instances in $\Xa$:
\[
\avcost(\sC,\Xa) = \frac{\sum\limits_{x \in X_{\sC}}\rrc(\sC,x)}{|X_\sC|} .
\]
Finally, let $\size(\sC) = |\sC|$ denote the cardinality of a set $\sC$.

As it is clear, multiple sets of actions can produce recourse for $\Xa$, and the quality of such a set is a factor of the three notions we introduced: effectiveness, cost, and size. An ideal global counterfactual should maximize effectiveness while minimizing the cost and the size, based on the properties that state-of-the-art works \citep{rawal2020beyond,kanamori2022counterfactual,pmlr-v202-ley23a,huang2023global} have argued are essential. 
The requirement for a small set of actions is to enhance the interpretability of the explanation. This can also be expressed as a constraint on the set size we can afford \cite{rawal2020beyond}. Therefore, instead of minimizing the size, we constrain it to 
$\size(\sC) \leq s$,  
% : \[\size(\sC) = s,\]  
where $s$ is a small positive integer (set to four for our empirical evaluation). For the rest of the paper, we will use the following problem formulation:

% Formally, we define the global counterfactual explanation problem as follows: {\bf DG: I THINK PROBLEM SHOULD BE REMOVED}

% \begin{problem}[Global Counterfactual Explanations (GCE)] \label{prob:gce}
%     Given a black box model $h$ that classifies the $\Xa$ instances to the negative class, our goal is to find the set $\sC \subseteq \sA, \sC \neq \emptyset$
%     % set of actions $A, A \neq \emptyset$ 
%     that represents a solution to the following multi-objective optimization:
%     % problem which is defined as the minimization of three objective functions:\
%     % \vspace{-.5em} 
%     \[\text{minimize} \quad \Big( \size(\sC), \; -\eff(\sC, \Xa), \;  \avcost(\sC, \Xa)\Big)\] 
%     % \[ f_1(\sC, \Xa) =  -\eff(\sC, \Xa), \; f_2(\sC, \Xa) = \avcost(\sC, \Xa) \text{ and }  f_3(\sC) = \size(\sC) \] 
% \end{problem}
% % \vspace{-.5em} 
% The requirement for a small set of actions is to enhance the interpretability of the explanation. This can also be expressed as a constraint on the set size we can afford. In this case, the first objective of Problem~\ref{prob:gce} is replaced by the constraint
% $\size(\sC) \leq s$,  
% % : \[\size(\sC) = s,\]  
% where $s$ is a small positive integer. For the rest of the paper, we will use the following problem formulation: 
% % \inote{more info for size, why do we want it small, etc}

\begin{problem}[$s$-GCE] \label{prob:gce-size}
    % Given the negative instances $\Xa$, the set of all actions $\sA$, our goal is to find the set $\sC \subseteq \sA$, $\size(\sC) =s$
    % % set of actions $A, A \neq \emptyset$ 
    % that represents a solution to the following bi-objective optimization problem:
    % \begin{align*}
    % \min \quad&\{-\eff(\sC, \Xa), \avcost(\sC, \Xa)\}\\
    % \text{s.t.} \quad&\size(\sC) =s
    % \end{align*}    
     Given a black box model $h$ that classifies the $\Xa$ instances to the negative class, our goal is to find the set $\sC \subseteq \sA$ 
     % set of actions $A, A \neq \emptyset$ 
    that represents a solution to the following multiobjective optimization problem:
% \vspace{-1em}
        \begin{align*}
    \text{minimize} \quad&\Big(-\eff(\sC, \Xa), \; \avcost(\sC, \Xa)\Big)\\
    \text{s.t.} \quad&\size(\sC) \leq s
    \end{align*}    
\end{problem}
We next show that a very restricted case of $s$-GCE is computationally difficult in the worst case.  

\begin{theorem}[NP-hardness]\label{thm:np-hard}
The special case of $s$-GCE, where the model $h$ and the set of allowable actions $\sA$ are explicitly given and the cost is ignored, is NP-hard. 
\end{theorem}

We prove that the decision version of the following special case of $s$-GCE is NP-complete. The input consists of a model $h$, a finite set of affected instances $\Xa = \{ x_1, \ldots, x_n \}$, a finite set of allowable actions $\sA = \{a_1, \ldots, a_m\}$ and a positive fractional number $E \in \mathbb{Q}$. We seek to determine if there is a subset of actions $\sC \subseteq \sA$ with $\size(\sC) \leq s$ and $\eff(\sC, \Xa) \geq E$. We assume that the model $h$ is defined only on $\Xa \cup \{ a_i(x_j)\,|\,a_i \in \sA\mbox{ and }x_j \in \Xa\}$ and its full description is given as part of the input. 
For the hardness part, we reduce Max $s$-Cover to the special case of $s$-GCE above. 
Max $s$-Cover is known to be NP-complete \citep[SP5]{GareyJ79} and inapproximable in polynomial time \citep[Theorem~5.3]{Feige98}. For the full proof, we refer the reader to Appendix~\ref{app:hardness}.

The fact that the very restricted special case of $s$-GCE in Theorem~\ref{thm:np-hard}, where the model $h$ and the set of allowable actions $\sA$ are explicitly given, is NP-hard (and NP-hard to approximate) indicates the computational challenges behind producing good enough solutions to $s$-GCE in practical settings, where we only have black-box access to $h$, the action cost is important and the set of allowable actions $\sA$ is unknown and potentially infinite. 

A key step of any approach is to efficiently generate a representative subset $\sA' \subseteq \sA$ of candidate actions, significantly larger than $s$, from which the final set of $s$ actions can be carefully chosen. 
Since actions that perform well locally may not generalize well as GCEs (as discussed, e.g., in Fig.~\ref{fig:example}), a myopic action selection (either as representative actions in $\sA'$ or as actions in the final solution $\sC$) may fail to produce results anywhere close to optimal. 
To further clarify this point, we note that effectiveness is a nondecreasing submodular function of the chosen actions set $\sC$ (and so well-fit for the standard greedy approach), but the average cost function is non-monotone in $\sC$, because its definition averages over the set of instances that receive recourse under $\sC$. 
At a conceptual level, increasing the size of $\sC$ may either leave effectiveness mostly unchanged, while potentially reducing the average cost, or improve effectiveness at an increased cost for the instances just added to $X_{\sC}$ (which might increase the average cost). Therefore, the addition of new actions must carefully balance gains in
effectiveness without disproportionately increasing the average cost. 

The discussion above aims to underscore the challenging trade-off between size, effectiveness, and average cost. 
Our method addresses this challenge by employing clustering in both feature and action space to ensure diversity and global representativeness of the preselected action set $\sA'$, thus maintaining high effectiveness and low average cost. 
A detailed analysis of how action set size impacts effectiveness and cost is provided in Appendix~\ref{app:tradeoffs}.

\section{GLANCE}
\label{sec:clustering}

We present \texttt{GLANCE}, a novel algorithm for solving Problem \ref{prob:gce-size}.
\texttt{GLANCE} is described in Algorithm~\ref{alg:it-merges} in pseudocode.

\begin{figure}[t!]
    \centering
    \begin{subfigure}[t]{0.3\textwidth}
        \centering
        \includegraphics[height=3.8cm]{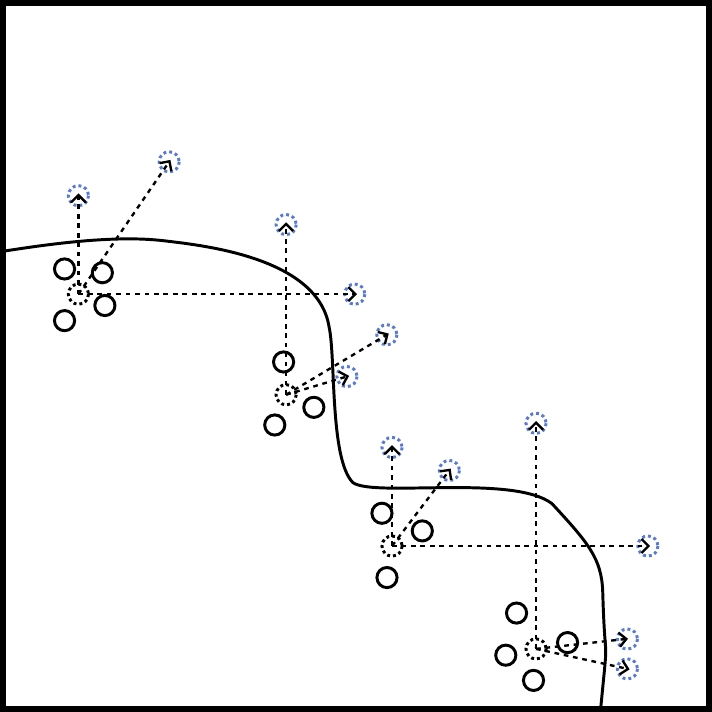}
        \caption{}
        \label{fig:actions}
    \end{subfigure}%
    ~ 
    \begin{subfigure}[t]{0.3\textwidth}
        \centering
        \includegraphics[height=3.8cm]{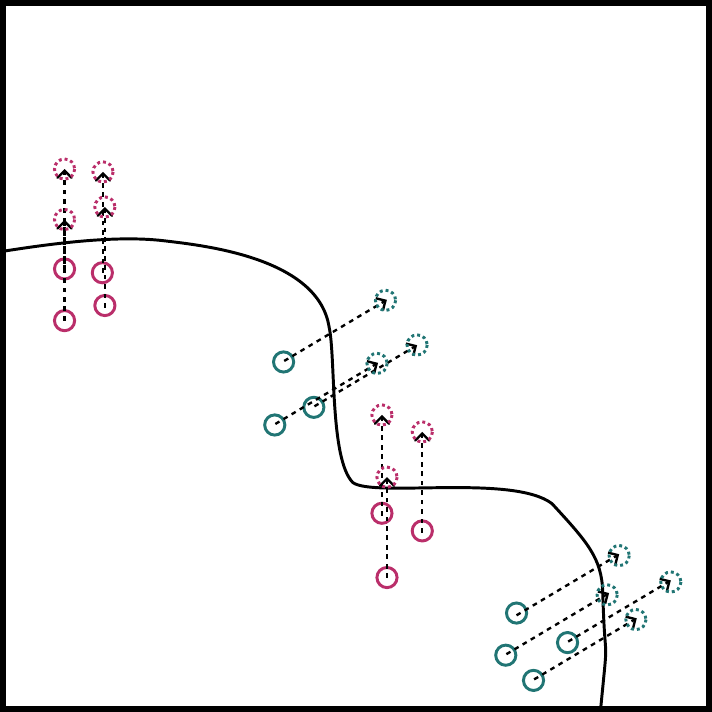}
        \caption{}
        \label{fig:glance}
    \end{subfigure}
    ~
    \begin{subfigure}[t]{0.3\textwidth}
        \centering
        \includegraphics[height=3.8cm]{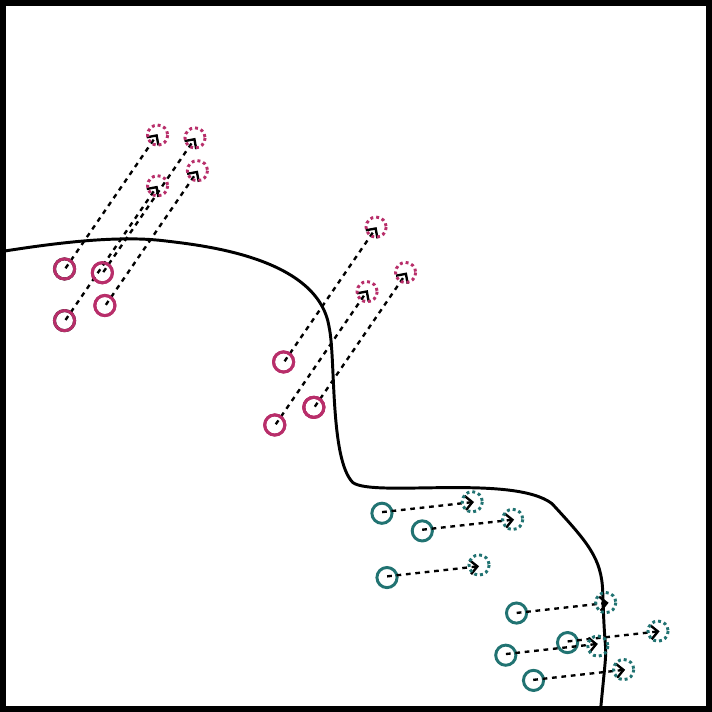}
        \caption{}
        \label{fig:clustering}
    \end{subfigure}%
    \caption{Intuition behind clustering approaches. (a) First, \texttt{GLANCE} generates diverse candidate actions from the centroids of feature-based clusters. 
    (b) Then, \texttt{GLANCE} merges clusters based on similarity in both feature and 
    action space, grouping instances that may be further apart but can be explained by similar actions.
    (c) Clustering solely in feature space can yield suboptimal global actions, 
    trading high effectiveness for high cost or low cost for low effectiveness.
    % either maximizing effectiveness at high cost or minimizing cost with low effectiveness.
    }
     \label{fig:motivation}
\end{figure}

\paragraph{Algorithm Description.}
The algorithm operates in two phases.
\emph{(1) Diverse Counterfactual Actions Generation.}
In the first phase, the algorithm produces a set of counterfactual actions while balancing the action diversity and sparsity (i.e., the number of actions).  The goal is to efficiently and effectively explore the action space by generating actions from widely dispersed points within the feature space and guiding them in diverse directions that cross the decision boundary. This is achieved by employing a clustering algorithm  (e.g., $k$-means, line 1, alg.~\ref{alg:it-merges}) to partition the feature space into $k$ clusters, based on the assumption that nearby points are likely to share similar views of the boundary. 
We compute the \emph{centroid} of each cluster and generate $m$ \emph{diverse} counterfactual actions for each centroid (line 3, alg.~\ref{alg:it-merges}), employing any candidate counterfactual generation method (see Appendix~\ref{app:candidate} for the methods used in the experimental evaluation). 
Fig.~\ref{fig:actions} shows an example with four fine-grained clusters, where three diverse actions are generated per centroid. Each action represents an alternative way to bring recourse to cluster members, giving \texttt{GLANCE} the flexibility to later select among them. 

\emph{(2) Extracting an optimal set of CGEs.} 
In the second phase, we identify a final set of $s$ actions from those produced in Phase 1. 
The approach combines the most similar clusters with the similarity function taking into account both their feature proximity and the proximity of their respective actions.
Specifically, similarity is computed by a metric \( D \), which is the sum of two components: \( d_1: \Xa \times \Xa \rightarrow \mathbb{R} \), the distance between the centroids of each cluster, and \( d_2: 2^\sA \times 2^\sA \rightarrow \mathbb{R} \), the distance between the action set of the clusters. The distance $d_2$ can be defined as the Wasserstein distance of the sets, though simpler formulations are also possible, e.g., computing the distance between the average counterfactual actions of each cluster; in the remainder of the paper, we adopt the latter formulation of $d_2$ for computational efficiency.
When similar actions bring recourse to individuals, it is easier to identify a common low-cost, highly effective action, even though they are dissimilar feature-wise. Fig.~\ref{fig:glance} shows an example where red-colored individuals are grouped through affinity either in the feature or in the action space. 
Until the desired number $s$ of clusters is reached, the algorithm iteratively merges (line 7, alg.~\ref{alg:it-merges}) the two clusters that minimize the total distance \( d_1 + d_2 \) (line 6, alg.~\ref{alg:it-merges}).
Merging two clusters also merges their corresponding action sets, ensuring that the most effective actions for each cluster are retained (line 8, alg.~\ref{alg:it-merges}). \\
Finally, for each of the $s$ clusters, we select a single action and return the compact set of $s$ global counterfactual actions (line 10, alg.~\ref{alg:it-merges}).
Assuming that action generation and subsequent merging have successfully grouped individuals that can achieve recourse through similar cost-efficient actions, the final step prioritizes effectiveness: we select, for each cluster, the optimal action in terms of effectiveness, among those associated with the cluster.

\begin{algorithm}[tb!]
   \caption{GLANCE}
   \label{alg:it-merges}

   \textbf{Input}: $\Xa$,  $m$, $k$, $s$ 
    \{$\Xa:=$ affected individuals ,  $m:=$ number of candidate actions to generate, $k:=$ initial number of clusters, $s:=$ number of global counterfactual actions\}\\
\textbf{Output}:  $s$ global counterfactual actions

\begin{algorithmic}[1]
\small
    \STATE $C \leftarrow \texttt{cluster}(\Xa,k) $ \COMMENT{Cluster $\Xa$ into $k$ initial clusters} 
    \FOR{$c \in C$} 
    \STATE $\bm{ca}(c) \leftarrow \texttt{actions}(\texttt{centroid}(c),m)$ \\ \COMMENT{For each cluster centroid generate $m$ actions}
    \ENDFOR
    \WHILE{$|C| > s$}
    \STATE $\{c_1,c_2\}$$\leftarrow$(\parbox[t]{.7\columnwidth}{$\argmin\limits_{\{c_1,c_2\}\subseteq C} d_1(\texttt{centroid}(c_1),\texttt{centroid}(c_2)) \\$+$d_2(\bm{ca}(c_1),\bm{ca}(c_2)))$\\
    \COMMENT{Find $c_1$,$c_2$ minimizing $d_1$+$d_2$}}
    \STATE $C\leftarrow \texttt{merge}(\{c_1,c_2\}\in C)$  
    \COMMENT{Replace $c_1$,$c_2$ in $C$ with the merged cluster}
    \STATE $\bm{ca}(\{c_1,c_2\}) \leftarrow \bm{ca}(c_1) \cup \bm{ca}(c_2)$ \COMMENT{Merge their action sets}
    \ENDWHILE
    \RETURN best action from each of the $s$ clusters
    % $\texttt{bca} (ca(c))$ for each $c \in C$ \COMMENT{Return optimal action from each of the $s$ clusters}
\end{algorithmic}
\end{algorithm}

\paragraph{Approach Strengths} 
As noted by \citet{kanamori2022counterfactual}, coarse-grained clustering approaches based solely on feature similarity lead to inadequate global actions that suffer either in cost or effectiveness (see Fig.~\ref{fig:clustering}). \texttt{GLANCE} mitigates these limitations by jointly considering both feature and action-space proximity, enabling the identification of cost-effective recourse actions. 

Although \texttt{GLANCE} explicitly prioritizes effectiveness, it inherently balances the effectiveness-cost trade-off. During the initial stage, refined clustering leads to diverse and typically low-cost candidate actions. In the second phase, the algorithm favors actions with maximal effectiveness, occasionally choosing higher-cost actions to achieve full coverage. For example, it may prefer an action with 100\% effectiveness and an average recourse cost of 2.3 over one with 98\% effectiveness and an average recourse cost of 1. However, affected individuals are ultimately free to select their preferred option among the final and applicable actions, often reducing the average recourse cost in practice.

While alternative selection strategies could explicitly incorporate cost, we deliberately avoided such heuristics. Recourse cost is often domain-specific, and \texttt{GLANCE} is designed to remain flexible for practitioners to adapt based on context and constraints.

Finally, \texttt{GLANCE} is modular and broadly applicable: it supports various clustering methods, cost metrics, and action generation techniques. 
As shown in Appendices \ref{app:exp_candidate} and \ref{app:exp_clustering}, different clustering methods and action generation techniques consistently lead to a near-optimal effectiveness-cost trade-off. 
Appendix~\ref{app:exp_initial} presents an ablation study analyzing the impact of the initial cluster number and the number of actions generated.
Across all datasets and models, \texttt{GLANCE} demonstrates fast runtimes, robust performance, and high-quality solutions.

\paragraph{Time Complexity}
Let $n$ denote the number of instances,  $k$ the number of initial clusters, $s$ global counterfactual actions, $m$ the number of candidate actions, and $d$ the number of features, respectively. Define $\gT_{\text{CF}}(d, \text{model})$ as the generation time for a single candidate counterfactual action and $\gT_{\text{model}}$ as the black-box model’s prediction time.
With $k$-means clustering requiring $I$ iterations to converge, 
the total time complexity of \texttt{GLANCE} is:
\[
\gO \big( k n d  I + k m  \gT_{\text{CF}}(d, \text{model}) +  k^2 (k-s) d  +k  n m   \gT_{\text{model}}) \big)
\]
We refer the reader to Appendix~\ref{app:complexity} for a detailed analysis.

\section{Evaluation}

\subsection{Experimental Setting}

\paragraph{Baselines} We compare \texttt{GLANCE} framework methods against state-of-the-art methods in Global Counterfactual Explanations, specifically: \texttt{AReS} (using the \texttt{Fast AReS} implementation), \texttt{CET}, \texttt{GroupCF}, and \texttt{GLOBE-CE}, all of which are constrained by a predefined action set size. 
Details on the implementation of each method can be found in Appendix~\ref{app:competing}.

\paragraph{Datasets} We use four established benchmark datasets from previous research: COMPAS \cite{COMPAS}, German Credit \cite{GermanCredit}, Default Credit \cite{Defaultcredit}, and HELOC \cite{HELOC}. Additionally, we introduce the Adult dataset \cite{Adult} for further evaluation. Details on the datasets and their preprocessing can be found in Appendix~\ref{app:datasets}.

\paragraph{Models} We trained three different model types: XGBoost (XGB), Logistic Regression (LR), and Deep Neural Network (DNN). 
Hyperparameters and training accuracy statistics are provided in Appendix~\ref{app:models}.
We used 5-fold cross-validation to also evaluate the robustness of the results.
\begin{table*}[tb]
\caption{Evaluating the effectiveness and average recourse cost of \texttt{GLANCE} against \texttt{Fast AReS}, \texttt{CET}, \texttt{Group-CF}, \texttt{GLOBE-CE}, and \texttt{dGLOBE-CE}) GCE methods for $s$-GCE problem with $s=4$. $s$-GCE solutions with effectiveness below 80\% (practicality threshold) are highlighted in \textcolor{red}{red}. Non-robust GCEs, identified by either a standard deviation (std) in effectiveness greater than 5\% across folds or a std in cost greater than half the average recourse cost, are highlighted in \textcolor{blue}{blue}.}
\label{tab:results_4}
\centering
\begin{tiny}
\resizebox{\textwidth}{!}{
\begin{tabular}{llcccccc}
\toprule
\multirow{2}{*}{Dataset} & \multirow{2}{*}{Method} & \multicolumn{2}{c}{DNN} & \multicolumn{2}{c}{LR} & \multicolumn{2}{c}{XGB} \\
\cmidrule(l){3-4}\cmidrule(l){5-6}\cmidrule(l){7-8}
 & & $\eff$ & $\avcost$ & $\eff$ & $\avcost$ & $\eff$ & $\avcost$ \\
\midrule
\multirow{6}{*}{Adult}
  & \texttt{Fast AReS}     & \textcolor{red}{12.39} ± \textcolor{red}{1.06} & \textcolor{red}{1.0} ± \textcolor{red}{0.0} & \textcolor{red}{11.74} ± \textcolor{red}{2.4} & \textcolor{red}{1.0} ± \textcolor{red}{0.0} & \textcolor{red}{6.13} ± \textcolor{red}{0.42} & \textcolor{red}{1.0} ± \textcolor{red}{0.0} \\
  & \texttt{CET}           & timeout & timeout & timeout & timeout & timeout & timeout \\
  & \texttt{Group-CF}      & 100.0 ± 0.0 & 10.08 ± 0.03 & 100.0 ± 0.0 & 1.71 ± 0.39 & 96.8 ± 1.72 & 1.41 ± 0.54 \\
  & \texttt{GLOBE-CE}      & 99.92 ± 0.0 & 3.34 ± 0.29 & 99.92 ± 0.0 & 2.34 ± 0.31 & 82.88 ± \textcolor{blue}{12.13} & 22.8 ± 7.87 \\
  & \texttt{dGLOBE-CE}     & 99.92 ± 0.0 & 10.89 ± 1.37 & 99.92 ± 0.0 & 5.91 ± 0.93 & 93.76 ± 1.98 & 64.76 ± 1.29 \\
  & \texttt{GLANCE}      & 100.0 ± 0.0 & 4.6 ± 0.73 & 100.0 ± 0.0 & 1.04 ± 0.07 & 99.85 ± 0.12 & 4.9 ± \textcolor{blue}{3.41} \\
\midrule
 \multirow{6}{*}{COMPAS}
  & \texttt{Fast AReS}     & \textcolor{red}{55.0} ± \textcolor{red}{0.86} & \textcolor{red}{1.21} ± \textcolor{red}{0.09} & \textcolor{red}{62.5} ± \textcolor{red}{1.82} & \textcolor{red}{1.24} ± \textcolor{red}{0.14} & \textcolor{red}{59.83} ± \textcolor{red}{3.12} & \textcolor{red}{1.1} ± \textcolor{red}{0.05} \\
  & \texttt{CET}           & \textcolor{red}{63.62} ± \textcolor{red}{10.35} & \textcolor{red}{0.96} ± \textcolor{red}{0.24} & \textcolor{red}{73.18} ± \textcolor{red}{4.34} & \textcolor{red}{1.24} ± \textcolor{red}{0.15} & \textcolor{red}{58.4} ± \textcolor{red}{9.3} & \textcolor{red}{1.06} ± \textcolor{red}{0.24} \\
  & \texttt{Group-CF}      & 100.0 ± 0.0 & 4.48 ± \textcolor{blue}{2.53} & 100.0 ± 0.0 & 3.97 ± \textcolor{blue}{2.38} & 100.0 ± 0.0 & 4.06 ± \textcolor{blue}{2.10} \\
  & \texttt{GLOBE-CE}      & 100.0 ± 0.0 & 2.82 ± 1.06 & 95.74 ± \textcolor{blue}{8.52} & 2.91 ± 0.57 & 87.17 ± \textcolor{blue}{11.09} & 4.73 ± 0.92 \\
  & \texttt{dGLOBE-CE}     & 100.0 ± 0.0 & 7.96 ± 3.91 & 100.0 ± 0.0 & 6.71 ± 0.23 & 99.84 ± 0.31 & 12.46 ± 3.42 \\
  & \texttt{GLANCE}      & 100.0 ± 0.0 & 2.34 ± 0.43 & 100.0 ± 0.0 & 2.33 ± 0.38 & 99.51 ± 0.46 & 2.96 ± 0.82 \\
\midrule
 \multirow{6}{*}{Default Credit}
  & \texttt{Fast AReS}     & \textcolor{red}{18.88} ± \textcolor{red}{2.16} & \textcolor{red}{1.0} ± \textcolor{red}{0.0} & \textcolor{red}{10.85} ± \textcolor{red}{5.45} & \textcolor{red}{1.07} ± \textcolor{red}{0.13} & \textcolor{red}{31.86} ± \textcolor{red}{5.12} & \textcolor{red}{1.05} ± \textcolor{red}{0.04} \\
  & \texttt{CET}           & 98.87 ± 0.62 & 6.32 ± 2.28 & 100.0 ± 0.0 & 3.79 ± 1.31 & 86.29 ± \textcolor{blue}{9.94} & 4.5 ± \textcolor{blue}{2.64} \\
  & \texttt{Group-CF}      & \textcolor{red}{79.6} ± \textcolor{red}{20.79} & \textcolor{red}{1.53} ± \textcolor{red}{0.62} & 95.4 ± \textcolor{blue}{9.2} & 1.94 ± \textcolor{blue}{1.2} & 95.2 ± 1.6 & 1.41 ± 0.64 \\
  & \texttt{GLOBE-CE}      & 81.19 ± 35.33 & 3.76 ± \textcolor{blue}{1.35} & 99.94 ± 0.07 & 2.91 ± \textcolor{blue}{1.55} & 83.69 ± \textcolor{blue}{6.72} & 17.211 ± 2.22 \\
  & \texttt{dGLOBE-CE}     & 87.38 ± \textcolor{blue}{18.69} & 5.96 ± \textcolor{blue}{4.14} & 99.94 ± 0.07 & 10.38 ± \textcolor{blue}{7.76} & 97.47 ± 0.82 & 42.58 ± 3.57 \\
  & \texttt{GLANCE}      & 100.0 ± 0.0 & 1.20 ± 0.40 & 100.0 ± 0.0 & 1.05 ± 0.11 & 98.13 ± 1.05 & 3.68 ± 1.64 \\
\midrule
 \multirow{6}{*}{German Credit}
  & \texttt{Fast AReS}     & \textcolor{red}{52.39} ± \textcolor{red}{1.63} & \textcolor{red}{1.0} ± \textcolor{red}{0.0} & \textcolor{red}{75.27} ± \textcolor{red}{2.96} & \textcolor{red}{1.0} ± \textcolor{red}{0.0} & \textcolor{red}{51.27} ± \textcolor{red}{1.57} & \textcolor{red}{1.0} ± \textcolor{red}{0.0} \\
  & \texttt{CET}           & 97.3 ± 2.46 & 1.58 ± 0.54 & 96.5 ± 2.85 & 2.42 ± 0.24 & 100.0 ± 0.0 & 2.73 ± 0.49 \\
  & \texttt{Group-CF}      & 97.8 ± 4.4 & 1.85 ± 0.13 & 97.6 ± 2.94 & 9.34 ± 3.85 & 100.0 ± 0.0 & 5.78 ± \textcolor{blue}{4.11} \\
  & \texttt{GLOBE-CE}      & 95.12 ± 2.04 & 2.11 ± 0.18 & \textcolor{red}{57.09} ± \textcolor{red}{20.03} & \textcolor{red}{2.27} ± \textcolor{red}{0.33} & \textcolor{red}{77.05} ± \textcolor{red}{11.26} & \textcolor{red}{2.52} ± \textcolor{red}{0.33} \\
  & \texttt{dGLOBE-CE}     & 97.36 ± 0.82 & 2.49 ± 0.27 & \textcolor{red}{69.89} ± \textcolor{red}{15.35} & \textcolor{red}{2.47} ± \textcolor{red}{0.23} & 86.96 ± \textcolor{blue}{9.79} & 2.66 ± 0.77 \\
  & \texttt{GLANCE}      & 95.31 ± 3.15 & 1.25 ± 0.33 & 100.0 ± 0.0 & 1.21 ± 0.06 & 100.0 ± 0.0 & 1.06 ± 0.03 \\
\midrule
 \multirow{6}{*}{HELOC}
  & \texttt{Fast AReS}     & \textcolor{red}{12.19} ± \textcolor{red}{0.58} & \textcolor{red}{1.03} ± \textcolor{red}{0.05} & \textcolor{red}{9.23} ± \textcolor{red}{1.24} & \textcolor{red}{1.12} ± \textcolor{red}{0.10} & \textcolor{red}{8.49} ± \textcolor{red}{1.32} & \textcolor{red}{1.16} ± \textcolor{red}{0.13} \\
  & \texttt{CET}           & 86.78 ± \textcolor{blue}{10.62} & 8.67 ± 3.25 & 100.0 ± 0.0 & 3.57 ± 1.48 & 86.78 ± \textcolor{blue}{6.70} & 12.51 ± 2.75 \\
  & \texttt{Group-CF}      & 80.4 ± \textcolor{blue}{10.17} & 3.09 ± 0.91 & 90.6 ± 3.93 & 2.40 ± \textcolor{blue}{1.38} & \textcolor{red}{78.4} ± \textcolor{red}{5.82} & \textcolor{red}{5.63} ± \textcolor{red}{1.93} \\
  & \texttt{GLOBE-CE}      & \textcolor{red}{47.18} ± \textcolor{red}{45.02} & \textcolor{red}{20.44} ± \textcolor{red}{24.18} & 99.9 ± 0.0 & 0.66 ± 0.10 & \textcolor{red}{28.33} ± \textcolor{red}{5.14} & \textcolor{red}{32.73} ± \textcolor{red}{0.48} \\
  & \texttt{dGLOBE-CE}     & 99.96 ± 0.05 & 11.07 ± \textcolor{blue}{8.6} & 99.9 ± 0.0 & 1.63 ± 0.35 & \textcolor{red}{77.64} ± \textcolor{red}{11.51} & \textcolor{red}{128.0} ±\textcolor{red}{0.0} \\
  & \texttt{GLANCE}      & 99.94 ± 0.05 & 11.24 ± 1.37 & 100.0 ± 0.0 & 1.55 ± 0.54 & 98.94 ± 0.66 & 19.99 ± 1.91 \\
\bottomrule

\end{tabular}
}
\end{tiny}
\end{table*}
\paragraph{Recourse Cost} Given the complexity of defining and computing the recourse cost for each action, we do not focus on the cost estimation and adhere to the guidelines established by \citet{pmlr-v202-ley23a}. 
Specifically, the cost in eq. (\ref{eq:rc}) is defined as the $L_1$ distance between the individual and the counterfactual point, i.e., $cost(a,x) = ||x-a(x)||_1$. For categorical features, this translates to the Hamming distance. For numerical features, this translates to the sum of the absolute differences in their values. However, due to some preprocessing (following the practices in \citet{pmlr-v202-ley23a}), the numerical features are split into 10 equal bins and their values are normalized to the bin size, resulting in a cost of one unit per decile.
\paragraph{Reproducibility}
All experiments were conducted on an in-house server with cloud infrastructure equipped with an Intel(R) Core(TM) i9-10900X CPU @ 3.70GHz, 128 GB of RAM. No GPU acceleration was utilized during these experiments. 
The code for reproducing our experiments can be found in the following GitHub repository: \url{https://github.com/AutoFairAthenaRC/GLANCE}.

\paragraph{Running Time}
In our experiments, \texttt{GLOBE-CE} and \texttt{dGLOBE-CE} were the fastest methods, typically delivering solutions in seconds, but required up to 30 seconds on datasets with many categorical features due to mandatory one-hot encoding, impacting scalability. \texttt{GLANCE} followed, typically completing in under 300 seconds. Other methods were slower, with \texttt{Fast AReS} ranging typically between 150--400 seconds, and peaking at 1,400 seconds in some runs. The least computationally efficient were \texttt{Group-CF} and \texttt{CET}, with maximum runtimes of 3,500 and 17,000 seconds, respectively. \texttt{CET} failed to solve the underlying optimization problem after 20 hours of runtime for the Adult dataset across all models.
\subsection{Experimental Evaluation}
Table~\ref{tab:results_4} presents the summarized results of all competing methods.
We compare \texttt{GLANCE} against the five other competitors across five datasets and three models, resulting in 75 head-to-head comparisons. However, since \texttt{CET} failed to solve the underlying optimization problem for the Adult dataset across all models, the final count is 72 comparisons per \texttt{GLANCE} method. 
Recall that all reported results concern the $s$-GCE problem for $s$ = $4$; 
% In the extended version, there is 
% \textcolor{purple}{The extended version of the paper presents full experimental details, along with additional results.}
Appendix~\ref{app:experiments} presents an extensive experimental evaluation, including detailed and additional results for $s$ = $4$ and results for $s$ = $8$ for all methods. 

\paragraph{Pareto Dominance Evaluation}
We summarize method performance by determining whether one solution dominates another based on effectiveness and cost.
Specifically, \emph{a solution $\sC$  of $s$-GCE \textbf{Pareto dominates} another solution $\sC'$ if it offers equal or better effectiveness and cost, and is strictly better in at least one of these objectives}.

\begin{table}[t!]
\caption{Pareto domination evaluation of solutions, for $s$-GCE problem with $s = 4$. The table reports the rate (number of times over available comparisons) at which \texttt{GLANCE} method dominates and is dominated by competitor methods.}
\label{tab:dominance}
\centering
\begin{small}
\begin{tabular}{cccccccccc}

\toprule
\texttt{GLANCE}  &     {\texttt{Fast AReS}} & {\texttt{CET}}  &{\texttt{GroupCF}} & {\texttt{GLOBE-CE}} & {\texttt{dGLOBE-CE}} & {$\sum_{\text{overall}}$}\\

% \texttt{GLANCE}  &     \rotatebox[origin=c]{90}{\texttt{Fast AReS}} & \rotatebox[origin=c]{90}{\texttt{CET}}  & \rotatebox[origin=c]{90}{\texttt{GroupCF}} & \rotatebox[origin=c]{90}{\texttt{GLOBE-CE}} & \rotatebox[origin=c]{90}{\texttt{dGLOBE-CE}} & \rotatebox[origin=c]{90}{$\sum_{\text{overall}}$}\\
\midrule
dominates  &      1/15       &           6/12          &        9/15        &         13/15 &12/15 &  41/72            \\
 is dominated &      0/15       &           0/12          &         0/15        &         0/15 &1/15  &  1/72            \\ 
\bottomrule
\end{tabular}
\end{small}
\end{table}

As shown in Table \ref{tab:dominance}, \texttt{GLANCE} dominates other methods in 41 out of 72 cases (57\%). It is dominated only once by \texttt{dGLOBE-CE} (in HELOC-DNN---cf.\ Table~\ref{tab:results_4}), where the performance of \texttt{GLANCE} is competitive in both terms of effectiveness and cost. 

\paragraph{Solution Practicality} In the prior Pareto-dominance evaluation (Table~\ref{tab:dominance}), we compare solutions with optimal or near-optimal effectiveness to many that exhibit insufficient
% unacceptable or unsatisfactory 
effectiveness. These lower-performing solutions are impractical for GCE, as the goal is to offer recourse to a large population segment. Solutions with low effectiveness fail to meet this goal, limiting their applicability in real-world scenarios. A solution that leaves a significant percentage of individuals without recourse undermines the very purpose of GCE, as noted by \citet{pmlr-v202-ley23a}. It is also important to note that achieving low recourse costs is easier for smaller subpopulations, especially those near the decision boundary, which explains the low domination scores for \texttt{Fast AReS} and \texttt{CET}. 
To address this, we include additional experiments (see Appendix~\ref{app:experiments}), where we cap \texttt{GLANCE}'s effectiveness, leading to lower costs and dominance over the baselines.

We consider a solution to be practical if it achieves effectiveness of at least 80\%. This number could be customizable depending on the dataset and the criticality of the application. Previous papers \cite{pmlr-v202-ley23a} used smaller thresholds that were too low to be meaningful
In Table~\ref{tab:results_4}, impractical solutions are shown in \textcolor{red}{red}. \texttt{GLANCE} never produces impractical solutions, whereas all \texttt{Fast AReS} outputs are impractical, and the remaining methods return 2--5 impractical solutions each.
\paragraph{Robustness} We expect methods to be robust, \emph{consistently} generating highly effective and low-cost GCEs across different data splits, which is crucial for real-world deployment.
Without robustness, recourse actions can vary significantly, undermining trust and leading to unfair outcomes, especially in critical areas like healthcare or finance. Evaluating the stability of effectiveness and cost metrics across different folds is key to determining the practical applicability of a counterfactual explanation method. 
Standard deviation measures this stability. An effectiveness deviation above 5\% indicates an inconsistency in providing recourse, while a cost deviation exceeding half the average suggests unpredictable actions; since cost scales vary by dataset, uniform robustness thresholds are not always appropriate.. These fluctuations make a solution unreliable, and we highlight them in \textcolor{blue}{blue} in Table~\ref{tab:results_4}. 
GLANCE is fully robust in effectiveness (15/15) and nearly so in cost (14/15). Even when excluding low-effectiveness ($<$80\%) solutions, it remains the top performer; the next best methods (dGLOBE-CE, Group-CF) achieve only 11/13 and 10/13.
Fig.~\ref{fig:cost-eff-plots-4acts} of the Appendix visualizes Table~\ref{tab:results_4}, showing effectiveness vs. cost (scaled 0–1) with standard deviation, illustrating cost-effectiveness trade-offs and method robustness across scenarios.

\section{User Study}

To better understand how humans evaluate Global Counterfactual Explanations (GCEs), we conducted an online user study following best practices outlined in \citep{pmlr-v202-ley23a,chowdhury2022equi,warren2023explaining}. The study consisted of two parts and had three primary objectives: (1) to assess how participants weigh trade-offs between effectiveness, average recourse cost, and size; (2) to validate our evaluation metrics, particularly the notions of practicality and robustness, by examining how variance in cost and effectiveness influences user preferences; and (3) to evaluate how participants rank \texttt{GLANCE} relative to baseline methods in both dominated and non-dominated scenarios. We recruited 55 participants--primarily PhD students and ML researchers--from six countries. Full user study details and data analysis are presented in Appendix~\ref{app:user}; below, we summarize the key findings.

In the first part of the study, participants ranked GCEs produced by three methods: \texttt{GLANCE} with $s$ = $3$, by \texttt{GLOBE-CE} with $s$ = $3$, and by \texttt{GLOBE-CE} in its default configuration (resulting in sizes ranging from 58 to 526). 
Rankings were aggregated using Borda Count \cite{fishburn1976borda}, resulting in the following order: \texttt{GLANCE} $|$ $s$ $=$ 3 $\succ$ \texttt{GLOBE-CE} $\succ$ \texttt{GLOBE-CE} $|$ $s$=3  with total Borda scores of 473, 277, and 241, respectively. \texttt{GLANCE} was significantly preferred over both variations of \texttt{GLOBE-CE}. While the default \texttt{GLOBE-CE} configuration achieved higher effectiveness and lower cost, participants frequently favored smaller action sets, prioritizing interpretability. These findings suggest that explanation size strongly shapes perceived usefulness, even when effectiveness and cost trade-offs are present. 

In the second part of the study, participants made pairwise comparisons between anonymized methods under various trade-off scenarios. \texttt{GLANCE} was unanimously preferred over impractical baselines. Notably, \texttt{GLANCE} was favored by  74.5\% of participants when % it was 
formally dominated by a competitor, with participants citing its lower variance as the rationale. In non-dominated scenarios, \texttt{GLANCE} was preferred in 71.5\% of cases on average. These preferences were statistically significant ($p$ $<$ $0.01$), indicating that robustness, and in particular, lower variance, can outweigh strict numerical dominance in human evaluations.

Participants' justifications revealed nuanced reasoning: over half prioritized effectiveness overall, but nearly all participants carefully considered trade-offs, including robustness in borderline cases. Several participants noted difficulty in interpreting cost without domain context,
% and knowledge,
underscoring the need for flexible cost metrics. These findings 
% not only 
validate our evaluation criteria and highlight the practical value of concise, stable, and interpretable action sets, such as those produced by \texttt{GLANCE}.

\section{Conclusion}

This paper presents \texttt{GLANCE}, a flexible framework for generating global counterfactual explanations that optimize the trade-off between effectiveness, cost, and interpretability. Extensive experiments show that \texttt{GLANCE} outperforms state-of-the-art methods by producing more effective and cost-efficient counterfactuals, under size constraints. A user study further confirms that the generated explanations are more interpretable, highlighting the practical advantages of our approach.

% \section*{Ethical Statement.}
% You can write a statement about the potential ethical impact of your work, including its broad societal implications, both positive and negative. If included, such statement must be written in an unnumbered section titled \emph{Ethical Statement}.

\section*{Acknowledgments and Disclosure of Funding}
This work was conducted while Dimitrios Rontogiannis was affiliated with the Athena Research Center and the National and Kapodistrian University of Athens.

This work has been partially supported by the European Union’s Horizon Europe research and innovation programme under Grant Agreements No.\ 101070568 (AutoFair), No.\ 101181895 (WiseFood), and No.\ 101135826 (AI-DAPT) and by the
CoDiet project, which is funded by the European Union under Horizon Europe
(grant number 101084642) and supported by UK Research and Innovation
(UKRI) under the UK government’s Horizon Europe funding guarantee
(grant number 101084642).

\bibliography{aaai2026}
\bibliographystyle{plainnat}
\newpage
\appendix

% \input{sections/intro}
% \input{sections/related}
% \input{sections/problem}
% \input{sections/clustering}
% % \input{sections/explainable}
% \input{sections/experimental}
% % \input{sections/user_study}
% \input{sections/conclusion}

\appendix
\section*{Technical Appendix}
The appendix is organized as follows:
\begin{itemize}
    \item Appendix \ref{app:hardness} includes the proof of Theorem 2.
    \item Appendix \ref{app:tradeoffs} provides a detailed analysis of the inherent trade-offs between solution size, effectiveness, and cost.
    \item Appendix \ref{app:complexity} presents a time complexity analysis of the \texttt{GLANCE} algorithm.
    \item Appendix \ref{app:datasets} describes the datasets used in our experiments, along with any preprocessing steps applied.
    \item Appendix \ref{app:models} describes the models utilized in our experiments, along with their hyperparameters, and the classification accuracy when trained. 
    \item Appendix \ref{app:competing} outlines the competing baseline methods, including implementation details and hyperparameters. It also includes results under no size constraints for \texttt{GLOBE-CE}.
    \item Appendix \ref{app:candidate} describes methods used to generate candidate counterfactual actions. 
    \item Appendix \ref{app:experiments} explains the experimental procedure, provides additional summary results for $s=8$, detailed results for $s=4$ and $s=8$, and examines cases where \texttt{GLANCE} intentionally sacrifices effectiveness by altering its default action selection method.
    \item Appendix \ref{app:exp_candidate} presents results for \texttt{GLANCE} using  different candidate counterfactual generation methods (as described in Appendix \ref{app:candidate}).
\item Appendix \ref{app:exp_clustering} reports results using different clustering algorithms in the initial clustering step.
    \item Appendix \ref{app:exp_local} compares the effectiveness and average recourse cost of local counterfactuals per individuals to the GCEs produced by \texttt{GLANCE} for $s=4$.
    \item Appendix \ref{app:exp_initial} investigates the effect of varying the number of initial clusters and candidate actions on \texttt{GLANCE}'s performance.
    \item Appendix \ref{app:edge} discusses potential edge cases that can arise in recourse algorithms.
    \item Appendix \ref{app:user} describes the user study setup, methodology, and full results including statistical analysis.
\end{itemize}

\section{Hardness}\label{app:hardness}
\begin{proof}[Proof of Theorem 2]
We prove that the decision version of the following special case of $s$-GCE is NP-complete. The input consists of a model $h$, a finite set of affected instances $\Xa = \{ x_1, \ldots, x_n \}$, a finite set of allowable actions $\sA = \{a_1, \ldots, a_m\}$ and a positive fractional number $E \in \mathbb{Q}$. We seek to determine if there is a subset of actions $\sC \subseteq \sA$ with $\size(\sC) \leq s$ and $\eff(\sC, \Xa) \geq E$. We assume that the model $h$ is defined only on $\Xa \cup \{ a_i(x_j)\,|\,a_i \in \sA\mbox{ and }x_j \in \Xa\}$ and its full description is given as part of the input. 

Membership in NP follows from the fact that for any given $\sC$, we can compute $\size(\sC)$ and $\eff(\sC, \Xa)$ in time polynomial in the size of the input. For the hardness part, we reduce Max $s$-Cover to the special case of $s$-GCE above. 
 
In (the decision version of) Max $s$-Cover, we are given a universe $\mathcal{X} = \{ x_1, \ldots, x_n \}$ and a family $\mathcal{A} = \{ A_1, \ldots, A_m \}$ of subsets of $\mathcal{X}$. We want to decide if there is a collection $\mathcal{C} \subseteq \mathcal{A}$ of size at most $s$ so that the union of the subsets in $\mathcal{C}$ equals $\mathcal{X}$. Max $s$-Cover is known to be NP-complete \citep[SP5]{GareyJ79}. \citet[Theorem~5.3]{Feige98} proved that it is NP-hard to distinguish between yes-instances of Max $s$-Cover and no-instances where the largest fraction of $\mathcal{X}$ that can be covered by any $\mathcal{C} \subseteq \mathcal{A}$ of size at most $s$ is $1-1/e+\epsilon$, for any $\epsilon \in (0, 1/e)$.

Given an instance of Max $s$-Cover, we construct an instance of $s$-GCE where the set of affected instances $\Xa = \mathcal{X}$ and $\sA$ consists of $n$ actions, one action $a_i$ for each subset $A_i \in \mathcal{A}$. The model $h$ is a function from $\Xa \cup \{ a_i(x_j)\,|\,a_i \in \sA \mbox{ and }x_j \in \Xa\}$ to $\{ -1, +1 \}$. The counterfactual instances $a_i(x_j)$ can be defined arbitrarily as long as $\Xa$ and the  sets $\{ a_i(x_j)\,|\,x_j \in \Xa\}$ are mutually disjoint. For each $x_j \in \Xa$, $h(x_j) = -1$. For each action $a_i$ and each $x_j \in \Xa$, we have that $h(a_i(x_j)) = +1$ if and only if $x_j \in A_i$. We set $E = 1.0$ (due to the inapproximability result of 
\citet[Theorem~5.3]{Feige98}, $E$ can be set to any fractional number larger than $1-1/e$). 

By construction, the effectiveness $\eff(\sC, \Xa)$ of any subset of actions $\sC \subseteq \sA$  under the model $h$ defined above equals the fraction of $\mathcal{X}$ covered by the union of the subsets $A_i \in \mathcal{A}$ that correspond to actions $a_i \in \sC$. Therefore, the instance of $s$-GCE above is a yes-instance if and only if the corresponding instance of Max $s$-Cover is a yes-instance. Moreover, the construction above along with \citep[Theorem~5.3]{Feige98} imply that approximating the optimal effectiveness $\eff(\sC, \Xa)$ achievable by the best subset $\sC \subseteq \sA$ of $\size(\sC) \leq s$ in polynomial time within any factor strictly larger than $1-1/e$ (even for this very restricted special case of $s$-CGE) would imply that $\mathrm{P} = \mathrm{NP}$.  
\end{proof}

\section{Size-Cost and Size-Effectiveness Trade-offs}
\label{app:tradeoffs}
Let $\sC$ denote a set of actions and consider the effect of including a new action $a$ to $\sC$, resulting in $\sC'=\sC \cup \{a\}$. We analyze how this increase in the solution size ($\size$) impacts the effectiveness ($\eff$) and the average recourse cost ($\avcost $).  
\begin{enumerate}
    \item[C1] \textbf{The new action $a$ does not flip any additional instances.}\\
    In this scenario, the effectiveness remains the same, as no new instances gain recourse. However, the average recourse cost may be affected depending on the cost of the new action relative to those in $\sC$:
    \begin{itemize}
        \item If the cost of $a$ is lower than that of existing actions, it may provide a cheaper alternative for instances already receiving recourse, thereby reducing the average recourse cost.
\item If the cost of 
$a$ is higher, the existing actions will continue to provide recourse to their respective instances, and the average recourse cost will remain unchanged.

    \end{itemize}
    \item[C2] \textbf{The new action 
$a$ flips previously uncovered instances.}\\
   In this scenario, the effectiveness increases due to additional instances receiving recourse. The impact on average recourse cost again depends on the cost of the new action:
    \begin{itemize}
        \item If the cost of $a$ is less than the current average recourse cost, it introduces lower-cost recourse for additional instances, thereby reducing the overall average recourse cost.
\item If the cost of $a$ is greater than the current average recourse cost, it increases the total cost, leading to a higher average recourse cost.
    \end{itemize}
\end{enumerate}
This analysis highlights the inherent trade-offs between size, effectiveness, and cost, emphasizing the importance of carefully selecting new actions to achieve a desirable balance. Notably, achieving near-optimal effectiveness (close to 100\%) often requires the inclusion of costly actions that address outliers. Nonetheless, our method demonstrates the ability to achieve near-optimal effectiveness with relatively low average recourse costs across a broad range of datasets.

\section{Time Complexity Analysis}
\label{app:complexity}
Let $n$ be the number of instances, $k$ the initial number of clusters, $s$ the number of global counterfactual actions, $m$ the number of candidate actions to generate, and $d$ the number of features. Let $\gT_{CF}(d, \text{model})$ be the time of generating a single candidate counterfactual action and $\gT_{\text{model}}$ be the prediction time of the black-box model. 
Let's assume that $k$-means is used for the clustering and $I$ is the number of iterations required for convergence. Then \texttt{GLANCE} time complexity is:
\begin{itemize}
    \item \textbf{Clustering (K-means):} $\gO\big(n k d I\big)$
    \item \textbf{Action Generation:} $\gO\big(k\cdot m \cdot \gT_{CF}(d, \text{model})\big)$
    \item \textbf{Merging Clusters:} $\gO\big((k-s) k^2 d \big)$
    \begin{itemize}
       \item \textbf{Finding pair of closest clusters:}
    \begin{itemize}
        \item For one iteration: $\gO\big(k^2 d\big)$
        \item For all iterations: $\gO\big((k-s)  k^2  d\big)$
    \end{itemize}
    \item \textbf{Merging pair of closest clusters:}
    \begin{itemize}
        \item For one iteration: $\gO\big( d + d\big)$
        \item For all iterations: $\gO\big((k-s)  d\big)$
    \end{itemize}
    \end{itemize}
    \item \textbf{Evaluation and selection of final actions:} $\gO\big(m k n  \gT_{\text{model}}\big)$
\end{itemize}
After simplifying common terms, the total complexity is:
\[
\gO \big( n  k d  I + k m  \gT_{\text{CF}}(d,\text{model}) + d (k-s) k^2 + n k m \gT_{\text{model}} \big)
\]

\section{Datasets \& Preprocessing}
\label{app:datasets}

\begin{table}[ht!]
\caption{Summary of the datasets used in our experiments. Specifically, we list the number of instances (No.Inst), input dimensions (i.e., the number of continuous features plus the number of categorical after the one-hot-encoding preprocessing step performed internally by the models), the number of categorical and continuous features, and the number of instances used for training and testing, evaluated using a 5-fold cross-validation strategy.
}
\label{apptab:dataset-info}
\begin{center}
\begin{tiny}
\begin{tabular}{lccccccr}
\toprule
Dataset & No. Instances & Input Dim. & Categorical & Continuous & Train & Test \\
\midrule
Adult & 30161 & 102 & 8 & 5 & 24128 & 6033  \\
COMPAS & 6172 & 15 & 4 & 2 & 4937 & 1235 \\
Default Credit & 30000 & 91 & 9 & 14 & 24000 & 6000 \\
German Credit & 1000 & 71 & 17 & 3 & 800 & 200 \\
HELOC & 9871 & 23 & 0 & 23 & 7896 & 1975  \\
\bottomrule
\end{tabular}
% \end{scriptsize}
\end{tiny}
\end{center}
\end{table}

We use five publicly available datasets as benchmarks. Our choice is based on their established use in previous works.
% A short description follows.
Table~\ref{apptab:dataset-info} summarizes the datasets' information, including the number of instances, the number of categorical and continuous features, the input dimensions (i.e., the number of continuous features plus the number of categorical after a preprocessing step using one-hot-encoding), and the number of instances used for training and testing, evaluated using a 5-fold cross-validation strategy. This approach illustrates the resilience of our method across various splits.

The dataset preprocessing follows the approach in \citet{pmlr-v202-ley23a}, using the publicly available repository at \url{https://github.com/danwley/GLOBE-CE/}. For completeness, we also provide a short description of each dataset below.

\paragraph{Adult} The Adult dataset (also known as ``Census Income'' dataset) is designed for the task of predicting whether an individual's income exceeds \$ 50K/yr, based on census data. The data as well as more detailed information can be obtained at \url{https://archive.ics.uci.edu/dataset/2/adult}.

For the preprocessing of this dataset, we first drop the ``education-num'' feature due to redundancy with the ``education'' feature (which takes on string values). Next, we remove missing values and map the class labels `$<$=50K' and `$>$50K' to 0 and 1, respectively. Additional minor transformations, mostly involving data types, are applied and can be reviewed in our source code.

\paragraph{COMPAS} The COMPAS dataset (Correctional Offender Management Profiling for Alternative Sanctions) \cite{COMPAS} is available at: 
\url{https://github.com/propublica/compas-analysis/blob/master/compas-scores-two-years.csv}. 
Detailed description and information on the dataset can be found at
\url{https://www.propublica.org/article/how-we-analyzed-the-compas-recidivism-algorithm}. It categorizes recidivism risk based on several factors, including race.

For the preprocessing of this dataset, we have to drop the feature ``days\_b\_screening\_arrest'', as it contains missing values. We also turn jail-in and jail-out dates to durations and turn negative durations to 0. Some additional filters are taken from the COMPAS analysis by ProPublica: \url{https://github.com/propublica/compas-analysis}. Finally, the target variable's values are transformed into the canonical 0 for the negative class and 1 for the positive class.

\paragraph{Default Credit} The Default Credit dataset \cite{Defaultcredit} is designed to classify the risk of default on customer payments, aiming to support the development and assessment of models for predicting creditworthiness and the likelihood of loan default. It can be obtained at  \url{https://archive.ics.uci.edu/ml/datasets/default+of+credit+card+clients}. 

To properly work with this dataset, we needed to drop the ``ID'' feature, since it holds no useful information, and transform the target labels into the canonical 0 - 1 values.

\paragraph{German Credit} The German Credit dataset \cite{GermanCredit} classifies people described by a set of attributes as good or bad credit risks. A detailed description and the dataset can be found in  \url{https://archive.ics.uci.edu/ml/datasets/statlog+(german+credit+data)}. 

The only preprocessing step we performed for this dataset was the transformation of the target variable's values into 0 - 1.

\paragraph{HELOC} The HELOC (Home Equity Line of Credit) dataset \cite{HELOC} contains anonymized information about home equity line of credit applications made by real homeowners, classifying credit risk. It is available at \url{https://community.fico.com/s/explainable-machine-learning-challenge}. All the features on this dataset are numeric.

A substantial percentage of these features' values are missing, so the main preprocessing step we performed here was to remove rows where all values are missing, and then replace all remaining missing values with the median of the respective feature. Other than that, we only needed to transform target labels to 0-1.

% \clearpage

% \newpage
\section{Models and Hyperparameters}
\label{app:models}
% \todo{tables for models should change}

% In our experimental framework, three distinct models are employed: XGBoost (XGB), Logistic Regression (LR), and Deep Neural Networks (DNNs). We train these models with \emph{the same} 80:20 train-test split as \citet{pmlr-v202-ley23a}; the distinctive hyperparameters for each model were also obtained from \citet{pmlr-v202-ley23a}, facilitating a standardized basis for the comparative analysis of our methodologies.
In our experimental evaluation, we utilize three distinct models: XGBoost (XGB), Logistic Regression (LR), and Deep Neural Networks (DNNs). Following \citet{pmlr-v202-ley23a}, we maintain an 80:20 train-test split, but, instead of splitting once, we perform 5-fold cross-validation, using each fold as a test set while training on the remaining four. The distinctive hyperparameters for each model are described in detail in this section. Additionally, we showcase the performance metric (accuracy), with all reported accuracies presented as the mean and standard deviation across the folds, providing a standardized foundation for comparative analysis of our methodologies.

\paragraph{XGBoost (XGB)} Implementation from the \texttt{xgboost}\footnote{\url{https://xgboost.readthedocs.io/en/stable/}} library. Hyperparameter values for each dataset and the model's accuracy on the test set are shown in Table~\ref{apptab:xgb-info}.

\begin{table}[ht!]
\caption{XGBoost Hyperparameter Configurations.}
\label{apptab:xgb-info}
\begin{center}
\begin{small}
% \resizebox{1\columnwidth}{!}{
\begin{tabular}{lcccc}
\toprule
Dataset & Depth & Estimators & $\gamma,\alpha,\lambda$ & Test Accuracy \\
\midrule
Adult & 6 & 100 & 0,0,1 & 86.62\% $\pm$ 0.18\% \\
COMPAS & 4 & 100 & 1,0,1 & 67.95\% $\pm$ 1.73\% \\
Default Credit & 10 & 200 & 2,4,1 & 81.35\% $\pm$ 0.18\% \\
German Credit & 6 & 500 & 0,0,1 & 76.9\% $\pm$ 0.86\% \\
HELOC & 6 & 100 & 4,4,1 & 72.97\% $\pm$ 1.14\% \\
\bottomrule
\end{tabular}
% }
\end{small}
\end{center}
\end{table}

\paragraph{Logistic Regression (LR)} Implementation from  
% \href{https://scikit-learn.org/stable/modules/generated/sklearn.linear_model.LogisticRegression.html}{\texttt{sklearn}} 
\texttt{sklearn}\footnote{\url{https://scikit-learn.org/stable/modules/generated/sklearn.linear_model.LogisticRegression.html}} 
library. Hyperparameter values for each dataset and the model's accuracy on the test set are shown in Table~\ref{apptab:lr-info}.

\begin{table}[ht!]
\caption{Logistic Regression Hyperparameter Configurations.}
\label{apptab:lr-info}
\begin{center}
\begin{small}
% \resizebox{1\columnwidth}{!}{

\begin{tabular}{lcccc}
\toprule
Dataset & Max Iter. & Class Weights(0:1) & Test Accuracy \\
\midrule
Adult & 100 & 1:1 & 79.09\% $\pm$ 0.12\% \\
COMPAS & 1000 & 1:1 & 66.69\% $\pm$ 0.97\% \\
Default Credit & 1000 & 1:1 & 82.08\% $\pm$ 0.19\% \\
German Credit & 1000 & 1:1 & 74.80\% $\pm$ 1.91\% \\
HELOC & 3000 & 1:1 & 73.06\% $\pm$ 0.96\% \\
\bottomrule
\end{tabular}
% }
\end{small}
\end{center}
\end{table}
\paragraph{Deep Neural Network (DNN)} Implementation using \texttt{pytorch}\footnote{\url{https://pytorch.org/docs/stable/index.html}} library. Hyperparameter values for each dataset and the model's accuracy on the test set are shown in Table~\ref{apptab:dnn-info}.

\begin{table}[ht!]
\caption{Deep Neural Network Hyperparameter Configurations.}
\label{apptab:dnn-info}
\begin{center}
\begin{small}
\begin{tabular}{lccccc}
\toprule
Dataset & Hidden Layers Width & Test Accuracy \\
\midrule
Adult & [64, 32, 16] & 84.13\% $\pm$ 0.30\% \\
COMPAS & [64, 32, 16] & 66.98\% $\pm$ 1.27\% \\
Default Credit & [64, 32, 16, 8] & 81.37\% $\pm$ 0.13\% \\
German Credit & [8, 4] & 71.3\% $\pm$ 2.27\% \\
HELOC & [64, 32, 16, 8] & 70.97\% $\pm$ 0.54\% \\
\bottomrule
\end{tabular}
\end{small}
\end{center}
\end{table}

% \newpage
\section{Competing Methods}
\label{app:competing}

\paragraph{AReS}  The \texttt{AReS} method, first introduced by \citet{rawal2020beyond}, was executed using the enhanced \texttt{Fast AReS} version implemented by \cite{pmlr-v202-ley23a}. The code for this version was obtained from the following GitHub repository: \url{https://github.com/danwley/GLOBE-CE}. We used the default hyperparameters provided in this implementation.

\paragraph{CET} The \texttt{CET} method, introduced by \citet{kanamori2022counterfactual}, was executed using the original implementation that was available on the following GitHub repository: \url{https://github.com/kelicht/cet}. We had to replace the solver they used for the MIP solved at each node of their tree. The IBM CPLEX solver is proprietary, and we replaced it with the GUROBI solver, for which we had an available license. We used the default hyperparameters provided in this implementation.

\paragraph{GroupCF} The \texttt{GroupCF} method, introduced by \citet{warren2023explaining}, was implemented using custom code due to some ambiguity in the instructions provided in the original GitHub repository (\url{https://github.com/e-delaney/group_cfe}), which made the original implementation challenging to follow.

\paragraph{GLOBE-CE} The \texttt{GLOBE-CE} method, introduced by \citet{pmlr-v202-ley23a}, was executed based on the implementation available on the following GitHub repository: \url{https://github.com/danwley/GLOBE-CE}.

% \begin{table}[ht!]
% \caption{\texttt{GLOBE-CE} results across datasets and models without the action set size constraints.}
% \label{apptab:globece_results}
% \begin{center}
% \begin{small}
% \begin{tabular}{lccccccccc}
% \toprule
% \multirow{2}{*}{Dataset} 
% & \multicolumn{3}{c}{DNN} 
% & \multicolumn{3}{c}{LR} 
% & \multicolumn{3}{c}{XGB} \\
% \cmidrule(r){2-4} \cmidrule{5-7} \cmidrule(l){8-10}
% & $\eff$ & $\avcost$ & $\size$ 
% & $\eff$ & $\avcost$ & $\size$ 
% & $\eff$ & $\avcost$ & $\size$ \\
% \midrule
% Adult 
% & 100.0 ± 0.0 & 1.24 ± 0.06 & 322.4
% & 100.0 ± 0.0 & 1.2 ± 0.03 & 545.2
% & 93.76 ± 0.91 & 4.28 ± 4.46 & 46.8 \\
% COMPAS 
% & 100.0 ± 0.0 & 1.11 ± 0.6 & 186.2
% & 98.92 ± 8.52 & 1.41 ± 0.19 & 97.8
% & 92.34 ± 8.87 & 1.22 ± 0.27 & 42.6 \\
% Default Credit 
% & 98.9 ± 0.73 & 1.1 ± 0.04 & 18.4
% & 99.94 ± 0.07 & 1.08 ± 0.01 & 4.6
% & 94.62 ± 2.85 & 1.05 ± 0.28 & 40.6 \\
% German Credit 
% & 95.69 ± 2.42 & 1.12 ± 0.09 & 4.2
% & 71.29 ± 14.27 & 1.05 ± 0.09 & 1.5
% & 84.97 ± 10.46 & 1.17 ± 0.15 & 2.2 \\
% HELOC 
% & 84.32 ± 19.32 & 2.63 ± 1.03 & 224.8
% & 100.0 ± 0.0 & 0.44 ± 0.06 & 512.8
% & 30.32 ± 4.07 & 2.11 ± 0.47 & 17 \\
% \bottomrule
% \end{tabular}
% \end{small}
% \end{center}
% \end{table}

\begin{table}[ht!]
\caption{\texttt{GLOBE-CE} results across datasets and models without the action set size constraints.}
\label{apptab:globece_results}
\begin{center}
\begin{small}
\begin{tabular}{llccc}
\toprule
Model & Dataset & $\eff$ & $\avcost$ & $\size$ \\
\midrule
\multirow{5}{*}{DNN} 
& Adult & 100.0 ± 0.0 & 1.24 ± 0.06 & 322.4 \\
& COMPAS & 100.0 ± 0.0 & 1.11 ± 0.6 & 186.2 \\
& Default Credit & 98.9 ± 0.73 & 1.1 ± 0.04 & 18.4 \\
& German Credit & 95.69 ± 2.42 & 1.12 ± 0.09 & 4.2 \\
& HELOC & 84.32 ± 19.32 & 2.63 ± 1.03 & 224.8 \\
\midrule
\multirow{5}{*}{LR} 
& Adult & 100.0 ± 0.0 & 1.2 ± 0.03 & 545.2 \\
& COMPAS & 98.92 ± 8.52 & 1.41 ± 0.19 & 97.8 \\
& Default Credit & 99.94 ± 0.07 & 1.08 ± 0.01 & 4.6 \\
& German Credit & 71.29 ± 14.27 & 1.05 ± 0.09 & 1.5 \\
& HELOC & 100.0 ± 0.0 & 0.44 ± 0.06 & 512.8 \\
\midrule
\multirow{5}{*}{XGB} 
& Adult & 93.76 ± 0.91 & 4.28 ± 4.46 & 46.8 \\
& COMPAS & 92.34 ± 8.87 & 1.22 ± 0.27 & 42.6 \\
& Default Credit & 94.62 ± 2.85 & 1.05 ± 0.28 & 40.6 \\
& German Credit & 84.97 ± 10.46 & 1.17 ± 0.15 & 2.2 \\
& HELOC & 30.32 ± 4.07 & 2.11 ± 0.47 & 17 \\
\bottomrule
\end{tabular}
\end{small}
\end{center}
\end{table}

Additional results of \texttt{GLOBE-CE} can be found in Table~\ref{apptab:globece_results}. Here, we do not constrain the set of actions, and we observe better performance, especially in terms of cost. However, these solutions do not represent solutions for $s$-GCE problem, since the solution size can be up to an average of 545.2 actions using 5-fold cross-validation (in Adult-LR case).

\paragraph{dGLOBE-CE} The \texttt{dGLOBE-CE} method, introduced by \citet{pmlr-v202-ley23a}, was executed using the implementation available on their GitHub repository (\url{https://github.com/danwley/GLOBE-CE}).

\section{Candidate Counterfactual Action Generators}
\label{app:candidate}
In this section, we outline the methods used to generate candidate counterfactual explanations and present comparative results employing different generation methods. To demonstrate the modularity of our framework, we developed and utilized various methods, showing that our approach is not tied to a specific counterfactual generation technique.
The comparative results of these experiments are shown in Table \ref{apptab:results_local}.
For the results presented in Table~1 %\ref{tab:results_4} 
of the main paper and Table \ref{apptab:results_8}, DiCE \citep{mothilal2020explaining} was used to generate candidate counterfactuals.

\paragraph{DiCE}

Detailed information on its methodology can be found in the corresponding paper by \citet{mothilal2020explaining}, and we encourage interested readers to refer to it for a deeper understanding.

For the scope of this paper, it suffices to say that DiCE offers a number of alternative algorithms for finding local counterfactuals. In our experiments, we use the `random' algorithm, which selects features to change at random and tests whether the change results in a successful counterfactual.

The other algorithms were either unsuitable for a black-box setting (e.g., requiring gradients), computationally inefficient (e.g., genetic algorithm), or failed to generate cost-efficient counterfactuals (e.g., KDTree) during our initial experimentation.

\paragraph{Random Sampling}

We developed a method called `Random Sampling'.
% , inspired by the DiCE framework's `random' approach. 
To find counterfactuals for an affected instance, this method iteratively modifies its features one at a time. The process begins by randomly altering one feature at a time, generating multiple new candidate instances. Each modified instance is then evaluated by querying the black-box model to determine whether it qualifies as a valid counterfactual. The method proceeds to modify additional features, building a set of potential counterfactuals.

Key differences between our method and DiCE's random approach include:
\begin{enumerate}
    \item We focus on modifying only the top $k_f$ most important features (using permutation feature importance), set to 3 in all experiments.
    \item For categorical features, only the top $k_c$ most frequent categories among unaffected individuals are considered replacement candidates (set to 10 for all experiments).
    \item We also introduce vectorization in certain operations, improving computational efficiency over DiCE's implementation.
\end{enumerate}
% The first in-house method we developed is called `Random Sampling,' inspired by the DiCE framework's `random' approach. To find counterfactuals for an affected instance, this method iteratively modifies its features one at a time. It starts by altering a single feature at random several times, to produce several new candidate instances. It checks if the new instances qualify as counterfactuals, and then proceeds to modify a second feature, then a third, and so on. Each randomly generated instance is evaluated by querying the black-box model to verify whether it constitutes a valid counterfactual.

% The key distinctions between our method and the DiCE framework's random approach are as follows:
% \begin{itemize}
%     \item we only try to perform changes on the $k_f$ most important features (according to the permuation feature importance technique), set to 3 for all experiments showcased in this paper.
%     \item For categorical features, only the $k_c$ most frequent categories among unaffected individuals are considered as candidate replacements (set to 10 for all experiments of this paper).
%     \item (we believe that) we managed to make this - very specific - part of the DiCE framework slighty more computationally efficient by moderately increasing the use of vectorization in certain operations.
% \end{itemize}

\paragraph{Nearest Neighbors}

This method is implemented by storing all unaffected individuals in memory. When queried to provide $k$ counterfactuals for an affected individual, it retrieves the $k$ nearest neighbors from the set of unaffected instances based on their proximity to the affected individual. This approach ensures that the generated counterfactuals are valid and closely aligned with the original instance, improving the relevance of the recommendations.
% This method is implemented by storing all unaffected individuals in memory. When queried to provide $k$ counterfactuals for an affected individual, the method retrieves the $k$ nearest neighbors from the set of unaffected instances based on their proximity to the affected individual.
% This is a very simple implementation which is fitted by storing in memory all unaffected individuals. Then, when asked to provide $k$ counterfactuals for an affected individual, it returns the $k$ nearest neighbors that this individual has in the set of unaffected instances.

\paragraph{Nearest Neighbors Scaled}

This algorithm closely resembles the `Nearest Neighbors' one but introduces a key enhancement. Rather than returning the $k$ nearest neighbors, it performs a localized search along the multidimensional line connecting the affected individual and any neighbor. We sample points along this line segment, and if any are classified as positive by the model, they are returned instead, as they are closer to the affected instance, potentially offering a more cost-efficient recourse.
% This algorithm closely resembles the previous one but introduces a key enhancement. Rather than immediately returning the $k$ nearest neighbors, it first conducts a localized search along the multidimensional line connecting the affected individual and an unaffected instance. If any points along this line are classified as positive by the model, they are returned instead, as their proximity to the original affected instance is smaller.
% This algorithm is almost the same as the previous one, with one important addition. Here, instead of returning immediately the $k$ nearest neighbors, we first perform a small search on the multidimensional `line' between the affected and the unaffected instance. If there happen to exist points between them which are still classified as positive by the model, then we return those, since the distance to the original affected instance will be smaller.

\section{Experimental Results}
\label{app:experiments}
\subsection{Experimental Procedure}
We evaluate our algorithms by comparing them with the following methods: \texttt{Fast AReS} by \citet{ley2022global}, \texttt{CET}  by  \citet{kanamori2022counterfactual}, \texttt{GroupCF} by \citet{warren2023explaining}, \texttt{GLOBE-CE}, and \texttt{dGLOBE-CE} by \citet{pmlr-v202-ley23a}. 
For the competing methods, we used the default hyperparameters as found in their respective GitHub repositories, referenced in Appendix~\ref{app:competing}.

For \texttt{GLANCE}, we used the following default hyperparameters: 100 initial clusters for all datasets (30 for German Credit), 10 candidate local counterfactuals were generated per cluster, $k$-means as the clustering method, DiCE as the candidate action generation method.
% For selecting actions within clusters, we adopted the maximum effectiveness strategy to prioritize the most impactful interventions. 
To ensure reproducibility, we fixed the random seed to 13 for both the clustering process and local counterfactual generation. 

All of the methods are constrained to solve $s$-GCE problem, with $s= 4$ and $s=8$. 
These constraints are applied differently depending on the method. In \texttt{GLANCE} and \texttt{GroupCF}, we configure the resulting number of clusters to match the target sizes.
\texttt{GLOBE-CE} is constrained by a maximum of 5  or 9 scalars, while \texttt{dGLOBE-CE} is limited to 4 or 8 directions, with up to two scalars per direction. Lastly, \texttt{CET} is constrained with a maximum of 4 or 8 leaves.

\begin{figure*}[ht!]

    \centering
    \includegraphics[width=0.9\linewidth]{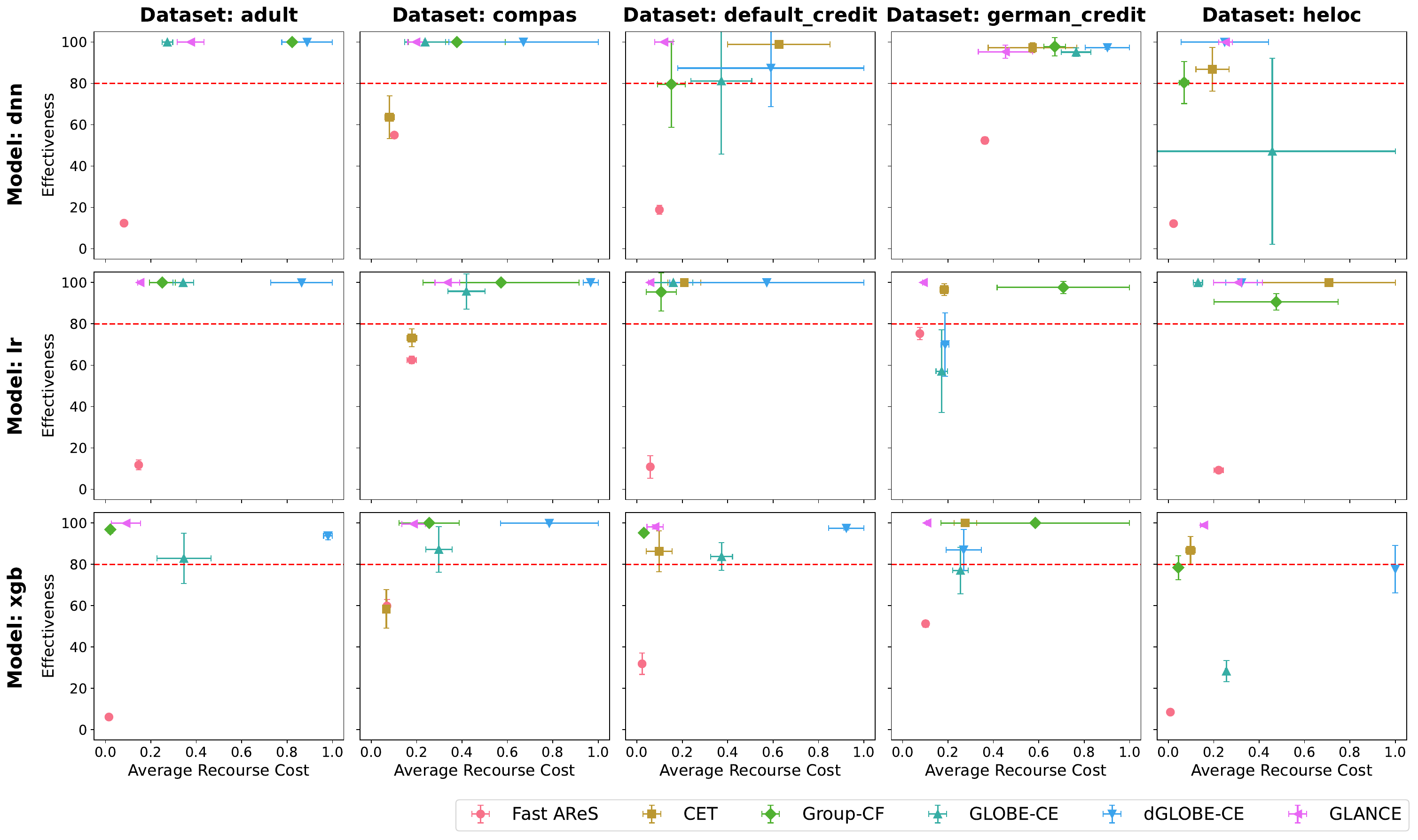}
  \caption{Comparison of effectiveness ($\eff$) and average recourse cost ($\avcost$), normalized with the maximum cost achieved in each dataset/model combination) for the solution of $s$-GCE with $s=4$. Standard deviations are represented by error bars. The red horizontal lines represent the $\eff>80\%$  threshold for evaluating the practicality of the solutions.}
  % \description{A visual representation of the results from Table 1, i.e., the results for the solution of s-GCE with s=4. The figure consists of a 3×5 grid, where each cell represents a dataset/model combination. Different methods are distinguished using unique symbols and colors. The position of each point reflects its effectiveness and normalized average cost, with lower cost and higher effectiveness being preferable. Error bars indicate standard deviations, helping to assess robustness. Overall, the figure highlights how different methods compare in performance across datasets.}
  \label{fig:cost-eff-plots-4acts}
\end{figure*}

\subsection{Summary of Experiments}
In Table 1 of the main paper, we present summarized results for $s = 4$, which are visualized in Figure~\ref{fig:cost-eff-plots-4acts}.  %~\ref{tab:results_4}
Complementary results for $s = 8$ can be found in Table~\ref{apptab:results_8} of the following section. The summarized results include the effectiveness and cost of each solution for all the datasets and model combinations. In Tables~\ref{apptab:results_adult}-\ref{apptab:results_heloc}, we provide detailed results for each model, adding runtime and the number of actions per dataset and method alongside effectiveness and cost.
\subsection{Experiments for Effectiveness-Cost Trade-offs} 
As shown in the main paper Table~1, %~\ref{tab:results_4}, 
our methods significantly outperform \texttt{CET}  and \texttt{Fast AReS} in terms of effectiveness. \texttt{Fast AReS} fails to exceed 80\% effectiveness in any case, and \texttt{CET} achieves near-optimal effectiveness in less than half the cases, sometimes failing to find a solution altogether. These low effectiveness levels make fair comparisons challenging, as solving for a small subset of the population is inherently easier.

To address this challenge, we demonstrate in Table~\ref{apptab:results_thresholds_4} that by intentionally lowering our solutions' effectiveness and selecting cost-efficient actions, our approach can still dominate other methods, achieving both greater effectiveness and lower cost. This adaptability underscores the strength of our methodology in maintaining superior performance even when optimizing for different trade-offs.

Our algorithm's default strategy selects optimal solutions based on maximum effectiveness. However, it also supports alternative strategies, such as selecting the lowest-cost solution, the lowest-cost solution above a specified effectiveness threshold, or the most effective solution below a specified cost. For the experiments presented Table~\ref{apptab:results_thresholds_4}, we used the ``lowest cost above a certain effectiveness threshold'' strategy in GLANCE, setting thresholds based on the effectiveness levels achieved by other methods. This flexibility highlights the algorithm's ability to align with various evaluation scenarios and user-defined trade-offs.

% \subsection{Additional Results for s = 4}

\begin{table}
\caption{Experimental results comparing \texttt{GLANCE} to baseline methods under constrained effectiveness. To enable fair comparison with less effective baselines, \texttt{GLANCE}'s effectiveness is intentionally reduced via thresholding, allowing the identification of cost-efficient dominating solutions in the low-effectiveness regime. We report the effectiveness ($\eff$) and average recourse cost ($\avcost$), and standard deviation for all methods.
}
\label{apptab:results_thresholds_4}
\begin{center}
\begin{small}
\resizebox{\textwidth}{!}{
\begin{tabular}{lclcccc}
\toprule
\multirow{2}{*}{Dataset} &\multirow{2}{*}{ Models} & \multirow{2}{*}{Competitors} & \multicolumn{2}{c}{Competitors Results} & \multicolumn{2}{c}{\texttt{GLANCE} Results} \\
\cmidrule(r){4-5} \cmidrule{6-7}
 & & &  $\eff$&  $\avcost$ & $\eff $ & $\avcost$\\

\midrule
\multirow{15}{*}{Adult}
 & \multirow{5}{*}{DNN} & \texttt{Fast AReS} & 12.39 ± 1.06 & 1.0 ± 0.0 & 26.59 ± 10.5 & 0.66 ± 0.23 \\
 &     & \texttt{CET} & NaN & NaN & NaN & NaN \\
 &     & \texttt{GroupCF} & 100.0 ± 0.0 & 10.08 ± 0.03 & 100.0 ± 0.0 & 1.73 ± 0.06 \\
 &     & \texttt{GLOBE-CE} & 99.92 ± 0.0 & 3.34 ± 0.29 & 99.96 ± 0.01 & 1.71 ± 0.05 \\
 &     & \texttt{dGLOBE-CE} & 99.92 ± 0.0 & 10.89 ± 1.37 & 99.96 ± 0.01 & 1.71 ± 0.05 \\ \cmidrule{3-7}
 & \multirow{5}{*}{LR}  & \texttt{Fast AReS} & 11.74 ± 2.4 & 1.0 ± 0.0 & 57.91 ± 24.77 & 0.63 ± 0.26 \\
 &     & \texttt{CET} & NaN & NaN & NaN & NaN \\
 &     & \texttt{GroupCF} & 100.0 ± 0.0 & 1.71 ± 0.39 & 100.0 ± 0.0 & 0.78 ± 0.06 \\
 &     & \texttt{GLOBE-CE} & 99.92 ± 0.0 & 2.34 ± 0.31 & 99.96 ± 0.02 & 0.78 ± 0.06 \\
 &     & \texttt{dGLOBE-CE} & 99.92 ± 0.0 & 5.91 ± 0.93 & 99.96 ± 0.02 & 0.78 ± 0.06 \\ \cmidrule{3-7}
 & \multirow{5}{*}{XGB} & \texttt{Fast AReS} & 6.13 ± 0.42 & 1.0 ± 0.0 & 31.16 ± 8.02 & 0.66 ± 0.13 \\
 &     & \texttt{CET} & NaN & NaN & NaN & NaN \\
 &     & \texttt{GroupCF} & 96.8 ± 1.72 & 1.41 ± 0.54 & 97.16 ± 0.41 & 0.81 ± 0.02 \\
 &     & \texttt{GLOBE-CE} & 82.88 ± 12.13 & 30.1 ± 10.39 & 94.04 ± 0.46 & 0.77 ± 0.01 \\
 &     & \texttt{dGLOBE-CE} & 93.76 ± 1.98 & 64.76 ± 1.29 & 94.04 ± 0.46 & 0.77 ± 0.01 \\
\midrule
\multirow{15}{*}{Default Credit}
 & \multirow{5}{*}{DNN} & \texttt{Fast AReS} & 18.88 ± 2.16 & 1.0 ± 0.0 & 21.54 ± 4.11 & 0.3 ± 0.1 \\
 &     & \texttt{CET} & 98.87 ± 0.62 & 6.32 ± 2.28 & 99.83 ± 0.1 & 0.82 ± 0.04 \\
 &     & \texttt{GroupCF} & 79.6 ± 20.79 & 1.53 ± 0.62 & 81.44 ± 2.18 & 0.8 ± 0.08 \\
 &     & \texttt{GLOBE-CE} & 81.19 ± 35.33 & 3.76 ± 1.35 & 81.44 ± 2.18 & 0.8 ± 0.08 \\
 &     & \texttt{dGLOBE-CE} & 87.38 ± 18.69 & 5.96 ± 4.14 & 91.22 ± 3.87 & 0.81 ± 0.07 \\ \cmidrule{3-7}
 & \multirow{5}{*}{LR}  & \texttt{Fast AReS} & 10.85 ± 5.45 & 1.07 ± 0.13 & 65.14 ± 15.62 & 0.55 ± 0.11 \\
 &     & \texttt{CET} & 100.0 ± 0.0 & 3.79 ± 1.31 & 100.0 ± 0.0 & 0.73 ± 0.07 \\
 &     & \texttt{GroupCF} & 95.4 ± 9.2 & 1.94 ± 1.2 & 99.05 ± 0.38 & 0.71 ± 0.07 \\
 &     & \texttt{GLOBE-CE} & 99.94 ± 0.07 & 2.91 ± 1.55 & 100.0 ± 0.0 & 0.73 ± 0.07 \\
 &     & \texttt{dGLOBE-CE} & 99.94 ± 0.07 & 10.38 ± 7.76 & 100.0 ± 0.0 & 0.73 ± 0.07 \\ \cmidrule{3-7}
 & \multirow{5}{*}{XGB} & \texttt{Fast AReS} & 31.86 ± 5.12 & 1.05 ± 0.04 & 40.03 ± 7.82 & 0.13 ± 0.04 \\
 &     & \texttt{CET} & 86.29 ± 9.94 & 4.5 ± 2.64 & 89.31 ± 2.77 & 0.73 ± 0.06 \\
 &     & \texttt{GroupCF} & 95.2 ± 1.6 & 1.41 ± 0.64 & 95.73 ± 0.18 & 1.01 ± 0.05 \\
 &     & \texttt{GLOBE-CE} & 83.69 ± 6.72 & 17.21 ± 2.22 & 83.77 ± 0.84 & 0.63 ± 0.08 \\
 &     & \texttt{dGLOBE-CE} & 97.47 ± 0.82 & 42.58 ± 3.57 & 97.64 ± 0.12 & 1.37 ± 0.25 \\
\midrule
\multirow{15}{*}{HELOC}
 & \multirow{5}{*}{DNN} & \texttt{Fast AReS} & 12.19 ± 0.58 & 1.03 ± 0.05 & 21.32 ± 5.13 & 0.11 ± 0.11 \\
 &     & \texttt{CET} & 86.78 ± 10.62 & 8.67 ± 3.25 & 92.67 ± 4.69 & 3.52 ± 1.09 \\
 &     & \texttt{GroupCF} & 80.4 ± 10.17 & 3.09 ± 0.91 & 82.31 ± 1.81 & 3.09 ± 0.9 \\
 &     & \texttt{GLOBE-CE} & 47.18 ± 45.02 & 20.44 ± 24.18 & 45.49 ± 2.1 & 1.19 ± 0.71 \\
 &     & \texttt{dGLOBE-CE} & 99.96 ± 0.05 & 11.07 ± 8.6 & 100.0 ± 0.0 & 4.22 ± 1.10 \\ \cmidrule{3-7}
 & \multirow{5}{*}{LR}  & \texttt{Fast AReS} & 9.23 ± 1.24 & 1.12 ± 0.1 & 71.55 ± 3.72 & 0.43 ± 0.04 \\
 &     & \texttt{CET} & 100.0 ± 0.0 & 3.57 ± 1.48 & 100.0 ± 0.0 & 0.72 ± 0.11 \\
 &     & \texttt{GroupCF} & 90.6 ± 3.93 & 2.4 ± 1.38 & 95.18 ± 1.75 & 0.6 ± 0.08 \\
 &     & \texttt{GLOBE-CE} & 99.9 ± 0.0 & 0.66 ± 0.1 & 99.92 ± 0.04 & 0.7 ± 0.11 \\
 &     & \texttt{dGLOBE-CE} & 99.9 ± 0.0 & 1.63 ± 0.35 & 99.92 ± 0.04 & 0.7 ± 0.11 \\ \cmidrule{3-7}
 & \multirow{5}{*}{XGB} & \texttt{Fast AReS} & 8.49 ± 1.32 & 1.16 ± 0.13 & 9.66 ± 1.12 & 0.4 ± 0.05 \\
 &     & \texttt{CET} & 86.78 ± 6.7 & 12.51 ± 2.75 & 90.72 ± 3.35 & 11.53 ± 2.94 \\
 &     & \texttt{GroupCF} & 78.4 ± 5.82 & 5.63 ± 1.93 & 82.46 ± 6.55 & 8.45 ± 3.1 \\
 &     & \texttt{GLOBE-CE} & 28.33 ± 5.14 & 32.73 ± 0.48 & 28.64 ± 0.99 & 0.85 ± 0.19 \\
 &     & \texttt{dGLOBE-CE} & 77.64 ± 11.51 & 128.0 ± 0.0 & 82.46 ± 6.55 & 8.45 ± 3.1 \\
\bottomrule
\end{tabular}
}
\end{small}
\end{center}
\end{table}

\subsection{Results for s = 8 } Table~\ref{apptab:results_8} presents the comparison of our method, \texttt{GLANCE}, with other state-of-the-art methods (\texttt{Fast AReS}, \texttt{CET}, \texttt{Group-CF}, \texttt{GLOBE-CE}, and \texttt{dGLOBE-CE}) for a maximum of 8 actions. The results indicate that \texttt{GLANCE} demonstrates superior performance in most cases, achieving nearly 100\% effectiveness with minimal and stable costs across all datasets and models. Exceptions are noted in the HELOC DNN and XGB experiments, where all methods report higher costs due to the dataset's numeric-only features. In contrast, \texttt{Fast AReS} struggles significantly, particularly on the Adult dataset, showing effectiveness as low as 13.8\% for DNN, 12.51\% for LR, and 7.19\% for XGB, explaining its lower costs.
Other methods, such as \texttt{GroupCF}, \texttt{CET}, \texttt{GLOBE-CE}, and \texttt{dGLOBE-CE}, generally perform well in terms of effectiveness but in almost all cases incur higher costs compared to \texttt{GLANCE}.

\begin{table}[tb]
\caption{Evaluating the effectiveness ($\eff$) and average recourse cost ($\avcost$) of \texttt{GLANCE} against \texttt{Fast AReS}, \texttt{CET}, \texttt{Group‑CF}, \texttt{GLOBE‑CE}, and \texttt{dGLOBE‑CE} GCE methods for the $s$‑GCE problem with $s=8$. $s$‑GCE solutions with effectiveness below 80\% are highlighted in \textcolor{red}{red}. Non‑robust GCEs, i.e., std in effectiveness greater than 5\% across folds or a std in cost greater than half the average recourse cost, are highlighted in  \textcolor{blue}{blue}.}
\label{apptab:results_8}
\begin{center}
\begin{small}
\resizebox{\textwidth}{!}{
\begin{tabular}{llcccccc}
\toprule
\multirow{2}{*}{Dataset} & \multirow{2}{*}{Method} & \multicolumn{2}{c}{DNN} & \multicolumn{2}{c}{LR} & \multicolumn{2}{c}{XGB} \\
 & & $\eff$ & $\avcost$ & $\eff$ & $\avcost$ & $\eff$ & $\avcost$ \\
\midrule
  \multirow{5}{*}{Adult}
  & \texttt{Fast AReS}     & \textcolor{red}{13.8} ± \textcolor{red}{0.92} & \textcolor{red}{1.0} ± \textcolor{red}{0.0} & \textcolor{red}{12.51} ± \textcolor{red}{2.44} & \textcolor{red}{1.0} ± \textcolor{red}{0.0} & \textcolor{red}{7.19} ± \textcolor{red}{0.3} & \textcolor{red}{1.0} ± \textcolor{red}{0.0} \\
  & \texttt{CET}           & nan ± nan & nan ± nan & nan ± nan & nan ± nan & nan ± nan & nan ± nan \\
  & \texttt{Group‑CF}      & 100.0 ± 0.0 & 11.54 ± 2.98 & 100.0 ± 0.0 & 2.33 ± \textcolor{blue}{1.91} & 99.4 ± 1.2 & 1.88 ± \textcolor{blue}{1.36} \\
  & \texttt{GLOBE‑CE}      & 99.92 ± 0.0 & 2.4 ± 0.12 & 99.92 ± 0.0 & 1.66 ± 0.9 & 86.33 ± \textcolor{blue}{8.81} & 11.7 ± 3.91 \\
  & \texttt{dGLOBE‑CE}     & 100.0 ± 0.0 & 11.85 ± 1.53 & 100.0 ± 0.0 & 6.12 ± 0.64 & 94.38 ± 2.01 & 62.37 ± 3.07 \\
  & \texttt{GLANCE}        & 100.0 ± 0.0 & 3.96 ± 0.56 & 100.0 ± 0.0 & 1.03 ± 0.07 & 99.87 ± 0.09 & 3.85 ± \textcolor{blue}{2.33} \\
\midrule
 \multirow{5}{*}{COMPAS}
  & \texttt{Fast AReS}     & \textcolor{red}{63.02} ± \textcolor{red}{1.58} & \textcolor{red}{1.11} ± \textcolor{red}{0.04} & \textcolor{red}{68.19} ± \textcolor{red}{0.6} & \textcolor{red}{1.03} ± \textcolor{red}{0.01} & \textcolor{red}{66.07} ± \textcolor{red}{2.5} & \textcolor{red}{1.14} ± \textcolor{red}{0.1} \\
  & \texttt{CET}           & \textcolor{red}{74.66} ± \textcolor{red}{5.33} & \textcolor{red}{1.02} ± \textcolor{red}{0.07} & 82.51 ± 2.97 & 1.51 ± 0.23 & \textcolor{red}{68.62} ± \textcolor{red}{15.67} & \textcolor{red}{1.3} ± \textcolor{red}{0.52} \\
  & \texttt{Group‑CF}      & 100.0 ± 0.0 & 5.65 ± \textcolor{blue}{3.14} & 100.0 ± 0.0 & 4.67 ± 2.29 & 100.0 ± 0.0 & 4.08 ± \textcolor{blue}{2.15} \\
  & \texttt{GLOBE‑CE}      & 100.0 ± 0.0 & 2.39 ± 0.87 & 95.74 ± \textcolor{blue}{8.52} & 2.25 ± 0.24 & 87.21 ± \textcolor{blue}{11.04} & 3.09 ± 0.47 \\
  & \texttt{dGLOBE‑CE}     & 100.0 ± 0.0 & 7.28 ± 1.25 & 100.0 ± 0.0 & 6.71 ± 0.23 & 99.96 ± 0.08 & 9.17 ± 2.01 \\
  & \texttt{GLANCE}        & 100.0 ± 0.0 & 1.77 ± 0.16 & 100.0 ± 0.0 & 1.69 ± 0.06 & 99.83 ± 0.24 & 2.06 ± 0.46 \\
\midrule
 \multirow{5}{*}{Default Credit}
  & \texttt{Fast AReS}     & \textcolor{red}{23.75} ± \textcolor{red}{1.81} & \textcolor{red}{1.02} ± \textcolor{red}{0.03} & \textcolor{red}{11.67} ± \textcolor{red}{6.18} & \textcolor{red}{1.04} ± \textcolor{red}{0.08} & \textcolor{red}{36.88} ± \textcolor{red}{5.46} & \textcolor{red}{1.07} ± \textcolor{red}{0.06} \\
  & \texttt{CET}           & 98.53 ± 2.11 & 3.68 ± 1.46 & 100.0 ± 0.0 & 2.0 ± \textcolor{blue}{1.36} & 90.01 ± \textcolor{blue}{10.44} & 2.61 ± 1.06 \\
  & \texttt{Group‑CF}      & 84.8 ± \textcolor{blue}{13.23} & 2.15 ± 0.98 & 100.0 ± 0.0 & 3.98 ± \textcolor{blue}{2.86} & 97.8 ± 0.75 & 1.92 ± \textcolor{blue}{1.33} \\
  & \texttt{GLOBE‑CE}      & 95.9 ± \textcolor{blue}{5.71} & 2.58 ± 0.78 & 99.94 ± 0.07 & 2.13 ± 0.89 & 85.63 ± \textcolor{blue}{6.0} & 9.78 ± 1.0 \\
  & \texttt{dGLOBE‑CE}     & 99.95 ± 0.04 & 6.77 ± \textcolor{blue}{3.67} & 99.94 ± 0.07 & 10.38 ± \textcolor{blue}{7.76} & 99.08 ± 0.66 & 28.73 ± 9.8 \\
  & \texttt{GLANCE}        & 100.0 ± 0.0 & 1.2 ± 0.4 & 100.0 ± 0.0 & 1.05 ± 0.11 & 98.87 ± 0.59 & 2.14 ± 0.18 \\
\midrule
\multirow{5}{*}{German Credit}
  & \texttt{Fast AReS}     & \textcolor{red}{65.47} ± \textcolor{red}{2.2} & \textcolor{red}{1.0} ± \textcolor{red}{0.0} & \textcolor{red}{71.9} ± \textcolor{red}{2.65} & \textcolor{red}{1.0} ± \textcolor{red}{0.0} & \textcolor{red}{68.88} ± \textcolor{red}{1.27} & \textcolor{red}{1.0} ± \textcolor{red}{0.0} \\
  & \texttt{CET}           & 100.0 ± 0.0 & 1.27 ± 0.37 & 98.75 ± 1.67 & 2.02 ± 0.33 & 100.0 ± 0.0 & 2.18 ± 0.37 \\
  & \texttt{Group‑CF}      & 99.6 ± 0.8 & 2.49 ± 0.82 & 98.9 ± 1.65 & 10.56 ± 2.43 & 100.0 ± 0.0 & 4.05 ± \textcolor{blue}{2.55} \\
  & \texttt{GLOBE‑CE}      & 95.55 ± 2.39 & 1.88 ± 0.36 & \textcolor{red}{57.55} ± \textcolor{red}{19.97} & \textcolor{red}{2.25} ± \textcolor{red}{0.33} & 77.05 ± \textcolor{blue}{11.26} & 2.49 ± 0.34 \\
  & \texttt{dGLOBE‑CE}     & 98.09 ± 1.72 & 2.34 ± 0.17 & \textcolor{red}{69.89} ± \textcolor{red}{15.35} & \textcolor{red}{2.47} ± \textcolor{red}{0.23} & 88.17 ± \textcolor{blue}{8.2} & 2.64 ± 0.78 \\
  & \texttt{GLANCE}        & 96.43 ± 3.6 & 0.86 ± 0.18 & 100.0 ± 0.0 & 1.18 ± 0.05 & 100.0 ± 0.0 & 1.02 ± 0.04 \\
\midrule
\multirow{5}{*}{HELOC}
  & \texttt{Fast AReS}     & \textcolor{red}{15.44} ± \textcolor{red}{1.91} & \textcolor{red}{1.05} ± \textcolor{red}{0.06} & \textcolor{red}{10.95} ± \textcolor{red}{1.73} & \textcolor{red}{1.11} ± \textcolor{red}{0.07} & \textcolor{red}{11.22} ± \textcolor{red}{2.05} & \textcolor{red}{1.09} ± \textcolor{red}{0.09} \\
  & \texttt{CET}           & 99.36 ± 0.92 & 8.86 ± 2.99 & 100.0 ± 0.0 & 2.99 ± 1.3 & 90.25 ± 2.51 & 12.99 ± 3.49 \\
  & \texttt{Group‑CF}      & 84.8 ± \textcolor{blue}{12.32} & 1.98 ± 0.5 & 96.6 ± 1.96 & 1.98 ± 0.5 & 81.4 ± 4.22 & 5.14 ± 1.46 \\
  & \texttt{GLOBE‑CE}      & \textcolor{red}{29.02} ± \textcolor{red}{39.47} & \textcolor{red}{7.88} ± \textcolor{red}{7.19} & 99.9 ± 0.0 & 0.55 ± \textcolor{blue}{0.08} & \textcolor{red}{29.18} ± \textcolor{red}{4.54} & \textcolor{red}{16.41} ± 0.3 \\
  & \texttt{dGLOBE‑CE}     & 100.0 ± 0.0 & 6.65 ± 0.15 & 99.9 ± 0.0 & 1.63 ± 0.35 & 81.95 ± \textcolor{blue}{10.38} & 128.0 ± 0.0 \\
  & \texttt{GLANCE}        & 99.98 ± 0.04 & 9.42 ± 2.22 & 100.0 ± 0.0 & 1.2 ± 0.12 & 99.0 ± 0.55 & 16.65 ± 2.44 \\
\bottomrule
\end{tabular}
}
\end{small}
\end{center}
\end{table}

We repeat the experimental evaluation utilizing Pareto-dominance for $s=8$.
As shown in Table~\ref{apptab:dominance_8} \texttt{GLANCE} dominates other methods in 38 out of 72 cases (53\%) while it is dominated once by \texttt{dGLOBE-CE} (in Heloc-DNN).

% \texttt{T-GLANCE} dominates other methods in 33 out of 72 comparisons (46\%),  and is dominated once by \texttt{dGLOBE-CE} (in Heloc-DNN). Overall, the \texttt{GLANCE} methods dominate other solutions in almost half of the cases (49\%) and are dominated in only 1\%. 

\begin{figure*}[ht!]
    \centering
    \includegraphics[width=0.9\linewidth]{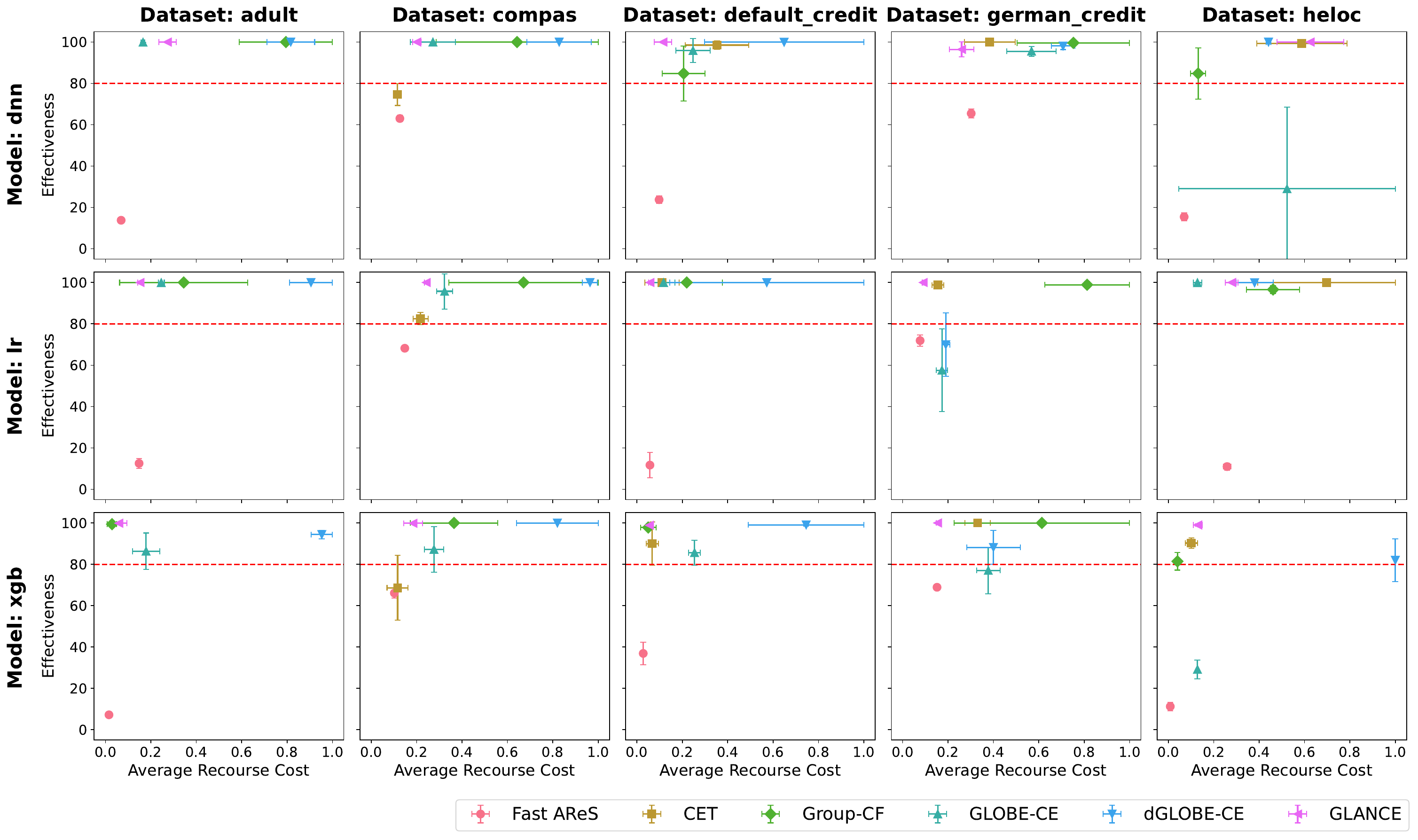}
    \caption{Comparison of effectiveness ($\eff$) and average recourse cost ($\avcost$), normalized with the maximum cost achieved in each dataset/model combination) for the solution of $s$-GCE with $s=8$. Standard deviations are represented by error bars.  The red horizontal lines represent the $\eff>80\%$ threshold for evaluating the practicality of the solution.}
    % \description{A visual representation of the results from Table 9, i.e., the results for the solution of s-GCE with s=8. The figure consists of a 3×5 grid, where each cell represents a dataset/model combination. Different methods are distinguished using unique symbols and colors. The position of each point reflects its effectiveness and normalized average cost, with lower cost and higher effectiveness being preferable. Error bars indicate standard deviations, helping to assess robustness. Overall, the figure highlights how different methods compare in performance across datasets.}
    \label{fig:cost-eff-plots-8acts}
\end{figure*}

\begin{table}[ht!]
\caption{Pareto domination evaluation of solutions, for $s$-GCE problem with $s = 8$. The table reports the rate (number of times over available comparisons) at which \texttt{GLANCE} method dominates and is dominated by competitor methods.}
\label{apptab:dominance_8}
\centering
\begin{center}
\resizebox{\textwidth}{!}{
\begin{tabular}{cccccccccc}
\toprule
\texttt{GLANCE}  &     {\texttt{Fast AReS}} & {\texttt{CET}}  &{\texttt{GroupCF}} & {\texttt{GLOBE-CE}} & {\texttt{dGLOBE-CE}} & {$\sum_{\text{overall}}$}\\
\midrule
 dominates  &      1/15       &           6/12          &          9/15        &         11/15 &11/15 & 38/72 \\   
 is dominated &      0/15       &           0/12          &          0/15        &         0/15 &1/15 & 1/72            \\ 
\bottomrule
\end{tabular}
}
\end{center}
\end{table}
% When evaluated within their respective explicit or implicit categories, \texttt{GLANCE} dominates other implicit methods in 69\% (31 out of 45) of the cases, outperforming its overall domination rate of 53\%. For \texttt{T-GLANCE}, the domination rate against explicit methods is 26\% (7 out of 27), lower than the 46\% overall domination rate. This decline is attributed to the high number of impractical solutions provided by competing explicit methods. Overall, the \texttt{GLANCE} framework demonstrates strong performance, dominating 49\% of cases within categories while keeping the percentage of dominated cases at a low 1\%.

\subsection{Detailed results}

\begin{table}[ht!]
\caption{Detailed results for the solution of $s$-GCE ($s=4$ and $s=8$) for Adult dataset. The table reports effectiveness ($\eff$), cost ($\avcost$), size ($\size$), and runtime, including their standard deviations for each method.}
\label{apptab:results_adult}
\begin{center}
\begin{small}
\resizebox{\textwidth}{!}{
\begin{tabular}{llcccccccc}
\toprule
\multirow{2}{*}{Model} & \multirow{2}{*}{Method} 
& \multicolumn{4}{c}{$s=4$} & \multicolumn{4}{c}{$s=8$} \\
\cmidrule(r){3-6} \cmidrule(l){7-10}
& & $\eff$ & $\avcost$ & $\size$ & RunTime 
  & $\eff$ & $\avcost$ & $\size$ & RunTime \\
\midrule
\multirow{6}{*}{DNN}
& \texttt{Fast AReS}     & 12.39 ± 1.06 & 1.0 ± 0.0 & 4.0 & 266.51 ± 127.13 & 13.8 ± 0.92 & 1.0 ± 0.0 & 8.0 & 271.04 ± 126.28 \\
& \texttt{CET}           & - & - & - & $>$20h or infeasible & - & - & - & $>$20h or infeasible \\
& \texttt{GroupCF}       & 100.0 ± 0.0 & 10.08 ± 0.03 & 4.0 & 3224.75 ± 0.03 & 100.0 ± 0.0 & 11.54 ± 2.98 & 8.0 & 3539.35 ± 2.98 \\
& \texttt{GLOBE-CE}      & 99.92 ± 0.0 & 3.34 ± 0.29 & 3.8 & 6.6 ± 0.22 & 99.92 ± 0.0 & 2.4 ± 0.12 & 7.0 & 6.56 ± 0.18 \\
& \texttt{dGLOBE-CE}     & 99.92 ± 0.0 & 10.89 ± 1.37 & 3.6 & 9.74 ± 0.21 & 100.0 ± 0.0 & 11.85 ± 1.53 & 6.4 & 14.09 ± 0.4 \\
& \texttt{GLANCE}        & 100.0 ± 0.0 & 4.6 ± 0.73 & 4.0 & 246.82 ± 0.0 & 100.0 ± 0.0 & 3.96 ± 0.56 & 8.0 & 240.16 ± 0.0 \\

\midrule
\multirow{6}{*}{LR}
& \texttt{Fast AReS}     & 11.74 ± 2.4 & 1.0 ± 0.0 & 4.0 & 124.2 ± 57.64 & 12.51 ± 2.44 & 1.0 ± 0.0 & 8.0 & 157.94 ± 73.44 \\
& \texttt{CET}           & - & - & - & $>$20h or infeasible & - & - & - & $>$20h or infeasible \\
& \texttt{GroupCF}       & 100.0 ± 0.0 & 1.71 ± 0.39 & 4.0 & 925.25 ± 0.39 & 100.0 ± 0.0 & 2.33 ± 1.91 & 8.0 & 872.1 ± 1.91 \\
& \texttt{GLOBE-CE}      & 99.92 ± 0.0 & 2.34 ± 0.31 & 4.0 & 6.48 ± 0.17 & 99.92 ± 0.0 & 1.66 ± 0.09 & 7.8 & 6.41 ± 0.08 \\
& \texttt{dGLOBE-CE}     & 99.92 ± 0.0 & 5.91 ± 0.93 & 4.0 & 9.36 ± 0.06 & 100.0 ± 0.0 & 6.12 ± 0.64 & 8.0 & 13.57 ± 0.22 \\
& \texttt{GLANCE}        & 100.0 ± 0.0 & 1.04 ± 0.07 & 4.0 & 194.83 ± 0.0 & 100.0 ± 0.0 & 1.03 ± 0.07 & 8.0 & 179.43 ± 0.0 \\

\midrule
\multirow{6}{*}{XGB}
& \texttt{Fast AReS}     & 6.13 ± 0.42 & 1.0 ± 0.0 & 4.0 & 108.66 ± 57.21 & 7.19 ± 0.3 & 1.0 ± 0.0 & 8.0 & 142.82 ± 67.84 \\
& \texttt{CET}           & - & - & - & $>$20h or infeasible & - & - & - & $>$20h or infeasible \\
& \texttt{GroupCF}       & 96.8 ± 1.72 & 1.41 ± 0.54 & 4.0 & 596.57 ± 0.54 & 99.4 ± 1.2 & 1.88 ± 1.36 & 8.0 & 1705.8 ± 1.36 \\
& \texttt{GLOBE-CE}      & 82.88 ± 12.13 & 22.8 ± 7.87 & 3 & 5.44 ± 0.18 & 86.33 ± 8.81 & 11.7 ± 3.91 & 3.6 & 5.58 ± 0.11 \\
& \texttt{dGLOBE-CE}     & 93.76 ± 1.98 & 64.76 ± 1.29 & 4.0 & 8.26 ± 0.14 & 94.38 ± 2.01 & 62.37 ± 3.07 & 7.4 & 12.01 ± 0.33 \\
& \texttt{GLANCE}        & 99.85 ± 0.12 & 5.98 ± 4.22 & 4.0 & 134.27 ± 0.0 & 99.87 ± 0.09 & 3.85 ± 2.33 & 8.0 & 126.65 ± 0.0 \\

\bottomrule
\end{tabular}
}
\end{small}
\end{center}
\end{table}

Table~\ref{apptab:results_adult} presents a comparative analysis of all algorithms for the Adult dataset. The \texttt{Fast AReS} method consistently performs poorly, achieving notably low effectiveness scores across all models. 

It is also noteworthy that in the Adult dataset, which is the largest of all in terms of instances and dimensions, the \texttt{CET} method did not run properly due to exceedingly high runtimes or infeasibility determined by the Gurobi optimizer. This is indicated by the “$>$20h or Infeasible (on Gurobi)” note present in the runtime column for \texttt{CET} (both $s=4$ and $s=8$) across all models.

On the other hand, \texttt{GroupCF}, \texttt{GLANCE}, and \texttt{GLOBE-CE} methods all perform notably well in terms of effectiveness. Under the DNN model \texttt {GLANCE}-4 and \texttt{GLANCE}-8 achieve perfect effectiveness scores of 100\% with similarly good cost performance. Notably, \texttt{GLOBE-CE}-8 achieves the best cost performance under the DNN model with an effectiveness score of 100\% and a cost of 2.4. 

In the LR model, \texttt{GLANCE} achieves 100\% effectiveness with costs of 1.04 and 1.03. Similarly, in the XGB model, \texttt{GLANCE}-4, and \texttt{GLANCE}-8 maintain high effectiveness scores of 99.85\%, and 99.87\%, with really good respective costs.

\begin{table}[ht!]
\caption{Detailed results for the solution of $s$-GCE ($s=4$ and $s=8$) for COMPAS dataset. The table reports effectiveness ($\eff$), cost ($\avcost$), size ($\size$), and runtime, including their standard deviations for each method.}
\label{apptab:results_compas}
\begin{center}
\begin{small}
\resizebox{\textwidth}{!}{
\begin{tabular}{llcccccccc}
\toprule
\multirow{2}{*}{Model} & \multirow{2}{*}{Method} & \multicolumn{4}{c}{$s=4$} & \multicolumn{4}{c}{$s=8$} \\
\cmidrule(l){3-6} \cmidrule(l){7-10}
& & $\eff$ & $\avcost$ & $\size$ & Runtime & $\eff$ & $\avcost$ & $\size$ & Runtime \\
\midrule
\multirow{6}{*}{DNN} 
& \texttt{Fast AReS} & 55.0 ± 0.86 & 1.21 ± 0.09 & 4 & 59.26 ± 26.71 & 63.02 ± 1.58 & 1.11 ± 0.04 & 8 & 65.84 ± 29.93 \\
& \texttt{CET} & 63.62 ± 10.35 & 0.96 ± 0.24 & 7.0 & 1339.42 ± 247.04 & 74.66 ± 5.33 & 1.02 ± 0.07 & 8.0 & 656.64 ± 119.51 \\
& \texttt{GroupCF} & 100.0 ± 0.0 & 4.48 ± 2.53 & 4 & 173.24 ± 2.53 & 100.0 ± 0.0 & 5.65 ± 3.14 & 8 & 210.34 ± 3.68 \\
& \texttt{GLOBE-CE} & 100.0 ± 0.0 & 2.82 ± 1.06 & 4.0 & 1.09 ± 0.06 & 100.0 ± 0.0 & 2.39 ± 0.87 & 7.8 & 1.06 ± 0.04 \\
& \texttt{dGLOBE-CE} & 100.0 ± 0.0 & 7.96 ± 3.91 & 3.4 & 1.55 ± 0.07 & 100.0 ± 0.0 & 7.28 ± 1.25 & 6.8 & 2.3 ± 0.09 \\
& \texttt{GLANCE} & 100.0 ± 0.0 & 2.34 ± 0.43 & 4.0 & 177.41 ± 0.0 & 100.0 ± 0.0 & 1.77 ± 0.16 & 8.0 & 177.35 ± 0.0 \\
\midrule
\multirow{6}{*}{LR} 
& \texttt{Fast AReS} & 62.5 ± 1.82 & 1.24 ± 0.14 & 4 & 11.52 ± 5.39 & 68.19 ± 0.6 & 1.03 ± 0.01 & 8 & 15.2 ± 6.9 \\
& \texttt{CET} & 73.18 ± 4.34 & 1.24 ± 0.15 & 4.0 & 663.22 ± 45.34 & 82.51 ± 2.97 & 1.51 ± 0.23 & 6.0 & 442.76 ± 64.47 \\
& \texttt{GroupCF} & 100.0 ± 0.0 & 3.97 ± 2.38 & 4 & 45.04 ± 2.38 & 100.0 ± 0.0 & 4.67 ± 2.29 & 8 & 48.53 ± 2.29 \\
& \texttt{GLOBE-CE} & 95.74 ± 8.52 & 2.91 ± 0.57 & 2.8 & 0.42 ± 0.01 & 95.74 ± 8.52 & 2.25 ± 0.24 & 6.4 & 0.45 ± 0.02 \\
& \texttt{dGLOBE-CE} & 100.0 ± 0.0 & 6.71 ± 0.23 & 4.0 & 0.61 ± 0.01 & 100.0 ± 0.0 & 6.71 ± 0.23 & 8.0 & 0.88 ± 0.01 \\
& \texttt{GLANCE} & 100.0 ± 0.0 & 2.33 ± 0.38 & 4.0 & 82.09 ± 0.0 & 100.0 ± 0.0 & 1.69 ± 0.06 & 8.0 & 84.11 ± 0.0 \\
\midrule
\multirow{6}{*}{XGB} 
& \texttt{Fast AReS} & 59.83 ± 3.12 & 1.1 ± 0.05 & 4 & 12.88 ± 5.5 & 66.07 ± 2.5 & 1.14 ± 0.1 & 8 & 17.69 ± 8.05 \\
& \texttt{CET} & 58.4 ± 9.3 & 1.06 ± 0.24 & 3.0 & 3280.17 ± 834.28 & 68.62 ± 15.67 & 1.3 ± 0.52 & 6.0 & 1787.18 ± 360.63 \\
& \texttt{GroupCF} & 100.0 ± 0.0 & 4.06 ± 2.1 & 4 & 36.07 ± 2.1 & 100.0 ± 0.0 & 4.08 ± 2.15 & 8 & 37.53 ± 2.15 \\
& \texttt{GLOBE-CE} & 87.17 ± 11.09 & 4.73 ± 0.92 & 2.4 & 1.06 ± 0.01 & 87.21 ± 11.04 & 3.09 ± 0.47 & 6.0 & 1.08 ± 0.03 \\
& \texttt{dGLOBE-CE} & 99.84 ± 0.31 & 12.46 ± 3.42 & 3.4 & 1.53 ± 0.03 & 99.96 ± 0.08 & 9.17 ± 2.01 & 5.6 & 2.23 ± 0.03 \\
& \texttt{GLANCE} & 99.51 ± 0.46 & 2.96 ± 0.82 & 4.0 & 286.26 ± 0.0 & 99.83 ± 0.24 & 2.06 ± 0.46 & 8.0 & 173.1 ± 0.0 \\
\bottomrule
\end{tabular}
}
\end{small}
\end{center}
\end{table}

Table~\ref{apptab:results_compas} presents a comparative analysis of all algorithms for the COMPAS dataset, revealing distinct performance patterns. \texttt{CET} and Fast \texttt{AReS}, both with 4 and 8 actions, show very low costs across all models; however, this is largely attributable to their relatively low effectiveness scores. 

Methods like \texttt{GLOBE-CE}, \texttt{dGLOBE-CE} display significantly higher effectiveness, mostly above 90\% across all models, \texttt{GroupCF} yields perfect effectiveness of 100\%. However, these methods come with higher costs, indicating that increased effectiveness is often tied to a rise in cost.

On the other hand, \texttt{GLANCE}  method consistently achieves near-perfect or perfect effectiveness ($\sim$ 100\%) across all models while maintaining much lower costs.
This balance of high effectiveness and low cost makes \texttt{GLANCE}the optimal method for the COMPAS dataset.
 
\begin{table}[ht!]
\caption{Detailed results for the solution of $s$-GCE ($s=4$ and $s=8$) for Default Credit dataset.  The table reports effectiveness ($\eff$), cost ($\avcost$), size ($\size$), and runtime, including their standard deviations for each method.}
\label{apptab:results_default}
\begin{center}
\begin{small}
\resizebox{\textwidth}{!}{
\begin{tabular}{llcccccccc}
\toprule
\multirow{2}{*}{Model} & \multirow{2}{*}{Method} 
& \multicolumn{4}{c}{$s=4$} & \multicolumn{4}{c}{$s=8$} \\
\cmidrule(r){3-6} \cmidrule(l){7-10}
& & $\eff$ & $\avcost$ & $\size$ & RunTime 
  & $\eff$ & $\avcost$ & $\size$ & RunTime \\
\midrule
\multirow{6}{*}{DNN} 
& \texttt{Fast AReS}   & 18.88 ± 2.16 & 1.0 ± 0.0 & 4 & 789.48 ± 306.69 & 23.75 ± 1.81 & 1.02 ± 0.03 & 8 & 428.04 ± 199.29 \\
& \texttt{CET}         & 98.87 ± 0.62 & 6.32 ± 2.28 & 3 & 4404.58 ± 453.7 & 98.53 ± 2.11 & 3.68 ± 1.46 & 4 & 2500.02 ± 328.94 \\
& \texttt{GroupCF}     & 79.6 ± 20.79 & 1.53 ± 0.62 & 4 & 2415.7 ± 0.62 & 84.8 ± 13.23 & 2.15 ± 0.98 & 8 & 2476.21 ± 0.82 \\
& \texttt{GLOBE-CE}    & 81.19 ± 35.33 & 3.76 ± 1.35 & 2.2 & 3.33 ± 0.2 & 95.9 ± 5.71 & 2.58 ± 0.78 & 4 & 3.49 ± 0.27 \\
& \texttt{dGLOBE-CE}   & 87.38 ± 18.69 & 5.96 ± 4.14 & 3.4 & 4.98 ± 0.21 & 99.95 ± 0.04 & 6.77 ± 3.67 & 4.8 & 6.79 ± 0.16 \\
& \texttt{GLANCE}      & 100.0 ± 0.0 & 1.2 ± 0.4 & 4 & 290.89 ± 0.0 & 100.0 ± 0.0 & 1.2 ± 0.4 & 8 & 289.46 ± 0.0 \\
\midrule
\multirow{6}{*}{LR} 
& \texttt{Fast AReS}   & 10.85 ± 5.45 & 1.07 ± 0.13 & 4 & 244.78 ± 128.93 & 11.67 ± 6.18 & 1.04 ± 0.08 & 8 & 272.64 ± 120.89 \\
& \texttt{CET}         & 100.0 ± 0.0 & 3.79 ± 1.31 & 1 & 3711.21 ± 164.94 & 100.0 ± 0.0 & 2.0 ± 1.36 & 2 & 1795.08 ± 182.55 \\
& \texttt{GroupCF}     & 95.4 ± 9.2 & 1.94 ± 1.2 & 4 & 306.76 ± 1.2 & 100.0 ± 0.0 & 3.98 ± 2.86 & 8 & 279.86 ± 2.86 \\
& \texttt{GLOBE-CE}    & 99.94 ± 0.07 & 2.91 ± 1.55 & 2 & 1.76 ± 0.04 & 99.94 ± 0.07 & 2.13 ± 0.89 & 2.4 & 1.75 ± 0.05 \\
& \texttt{dGLOBE-CE}   & 99.94 ± 0.07 & 10.38 ± 7.76 & 1.2 & 2.54 ± 0.07 & 99.94 ± 0.07 & 10.38 ± 7.76 & 1.2 & 3.51 ± 0.07 \\
& \texttt{GLANCE}      & 100.0 ± 0.0 & 1.05 ± 0.11 & 4 & 243.45 ± 0.0 & 100.0 ± 0.0 & 1.05 ± 0.11 & 8 & 226.96 ± 0.0 \\
\midrule
\multirow{6}{*}{XGB} 
& \texttt{Fast AReS}   & 31.86 ± 5.12 & 1.05 ± 0.04 & 4 & 226.08 ± 107.9 & 36.88 ± 5.46 & 1.07 ± 0.06 & 8 & 280.88 ± 135.37 \\
& \texttt{CET}         & 86.29 ± 9.94 & 4.5 ± 2.64 & 3 & 7982.05 ± 668.76 & 90.01 ± 10.44 & 2.61 ± 1.06 & 5 & 5105.11 ± 923.27 \\
& \texttt{GroupCF}     & 95.2 ± 1.6 & 1.41 ± 0.64 & 4 & 991.53 ± 0.64 & 97.8 ± 0.75 & 1.92 ± 1.33 & 8 & 631.31 ± 1.33 \\
& \texttt{GLOBE-CE}    & 83.69 ± 6.72 & 17.21 ± 2.22 & 3.4 & 2.44 ± 0.05 & 86.63 ± 6.0 & 9.78 ± 1.0 & 6.4 & 2.47 ± 0.06 \\
& \texttt{dGLOBE-CE}   & 97.47 ± 0.82 & 42.58 ± 3.57 & 4 & 3.48 ± 0.09 & 99.08 ± 0.66 & 28.73 ± 9.8 & 8 & 5.05 ± 0.12 \\
& \texttt{GLANCE}      & 98.13 ± 1.05 & 3.68 ± 1.64 & 4 & 319.45 ± 0.0 & 98.87 ± 0.59 & 2.14 ± 0.18 & 8 & 257.39 ± 0.0 \\
\bottomrule
\end{tabular}
}
\end{small}
\end{center}
\end{table}

Table~\ref{apptab:results_default} provides a comparative analysis of all algorithms for the Default Credit dataset. \texttt{GLANCE} method exhibits perfect effectiveness scores across both the DNN and LR models while maintaining the lowest costs, second only to \texttt{Fast AReS}, highlighting their superior cost-effectiveness.

In the XGB model, \texttt{GroupCF}-4 achieves a smaller cost compared to \texttt{GLANCE}-4, which, however, has better performance in terms of effectiveness. Concerning the case $s = 8$, \texttt{GroupCF}-8 has slightly better cost but worse effectiveness than \texttt{GLANCE}-8, and slightly better effectiveness. Overall, our method ensures competitive, robust,t and highly efficient performance.

\begin{table}[ht!]
\caption{
Detailed results for the solution of $s$-GCE ($s=4$ and $s=8$) for the German Credit dataset.  The table reports effectiveness ($\eff$), cost ($\avcost$), size ($\size$), and runtime, including their standard deviations for each method. }
\label{apptab:results_german}
\begin{center}
\begin{small}
\resizebox{\textwidth}{!}{
\begin{tabular}{llcccccccc}
\toprule
\multirow{2}{*}{Model} & \multirow{2}{*}{Method} 
& \multicolumn{4}{c}{$s=4$} & \multicolumn{4}{c}{$s=8$} \\
\cmidrule(l){3-6} \cmidrule(l){7-10}
& & $\eff$ & $\avcost$ & $\size$ & Runtime & $\eff$ & $\avcost$ & $\size$ & Runtime \\
\midrule
\multirow{6}{*}{DNN}
& \texttt{Fast AReS} & 52.39 ± 1.63 & 1.0 ± 0.0 & 4 & 119.34 ± 54.6 & 65.47 ± 2.2 & 1.0 ± 0.0 & 8 & 415.74 ± 175.82 \\
& \texttt{CET} & 97.3 ± 2.46 & 1.58 ± 0.54 & 3.0 & 230.36 ± 47.15 & 100.0 ± 0.0 & 1.27 ± 0.37 & 1.0 & 142.29 ± 27.72 \\
& \texttt{GroupCF} & 97.8 ± 4.4 & 1.85 ± 0.13 & 4 & 25.54 ± 0.13 & 99.6 ± 0.8 & 2.49 ± 0.82 & 8 & 28.93 ± 0.82 \\
& \texttt{GLOBE-CE} & 95.82 ± 2.04 & 2.11 ± 0.18 & 2.25 & 1.02 ± 0.01 & 95.55 ± 2.39 & 1.88 ± 0.36 & 2.2 & 1.03 ± 0.01 \\
& \texttt{dGLOBE-CE} & 97.36 ± 0.82 & 2.49 ± 0.27 & 4.0 & 1.57 ± 0.01 & 98.09 ± 1.72 & 2.34 ± 0.17 & 7.8 & 2.34 ± 0.03 \\
& \texttt{GLANCE} & 95.31 ± 3.15 & 1.25 ± 0.33 & 4.0 & 78.45 ± 0.0 & 96.43 ± 3.6 & 0.86 ± 0.18 & 8.0 & 78.38 ± 0.0 \\
\midrule
\multirow{6}{*}{LR}
& \texttt{Fast AReS} & 75.27 ± 2.96 & 1.0 ± 0.0 & 4 & 47.88 ± 23.06 & 71.9 ± 2.65 & 1.0 ± 0.0 & 8 & 59.8 ± 32.65 \\
& \texttt{CET} & 96.5 ± 2.85 & 2.42 ± 0.24 & 3.0 & 237.17 ± 11.2 & 98.75 ± 1.67 & 2.02 ± 0.33 & 6.0 & 156.14 ± 15.7 \\
& \texttt{GroupCF} & 97.6 ± 2.94 & 9.34 ± 3.85 & 4 & 7.91 ± 3.85 & 98.9 ± 1.65 & 10.56 ± 2.43 & 8 & 15.6 ± 2.33 \\
& \texttt{GLOBE-CE} & 57.09 ± 20.03 & 2.27 ± 0.33 & 1.0 & 0.44 ± 0.0 & 57.55 ± 19.97 & 2.25 ± 0.33 & 1.0 & 0.45 ± 0.0 \\
& \texttt{dGLOBE-CE} & 69.89 ± 15.35 & 2.47 ± 0.23 & 3.2 & 0.7 ± 0.01 & 69.89 ± 15.35 & 2.47 ± 0.23 & 5.6 & 1.02 ± 0.04 \\
& \texttt{GLANCE} & 100.0 ± 0.0 & 1.21 ± 0.06 & 4.0 & 61.58 ± 0.0 & 100.0 ± 0.0 & 1.18 ± 0.05 & 8.0 & 63.1 ± 0.0 \\
\midrule
\multirow{6}{*}{XGB}
& \texttt{Fast AReS} & 51.27 ± 1.57 & 1.0 ± 0.0 & 4 & 55.01 ± 27.27 & 68.88 ± 1.27 & 1.0 ± 0.0 & 8 & 64.77 ± 29.67 \\
& \texttt{CET} & 100.0 ± 0.0 & 2.73 ± 0.49 & 3.0 & 347.41 ± 30.21 & 100.0 ± 0.0 & 2.18 ± 0.37 & 5.0 & 239.2 ± 25.84 \\
& \texttt{GroupCF} & 100.0 ± 0.0 & 5.78 ± 4.11 & 4 & 5.87 ± 4.11 & 100.0 ± 0.0 & 4.05 ± 2.55 & 8 & 7.84 ± 2.55 \\
& \texttt{GLOBE-CE} & 77.05 ± 11.26 & 2.52 ± 0.33 & 1.0 & 1.34 ± 0.03 & 77.05 ± 11.26 & 2.49 ± 0.34 & 1.0 & 1.36 ± 0.05 \\
& \texttt{dGLOBE-CE} & 86.96 ± 9.79 & 2.66 ± 0.77 & 3.6 & 2.09 ± 0.04 & 88.17 ± 8.2 & 2.64 ± 0.78 & 6.2 & 3.06 ± 0.04 \\
& \texttt{GLANCE} & 100.0 ± 0.0 & 1.06 ± 0.03 & 4.0 & 67.36 ± 0.0 & 100.0 ± 0.0 & 1.02 ± 0.04 & 8.0 & 68.96 ± 0.0 \\
\bottomrule
\end{tabular}
}
\end{small}
\end{center}
\end{table}

Table~\ref{apptab:results_german} presents a comparative analysis of all algorithms for the German Credit dataset. \texttt{Fast AReS} remains the weakest method overall, continuing to exhibit poor effectiveness. Notably, while \texttt{CET} incurs a slightly higher cost than on the COMPAS dataset, it achieves significantly greater effectiveness, and the increase in cost is quite reasonable.
The \texttt{GLOBE-CE} and \texttt{dGLOBE-CE} methods demonstrate mixed results. They perform relatively well under the DNN model, achieving effectiveness scores above 90\%, but their effectiveness declines when applied to the LR and XGB models.
On the other hand, \texttt{GLANCE} continues to excel due to its robustness and consistently high performance. This method consistently reaches near-perfect effectiveness ($\sim$100\%, and only for DNN 95\%-97\%) across all models while maintaining very low costs. Notably, \texttt{GLANCE}-8 achieves the lowest cost of 0.86 under the DNN model, while still maintaining high effectiveness.

\begin{table}[ht!]
\caption{Detailed results for the solution of $s$-GCE ($s=4$ and $s=8$) for HELOC dataset.  The table reports effectiveness ($\eff$), cost ($\avcost$), size ($\size$), and runtime, including their standard deviations for each method.}
\label{apptab:results_heloc}
\begin{center}
\begin{small}
\resizebox{\textwidth}{!}{
\begin{tabular}{llcccccccc}
\toprule
\multirow{2}{*}{Model} & \multirow{2}{*}{Method} 
& \multicolumn{4}{c}{$s=4$} & \multicolumn{4}{c}{$s=8$} \\
\cmidrule(l){3-6} \cmidrule(l){7-10}
& & $\eff$ & $\avcost$ & $\size$ & RunTime 
  & $\eff$ & $\avcost$ & $\size$ & RunTime \\
\midrule

\multirow{6}{*}{DNN}
& \texttt{Fast AReS} & 12.19 ± 0.58 & 1.03 ± 0.05 & 4 & 1562.21 ± 656.59 & 15.44 ± 1.91 & 1.05 ± 0.06 & 8 & 1070.19 ± 558.28 \\
& \texttt{CET} & 86.78 ± 10.62 & 8.67 ± 3.25 & 1.0 & 11090.16 ± 4597.13 & 99.36 ± 0.92 & 8.86 ± 2.99 & 4.0 & 6758.87 ± 3384.51 \\
& \texttt{GroupCF} & 80.4 ± 10.17 & 3.09 ± 0.91 & 4 & 468.52 ± 0.91 & 84.8 ± 12.32 & 1.98 ± 0.5 & 8 & 501.45 ± 2.5 \\
& \texttt{GLOBE-CE} & 47.18 ± 45.02 & 20.44 ± 24.18 & 2.4 & 1.43 ± 0.13 & 29.02 ± 39.47 & 7.88 ± 7.19 & 2.2 & 1.44 ± 0.02 \\
& \texttt{dGLOBE-CE} & 99.96 ± 0.05 & 11.07 ± 8.6 & 2.0 & 2.2 ± 0.14 & 100.0 ± 0.0 & 6.65 ± 0.15 & 4.4 & 3.2 ± 0.22 \\
& \texttt{GLANCE} & 99.94 ± 0.05 & 11.24 ± 1.37 & 4.0 & 307.06 ± 0.0 & 99.98 ± 0.04 & 9.42 ± 2.22 & 8.0 & 301.47 ± 0.0 \\
\midrule

\multirow{6}{*}{LR}
& \texttt{Fast AReS} & 9.23 ± 1.24 & 1.12 ± 0.1 & 4 & 961.23 ± 541.41 & 10.95 ± 1.73 & 1.11 ± 0.07 & 8 & 901.3 ± 493.42 \\
& \texttt{CET} & 100.0 ± 0.0 & 3.57 ± 1.48 & 2.0 & 2662.77 ± 739.88 & 100.0 ± 0.0 & 2.99 ± 1.3 & 2.0 & 1628.08 ± 344.57 \\
& \texttt{GroupCF} & 90.6 ± 3.93 & 2.4 ± 1.38 & 4 & 209.26 ± 1.38 & 96.6 ± 1.96 & 1.98 ± 0.5 & 8 & 249.14 ± 0.5 \\
& \texttt{GLOBE-CE} & 99.9 ± 0.0 & 0.66 ± 0.1 & 4.0 & 0.65 ± 0.06 & 99.9 ± 0.0 & 0.55 ± 0.08 & 8.0 & 0.64 ± 0.02 \\
& \texttt{dGLOBE-CE} & 99.9 ± 0.0 & 1.63 ± 0.35 & 4.0 & 0.94 ± 0.05 & 99.9 ± 0.0 & 1.63 ± 0.35 & 8.0 & 1.35 ± 0.04 \\
& \texttt{GLANCE} & 100.0 ± 0.0 & 1.55 ± 0.54 & 4.0 & 214.6 ± 0.0 & 100.0 ± 0.0 & 1.2 ± 0.12 & 8.0 & 217.9 ± 0.0 \\
\midrule

\multirow{6}{*}{XGB}
& \texttt{Fast AReS} & 8.49 ± 1.32 & 1.16 ± 0.13 & 4 & 963.0 ± 492.98 & 11.22 ± 2.05 & 1.09 ± 0.09 & 8 & 829.62 ± 420.35 \\
& \texttt{CET} & 86.78 ± 6.7 & 12.51 ± 2.75 & 2.0 & 17339.7 ± 1363.44 & 90.25 ± 2.51 & 12.99 ± 3.49 & 6.0 & 9108.94 ± 1624.21 \\
& \texttt{GroupCF} & 78.4 ± 5.82 & 5.63 ± 1.93 & 4 & 201.24 ± 1.93 & 81.4 ± 4.22 & 5.14 ± 1.46 & 8 & 197.68 ± 1.46 \\
& \texttt{GLOBE-CE} & 28.33 ± 5.14 & 32.73 ± 0.48 & 3.2 & 1.18 ± 0.02 & 29.18 ± 4.54 & 16.41 ± 0.3 & 2.8 & 1.18 ± 0.02 \\
& \texttt{dGLOBE-CE} & 77.64 ± 11.51 & 128.0 ± 0.0 & 4.0 & 1.77 ± 0.03 & 81.95 ± 10.38 & 128.0 ± 0.0 & 8.0 & 2.55 ± 0.04 \\
& \texttt{GLANCE} & 98.94 ± 0.66 & 19.99 ± 1.91 & 4.0 & 183.45 ± 0.0 & 99.0 ± 0.55 & 16.65 ± 2.44 & 8.0 & 182.83 ± 0.0 \\
\bottomrule
\end{tabular}
}
\end{small}
\end{center}
\end{table}

Table~\ref{apptab:results_heloc} presents a comparative analysis of all algorithms for the Heloc dataset, which, due to its exclusively numeric features, poses challenges in achieving low costs. Despite this complexity, the methods  \texttt{GLANCE}, and \texttt{dGLOBE-CE} stand out by delivering near-perfect or perfect effectiveness in the DNN models, while \texttt{GLANCE}, \texttt{dGLOBE-CE}, and \texttt{GLOBE-CE} excel in the LR models. \texttt{dGLOBE-CE}-8 provides the best combination of effectiveness and cost for DNN, and \texttt{GLOBE-CE}-8 achieves the same for LR. For the XGB models, \texttt{GLANCE} leads in effectiveness across all methods. Overall, our methods demonstrate strong and consistent performance, unlike other methods, where results fluctuate considerably.

Upon evaluating the performance across the five datasets — COMPAS, German Credit, Default Credit, HELOC, and Adult Income — it is evident that our method,   \texttt{GLANCE}, consistently demonstrates concrete performance. They achieve near-perfect or perfect effectiveness scores while maintaining low costs across different models, such as DNN, LR, and XGB. This robustness and consistency underline their efficiency and practicality in handling diverse datasets and complexities. 

\section{Comparative Evaluation with Different Counterfactual Generation Methods}
\label{app:exp_candidate}

\begin{table}[ht!]
\caption{
Evaluation of the effectiveness ($\eff$) and average recourse cost ($\avcost$) of \texttt{GLANCE} with different candidate action generation methods for $s$-GCE ($s = 4$), across multiple datasets and models.}
\label{apptab:results_local}
\begin{center}
\begin{small}
\resizebox{\textwidth}{!}{
\begin{tabular}{lclcccccc}
\toprule
\multirow{2}{*}{Dataset}& \multirow{2}{*}{$\size$}  & \multirow{2}{*}{Generation Algorithm} & \multicolumn{2}{c}{DNN} & \multicolumn{2}{c}{LR} & \multicolumn{2}{c}{XGB} \\
& & & $\eff$ & $\avcost$ & $\eff$ & $\avcost$ & $\eff$ & $\avcost$ \\
\midrule
\multirow{8}{*}{Adult} 
   & \multirow{4}{*}{4} & Dice & 100.0 ± 0.0 & 4.6 ± 0.73 & 100.0 ± 0.0 & 1.04 ± 0.07 & 99.85 ± 0.12 & 5.98 ± 4.22 \\
  && NearestNeighbors & 100.0 ± 0.0 & 6.04 ± 0.35 & 100.0 ± 0.0 & 3.76 ± 0.96 & 100.0 ± 0.01 & 4.53 ± 0.56 \\
  && NearestNeighborsScaled & 99.99 ± 0.03 & 5.04 ± 0.51 & 100.0 ± 0.0 & 3.19 ± 0.51 & 99.98 ± 0.03 & 4.12 ± 0.31 \\
  && RandomSampling & 98.01 ± 0.66 & 2.03 ± 0.01 & 100.0 ± 0.0 & 0.72 ± 0.12 & 99.85 ± 0.19 & 2.1 ± 0.19 \\\cmidrule{3-9}
  &\multirow{4}{*}{8}
    & Dice & 100.0 ± 0.0 & 3.96 ± 0.56 & 100.0 ± 0.0 & 1.03 ± 0.07 & 99.87 ± 0.09 & 3.85 ± 2.33 \\
  &  & NearestNeighbors & 100.0 ± 0.01 & 4.99 ± 0.62 & 100.0 ± 0.0 & 3.22 ± 0.49 & 100.0 ± 0.0 & 4.43 ± 0.55 \\
  &  & NearestNeighborsScaled & 100.0 ± 0.0 & 4.44 ± 0.15 & 100.0 ± 0.0 & 2.96 ± 0.62 & 99.98 ± 0.03 & 3.9 ± 0.41 \\
  &  & RandomSampling & 95.36 ± 3.8 & 2.84 ± 0.8 & 100.0 ± 0.0 & 0.67 ± 0.07 & 99.76 ± 0.37 & 2.13 ± 0.17 \\
\midrule
\multirow{8}{*}{COMPAS}
  & \multirow{4}{*}{4} & Dice & 100.0 ± 0.0 & 2.34 ± 0.43 & 100.0 ± 0.0 & 2.33 ± 0.38 & 99.51 ± 0.46 & 2.96 ± 0.82 \\
 & & NearestNeighbors & 100.0 ± 0.0 & 2.27 ± 0.31 & 100.0 ± 0.0 & 2.53 ± 0.32 & 99.6 ± 0.34 & 2.95 ± 0.58 \\
 & & NearestNeighborsScaled & 100.0 ± 0.0 & 1.74 ± 0.38 & 100.0 ± 0.0 & 2.14 ± 0.15 & 99.44 ± 0.54 & 3.0 ± 0.33 \\
 & & RandomSampling & 100.0 ± 0.0 & 2.08 ± 0.21 & 100.0 ± 0.0 & 2.21 ± 0.2 & 99.88 ± 0.16 & 2.83 ± 0.76 \\\cmidrule{3-9}
 &\multirow{4}{*}{8}
    & Dice & 100.0 ± 0.0 & 1.77 ± 0.16 & 100.0 ± 0.0 & 1.69 ± 0.06 & 99.83 ± 0.24 & 2.06 ± 0.46 \\
  &   & NearestNeighbors & 100.0 ± 0.0 & 1.71 ± 0.18 & 100.0 ± 0.0 & 1.90 ± 0.12 & 99.60 ± 0.34 & 1.80 ± 0.16 \\
  &   & NearestNeighborsScaled & 100.0 ± 0.0 & 1.08 ± 0.08 & 100.0 ± 0.0 & 1.81 ± 0.18 & 99.48 ± 0.57 & 1.88 ± 0.09 \\
   &  & RandomSampling & 100.0 ± 0.0 & 1.74 ± 0.22 & 100.0 ± 0.0 & 1.78 ± 0.25 & 99.84 ± 0.24 & 2.06 ± 0.21 \\
\midrule
\multirow{8}{*}{Default Credit}
  & \multirow{4}{*}{4} & Dice & 100.0 ± 0.0 & 1.2 ± 0.4 & 100.0 ± 0.0 & 1.05 ± 0.11 & 98.13 ± 1.05 & 3.68 ± 1.64 \\
 & & NearestNeighbors & 100.0 ± 0.0 & 4.07 ± 0.08 & 100.0 ± 0.0 & 1.78 ± 0.39 & 100.0 ± 0.0 & 3.8 ± 0.44 \\
 & & NearestNeighborsScaled & 100.0 ± 0.0 & 3.02 ± 0.58 & 100.0 ± 0.0 & 1.11 ± 0.14 & 100.0 ± 0.0 & 3.09 ± 0.42 \\
 & & RandomSampling & 100.0 ± 0.0 & 1.0 ± 0.0 & 99.94 ± 0.12 & 1.0 ± 0.0 & 91.52 ± 4.44 & 2.25 ± 1.06 \\\cmidrule{3-9}
 &\multirow{4}{*}{8}
    & Dice & 100.0 ± 0.0 & 1.20 ± 0.40 & 100.0 ± 0.0 & 1.05 ± 0.11 & 98.87 ± 0.59 & 2.14 ± 0.18 \\
  &  & NearestNeighbors & 100.0 ± 0.0 & 3.84 ± 0.26 & 100.0 ± 0.0 & 1.39 ± 0.11 & 100.0 ± 0.0 & 3.20 ± 0.45 \\
   & & NearestNeighborsScaled & 100.0 ± 0.0 & 2.50 ± 0.24 & 100.0 ± 0.0 & 1.01 ± 0.00 & 100.0 ± 0.0 & 2.40 ± 0.31 \\
   & & RandomSampling & 99.82 ± 0.29 & 1.07 ± 0.14 & 99.94 ± 0.12 & 1.00 ± 0.00 & 93.86 ± 2.69 & 1.51 ± 0.52 \\
\midrule
\multirow{8}{*}{German Credit}
 & \multirow{4}{*}{4}  & Dice & 99.46 ± 1.08 & 1.22 ± 0.41 & 100.0 ± 0.0 & 1.21 ± 0.06 & 100.0 ± 0.0& 1.05 ± 0.02 \\
 & & NearestNeighbors & 100.0 ± 0.0 & 4.9 ± 0.69 & 100.0 ± 0.0 & 4.21 ± 0.56 & 100.0 ± 0.0 & 3.67 ± 0.58 \\
 & & NearestNeighborsScaled & 100.0 ± 0.0 & 2.1 ± 0.4 & 100.0 ± 0.0 & 2.02 ± 0.27 & 100.0 ± 0.0 & 1.75 ± 0.2 \\
 &  & RandomSampling & 99.43 ± 1.14 & 1.0 ± 0.2 & 100.0 ± 0.0 & 1.31 ± 0.15 & 100.0 ± 0.0 & 1.14 ± 0.07 \\\cmidrule{3-9}
 & \multirow{4}{*}{8} & Dice & 99.46 ± 1.08 & 0.78 ± 0.27 & 100.0 ± 0.0 & 1.15 ± 0.03 & 100.0 ± 0.0 & 0.97 ± 0.08 \\
  &  & NearestNeighbors & 100.0 ± 0.0 & 3.94 ± 0.39 & 100.0 ± 0.0 & 3.98 ± 0.55 & 100.0 ± 0.0 & 3.63 ± 0.58 \\
 &   & NearestNeighborsScaled & 100.0 ± 0.0 & 1.65 ± 0.14 & 100.0 ± 0.0 & 1.70 ± 0.20 & 100.0 ± 0.0 & 1.62 ± 0.13 \\
 &   & RandomSampling & 100.0 ± 0.0 & 0.70 ± 0.18 & 100.0 ± 0.0 & 1.19 ± 0.05 & 100.0 ± 0.0 & 1.06 ± 0.10 \\
\midrule
\multirow{8}{*}{HELOC}& \multirow{4}{*}{4}
  & Dice & 99.94 ± 0.05 & 11.24 ± 1.37 & 100.0 ± 0.0 & 1.55 ± 0.54 & 98.94 ± 0.66 & 19.99 ± 1.91 \\
  && NearestNeighbors & 99.82 ± 0.15 & 20.18 ± 3.28 & 99.94 ± 0.05 & 21.52 ± 1.33 & 99.51 ± 0.42 & 19.3 ± 2.66 \\
  && NearestNeighborsScaled & 99.79 ± 0.3 & 18.88 ± 2.11 & 99.94 ± 0.12 & 20.95 ± 3.47 & 99.68 ± 0.13 & 20.92 ± 3.96 \\
  && RandomSampling & 94.02 ± 4.28 & 2.53 ± 1.12 & 100.0 ± 0.0 & 1.4 ± 0.27 & 82.75 ± 5.2 & 9.6 ± 1.84\\\cmidrule{3-9}
    & \multirow{4}{*}{8}& Dice & 99.98 ± 0.04 & 9.42 ± 2.22 & 100.0 ± 0.0 & 1.20 ± 0.12 & 99.00 ± 0.55 & 16.65 ± 2.44 \\
    && NearestNeighbors & 99.75 ± 0.26 & 16.07 ± 1.32 & 99.88 ± 0.14 & 17.30 ± 1.90 & 99.49 ± 0.47 & 15.76 ± 1.71 \\
    && NearestNeighborsScaled & 99.71 ± 0.15 & 16.17 ± 1.72 & 99.88 ± 0.15 & 17.39 ± 0.69 & 99.72 ± 0.17 & 18.58 ± 1.88 \\
    && RandomSampling & 98.55 ± 0.45 & 10.01 ± 0.53 & 100.0 ± 0.0 & 1.19 ± 0.15 & 84.76 ± 4.77 & 7.11 ± 1.13 \\
\bottomrule
\end{tabular}
}
\end{small}
\end{center}
\end{table}

Table~\ref{apptab:results_local} presents a comprehensive comparison of \texttt{GLANCE} utilizing the counterfactual generation methods described in Appendix~\ref{app:candidate}, i.e.,  DiCE, NearestNeighbors, NearestNeighborsScaled, and RandomSampling, for the $s$-GCE problem with $s=4$ and $s=8$. Across all datasets and models, \texttt{GLANCE} consistently exhibits superb performance in terms of effectiveness, frequently achieving near 100\%. 
Since the performance in terms of effectiveness is comparable, we will focus on the cost our algorithms achieved under the different generation methods. 
The best performance is observed with the RandomSampling method, achieving the best results in 41 out of 60 cases. DiCE, which is regarded as state-of-the-art, follows, achieves the lowest cost in 15 out of 60 cases. NearestNeighbors and NearestNeighborsScaled achieve the best cost in only 
1 out of 60 and 3 out of 60 cases, respectively. In general, \texttt{GLANCE} can demonstrate great performance under all action generation methods, maintaining high effectiveness 
and low costs in almost all cases.

\section{Impact of Different Clustering Algorithms}
\label{app:exp_clustering}
In the initial step of GLANCE, we used k-means as the default clustering method. We also experimented with Gaussian Mixture Models (GMM) and agglomerative clustering on the COMPAS and German Credit datasets, using 100 initial clusters (n\_initial\_clusters = 100) and DiCE as the local counterfactual generation method. Across these settings, we did not find any meaningful differences in terms of effectiveness or cost.  This was expected, as the primary goal at this stage is to obtain a refined partitioning into many small clusters. In Table~ \ref{apptab:results_diff_clusters} we showcase our results for $s$ = 4 and $s$ = 8 global actions.

\begin{table}[ht!]
\caption{
Evaluation of the effectiveness ($\eff$) and average recourse cost ($\avcost$) of \texttt{GLANCE} with different clustering algorithms for $s$-GCE ($s = 4$ and $s=8$), across multiple datasets and models.
}
\label{apptab:results_diff_clusters}
\begin{center}
\begin{small}
\resizebox{\textwidth}{!}{
\begin{tabular}{lclcccccc}
\toprule
\multirow{2}{*}{Dataset} & \multirow{2}{*}{$\size$} & \multirow{2}{*}{Clustering Algorithm} & \multicolumn{2}{c}{DNN} & \multicolumn{2}{c}{LR} & \multicolumn{2}{c}{XGB} \\ \cmidrule(l){4-5} \cmidrule(l){6-7} \cmidrule(l){8-9}
& & & $\eff$ & $\avcost$ & $\eff$ & $\avcost$ & $\eff$ & $\avcost$ \\
\midrule
\multirow{6}{*}{COMPAS}
    & \multirow{3}{*}{4}
        & Kmeans        & 100.0 ± 0.0 & 2.34 ± 0.43 & 100.0 ± 0.0 & 2.33 ± 0.38 & 99.51 ± 0.46 & 2.96 ± 0.82 \\
    &                   & Agglomerative & 100.0 ± 0.0 & 2.85 ± 0.38 & 100.0 ± 0.0 & 2.27 ± 0.43 & 99.88 ± 0.16 & 3.33 ± 0.89 \\
    &                   & GMM           & 100.0 ± 0.0 & 2.39 ± 0.61 & 100.0 ± 0.0 & 2.33 ± 0.61 & 99.84 ± 0.23 & 3.5 ± 0.56 \\ \cmidrule{3-9}
    & \multirow{3}{*}{8}
        & Kmeans        & 100.0 ± 0.0 & 1.77 ± 0.16 & 100.0 ± 0.0 & 1.69 ± 0.06 & 99.83 ± 0.24 & 2.06 ± 0.46 \\
    &                   & Agglomerative & 100.0 ± 0.0 & 1.82 ± 0.15 & 100.0 ± 0.0 & 1.81 ± 0.13 & 99.96 ± 0.08 & 2.02 ± 0.18 \\
    &                   & GMM           & 100.0 ± 0.0 & 1.77 ± 0.11 & 100.0 ± 0.0 & 1.71 ± 0.07 & 99.96 ± 0.08 & 2.27 ± 0.38 \\
\midrule
\multirow{6}{*}{German Credit}
    & \multirow{3}{*}{4}
        & Kmeans        & 95.31 ± 3.15 & 1.25 ± 0.33 & 100.0 ± 0.0 & 1.21 ± 0.06 & 100.0 ± 0.0 & 1.06 ± 0.03 \\
    &                   & Agglomerative & 99.46 ± 1.08 & 1.46 ± 0.68 & 100.0 ± 0.0 & 1.45 ± 0.32 & 100.0 ± 0.0 & 1.05 ± 0.06 \\
    &                   & GMM           & 99.46 ± 1.08 & 1.23 ± 0.37 & 100.0 ± 0.0 & 1.17 ± 0.04 & 100.0 ± 0.0 & 1.16 ± 0.21 \\ \cmidrule{3-9}
    & \multirow{3}{*}{8}
        & Kmeans        & 96.43 ± 3.6 & 0.86 ± 0.18 & 100.0 ± 0.0 & 1.18 ± 0.05 & 100.0 ± 0.0 & 1.02 ± 0.04 \\
    &                   & Agglomerative & 99.46 ± 1.08 & 0.94 ± 0.29 & 100.0 ± 0.0 & 1.15 ± 0.02 & 100.0 ± 0.0 & 0.96 ± 0.11 \\ 
    &                   & GMM           & 100.0 ± 0.0 & 0.77 ± 0.28 & 100.0 ± 0.0 & 1.15 ± 0.02 & 100.0 ± 0.0 & 1.01 ± 0.03 \\
\midrule
\multirow{6}{*}{Default Credit}
    & \multirow{3}{*}{4}
        & Kmeans        & 100.0 ± 0.0 & 1.20 ± 0.40 & 100.0 ± 0.0 & 1.05 ± 0.11 & 98.13 ± 1.05 & 3.68 ± 1.64 \\
    &                   & Agglomerative & 99.97 ± 0.06 & 1.11 ± 0.45 & 100.0 ± 0.0 & 1.33 ± 0.28 & 95.83 ± 3.29 & 3.66 ± 1.53 \\
    &                   & GMM           & 100.0 ± 0.0 & 1.01 ± 0.01 & 100.0 ± 0.0 & 1.11 ± 0.21 & 97.26 ± 1.67 & 3.07 ± 1.02 \\ \cmidrule{3-9}
    & \multirow{3}{*}{8}
        & Kmeans        & 100.0 ± 0.0 & 1.2 ± 0.4 & 100.0 ± 0.0 & 1.05 ± 0.11 & 98.87 ± 0.59 & 2.14 ± 0.18 \\
    &                   & Agglomerative & 100.0 ± 0.0 & 1.05 ± 0.08 & 100.0 ± 0.0 & 1.1 ± 0.29 & 98.42 ± 1.09 & 2.03 ± 0.28 \\
    &                   & GMM           & 100.0 ± 0.0 & 1.02 ± 0.13 & 100.0 ± 0.0 & 1.1 ± 0.18 & 99.22 ± 0.78 & 2.43 ± 0.8 \\
\midrule
\multirow{6}{*}{Adult}
    & \multirow{3}{*}{4}
        & Kmeans        & 100.0 ± 0.0 & 4.6 ± 0.73 & 100.0 ± 0.0 & 1.04 ± 0.07 & 99.85 ± 0.12 & 4.9 ± 3.41 \\
    &                   & Agglomerative & 100.0 ± 0.0 & 4.73 ± 0.45 & 100.0 ± 0.0 & 1.08 ± 0.13 & 99.77 ± 0.2 & 4.15 ± 1.7 \\
    &                   & GMM           & 100.0 ± 0.0 & 4.43 ± 0.39  & 100.0 ± 0.0 & 1.1 ± 0.24 & 99.74 ± 0.24 & 5.07 ± 3.8 \\ \cmidrule{3-9}
    & \multirow{3}{*}{8}
        & Kmeans        & 100.0 ± 0.0 & 3.96 ± 0.56 & 100.0 ± 0.0 & 1.03 ± 0.07 & 99.87 ± 0.09 & 3.85 ± 2.33 \\
    &                   & Agglomerative & 100.0 ± 0.0 & 3.99 ± 0.21 & 100.0 ± 0.0 & 1.02 ± 0.22 & 99.89 ± 0.06 & 4.31 ± 2.47 \\
    &                   & GMM           & 100.0 ± 0.0 & 3.68 ± 0.52 & 100.0 ± 0.0 & 0.99 ± 0.19 & 99.78 ± 0.24 & 2.75 ± 0.93 \\
\midrule
\multirow{6}{*}{HELOC}
    & \multirow{3}{*}{4}
        & Kmeans        & 99.94 ± 0.05 & 11.24 ± 1.37 & 100.0 ± 0.0 & 1.55 ± 0.54 & 98.94 ± 0.66 & 19.99 ± 1.91 \\
    &                   & Agglomerative & 99.96 ± 0.08 & 8.16 ± 1.77 & 100.0 ± 0.0 & 1.43 ± 0.16 & 99.41 ± 0.49 & 23.61 ± 5.17 \\
    &                   & GMM           & 100.0 ± 0.0 & 8.01 ± 2.3 & 100.0 ± 0.0 & 1.45 ± 0.28 & 99.04 ± 0.51 & 22.62 ± 6.0 \\ \cmidrule{3-9}
    & \multirow{3}{*}{8}
        & Kmeans        & 99.98 ± 0.04 & 9.42 ± 2.22 & 100.0 ± 0.0 & 1.2 ± 0.12 & 99.0 ± 0.55 & 16.65 ± 2.44 \\
    &                   & Agglomerative & 99.96 ± 0.08 & 6.72 ± 0.81 & 100.0 ± 0.0 & 1.25 ± 0.18 & 99.94 ± 0.33 & 14.13 ± 1.78 \\
    &                   & GMM           & 100.0 ± 0.0 & 7.36 ± 2.85 & 100.0 ± 0.0 & 1.2 ± 0.2 & 99.32 ± 0.71 & 14.67 ± 2.04 \\
\bottomrule
\end{tabular}
}

\end{small}
\end{center}
\end{table}

\section{Comparing Local and Global Counterfactual Costs}
\label{app:exp_local}
To illustrate the challenges in defining optimal solutions for global counterfactual explanations, we conducted an experiment comparing the cost of local counterfactuals to the global counterfactual actions generated by our method. Using the Compas and German Credit datasets, we generated all possible local counterfactuals for each instance and calculated the average recourse cost across the datasets. The results are presented in Table~\ref{fig:app-local-vs-global} under the row ``Local''.

Surprisingly, although \texttt{GLANCE} generates only 4 global counterfactual actions, it achieves significantly lower average recourse costs than the local counterfactuals for the majority of model-dataset combinations, even though both were generated using the same local counterfactual method. This counterintuitive result reveals the inherent variability and complexity in local counterfactual generation, emphasizing the difficulty in defining an optimal solution at both local and global levels.

Overall, this experiment demonstrates that our global approach not only achieves cost-efficient global counterfactuals but also highlights the practical and theoretical limitations of the domain.

\begin{table}
\caption{Effectiveness ($\eff$) and average recourse cost ($\avcost$) of GCEs generated by \texttt{GLANCE} for $s=4$ compared to local counterfactuals. Despite using only four global actions, \texttt{GLANCE} consistently achieves lower costs, indicating higher efficiency and generalizability.}
\label{fig:app-local-vs-global}
\begin{center}
\begin{small}

\resizebox{\textwidth}{!}{
\begin{tabular}{llcccccc}
\toprule
\multirow{2}{*}{Dataset} & \multirow{2}{*}{Method} 
    & \multicolumn{2}{c}{DNN} 
    & \multicolumn{2}{c}{LR} 
    & \multicolumn{2}{c}{XGB} \\ \cmidrule(l){3-4}\cmidrule(l){5-6}\cmidrule(l){7-8}
& & $\eff$ & $\avcost$ & $\eff$ & $\avcost$ & $\eff$ & $\avcost$ \\
\midrule
\multirow{2}{*}{COMPAS}
    & \texttt{GLANCE} & 100.0 ± 0.0 & 2.34 ± 0.43 & 100.0 ± 0.0 & 2.33 ± 0.38 & 99.51 ± 0.46 & 2.96 ± 0.82 \\
    & Local           & 100.0 ± 0.0 & 2.75 ± 0.53 & 100.0 ± 0.0 & 2.73 ± 0.23 & 100.0 ± 0.0 & 2.87 ± 0.27 \\
\midrule
\multirow{2}{*}{German Credit}
    & \texttt{GLANCE} & 95.31 ± 3.15 & 1.25 ± 0.33 & 100.0 ± 0.0 & 1.21 ± 0.06 & 100.0 ± 0.0 & 1.06 ± 0.03 \\
    & Local           & 100.0 ± 0.0 & 4.74 ± 0.93 & 100.0 ± 0.0 & 4.90 ± 0.40 & 100.0 ± 0.0 & 5.09 ± 0.33 \\
\midrule
\multirow{2}{*}{Default Credit}
    & \texttt{GLANCE} & 100.0 ± 0.0 & 1.20 ± 0.40 & 100.0 ± 0.0 & 1.05 ± 0.11 & 98.13 ± 1.05 & 3.68 ± 1.64 \\
    & Local           & 100.0 ± 0.0 & 6.41 ± 0.24 & 100.0 ± 0.0 & 7.02 ± 0.29 & 100.0 ± 0.0 & 5.67 ± 0.19 \\
\midrule
\multirow{2}{*}{Adult}
    & \texttt{GLANCE} & 100.0 ± 0.0 & 4.6 ± 0.73 & 100.0 ± 0.0 & 1.04 ± 0.07 & 99.85 ± 0.12 & 4.9 ± 3.41 \\
    & Local           & 100.0 ± 0.0 & 6.61 ± 0.16 & 100.0 ± 0.0 & 6.34 ± 0.14 & 100.0 ± 0.0 & 6.68 ± 0.08 \\
\midrule
\multirow{2}{*}{HELOC}
    & \texttt{GLANCE} & 99.94 ± 0.05 & 11.24 ± 1.37 & 100.0 ± 0.0 & 1.55 ± 0.54 & 98.94 ± 0.66 & 19.99 ± 1.91 \\
    & Local           & 100.0 ± 0.0 & 10.45 ± 0.31 & 100.0 ± 0.0 & 8.03 ± 0.27 & 100.0 ± 0.0 & 10.38 ± 0.56 \\
\bottomrule
\end{tabular}
}

\end{small}
\end{center}
\end{table}

\section{Effect of Initial Clusters and Number of Diverse Actions on \texttt{GLANCE}}
\label{app:exp_initial}
To evaluate how key parameters influence the quality of solutions in \texttt{GLANCE}, we systematically analyzed the role of initial clusters and the number of diverse candidate actions generated per cluster. The product of these parameters defines the total number of generated actions, which directly impacts the algorithm's performance.

Increasing the number of generated actions often improves effectiveness but typically comes at the expense of increased cost. However, selecting a subset of actions with optimal effectiveness tends to reduce overall cost. This pattern is clearly observed in Tables \ref{fig:app-impact-default-eff} to \ref{fig:app-impact-heloc-cost}.

Regarding individual parameter effects, increasing the number of initial clusters proves particularly beneficial for larger datasets with widely distributed points. This approach allows for better grouping in the feature space, thereby enhancing the relevance and feasibility of the generated counterfactual actions. Conversely, increasing the diversity of candidate actions is more advantageous for complex models with intricate decision boundaries. A higher diversity ensures that the generated counterfactual actions are well-aligned with the model's structure, leading to better adaptation to the complexities of the decision boundary.

\begin{table}
\caption{
% Default Credit Effectiveness Mean
Effectiveness ($\eff$) of GCEs generated by \texttt{GLANCE} with $s=4$ on the Default Credit dataset. We vary the number of initial clusters from 10 to 100 (step size 10) and the number of diverse actions per cluster from 5 to 50 (step size 5). Effectiveness is reported as the mean over 5-fold cross-validation.
}
\label{fig:app-impact-default-eff}
\begin{center}
\resizebox{\textwidth}{!}{
\begin{tabular}{lccccccccccc}
\toprule
\multirow{2}{*}{Model} & \multirow{2}{*}{Diverse Actions} & \multicolumn{10}{c}{Number of Initial Clusters}\\ \cmidrule(l){3-12}
 && 10.0 & 20.0 & 30.0 & 40.0 & 50.0 & 60.0 & 70.0 & 80.0 & 90.0 & 100.0 \\
\midrule
\multirow{10}{*}{DNN} & 5 & 99.91 & 100.00 & 99.86 & 99.97 & 100.00 & 99.94 & 100.00 & 99.97 & 99.97 & 100.00 \\
 & 10 & 99.91 & 100.00 & 99.88 & 99.97 & 99.97 & 100.00 & 99.97 & 99.97 & 99.97 & 99.97 \\
 & 15 & 99.89 & 100.00 & 99.94 & 100.00 & 99.97 & 99.94 & 99.97 & 99.97 & 100.00 & 100.00 \\
 & 20 & 99.91 & 100.00 & 99.97 & 100.00 & 99.97 & 99.97 & 100.00 & 99.97 & 99.94 & 100.00 \\
 & 25 & 99.89 & 100.00 & 100.00 & 99.97 & 99.97 & 99.97 & 99.94 & 100.00 & 99.97 & 99.97 \\
 & 30 & 99.94 & 100.00 & 99.97 & 100.00 & 99.97 & 100.00 & 100.00 & 100.00 & 100.00 & 100.00 \\
 & 35 & 99.94 & 99.97 & 99.97 & 99.97 & 99.97 & 100.00 & 99.97 & 99.97 & 99.97 & 100.00 \\
 & 40 & 99.94 & 100.00 & 100.00 & 99.97 & 99.97 & 99.97 & 99.97 & 99.97 & 100.00 & 100.00 \\
 & 45 & 99.94 & 100.00 & 99.94 & 100.00 & 100.00 & 99.97 & 99.97 & 99.97 & 100.00 & 100.00 \\
 & 50 & 99.91 & 100.00 & 99.97 & 99.97 & 99.97 & 99.97 & 100.00 & 100.00 & 100.00 & 100.00 \\
\midrule
\multirow{10}{*}{LR}& 5 & 100.00 & 100.00 & 100.00 & 100.00 & 100.00 & 100.00 & 100.00 & 100.00 & 100.00 & 100.00 \\
 & 10 & 100.00 & 100.00 & 100.00 & 100.00 & 100.00 & 100.00 & 100.00 & 100.00 & 100.00 & 100.00 \\
 & 15 & 100.00 & 100.00 & 100.00 & 100.00 & 100.00 & 100.00 & 100.00 & 100.00 & 100.00 & 100.00 \\
 & 20 & 100.00 & 100.00 & 100.00 & 100.00 & 100.00 & 100.00 & 100.00 & 100.00 & 100.00 & 100.00 \\
 & 25 & 100.00 & 100.00 & 100.00 & 100.00 & 100.00 & 100.00 & 100.00 & 100.00 & 100.00 & 100.00 \\
 & 30 & 100.00 & 100.00 & 100.00 & 100.00 & 100.00 & 100.00 & 100.00 & 100.00 & 100.00 & 100.00 \\
 & 35 & 100.00 & 100.00 & 100.00 & 100.00 & 100.00 & 100.00 & 100.00 & 100.00 & 100.00 & 100.00 \\
 & 40 & 100.00 & 100.00 & 100.00 & 100.00 & 100.00 & 100.00 & 100.00 & 100.00 & 100.00 & 100.00 \\
 & 45 & 100.00 & 100.00 & 100.00 & 100.00 & 100.00 & 100.00 & 100.00 & 100.00 & 100.00 & 100.00 \\
 & 50 & 100.00 & 100.00 & 100.00 & 100.00 & 100.00 & 100.00 & 100.00 & 100.00 & 100.00 & 100.00 \\
\midrule
\multirow{10}{*}{XGB} & 5 & 94.42 & 94.39 & 95.48 & 96.50 & 95.53 & 94.95 & 96.29 & 96.14 & 96.42 & 94.59 \\
 & 10 & 94.03 & 96.08 & 97.38 & 95.99 & 95.85 & 98.24 & 96.79 & 96.94 & 95.54 & 96.68 \\
 & 15 & 95.95 & 95.97 & 97.73 & 96.30 & 97.39 & 98.01 & 97.97 & 97.94 & 97.92 & 96.49 \\
 & 20 & 96.02 & 96.62 & 97.79 & 96.19 & 98.54 & 98.54 & 98.62 & 98.52 & 98.11 & 98.34 \\
 & 25 & 96.66 & 95.72 & 98.57 & 96.84 & 98.62 & 98.65 & 98.57 & 97.76 & 98.21 & 98.36 \\
 & 30 & 96.68 & 96.69 & 98.62 & 96.31 & 98.49 & 98.67 & 98.46 & 98.49 & 98.43 & 98.97 \\
 & 35 & 96.73 & 97.17 & 99.00 & 97.56 & 98.67 & 98.85 & 98.46 & 98.84 & 98.58 & 98.70 \\
 & 40 & 96.94 & 97.69 & 98.82 & 97.44 & 98.90 & 98.63 & 98.92 & 99.07 & 98.80 & 98.99 \\
 & 45 & 97.76 & 98.51 & 98.82 & 97.69 & 98.82 & 99.00 & 98.44 & 98.89 & 98.82 & 99.07 \\
 & 50 & 97.20 & 98.49 & 98.82 & 98.63 & 98.74 & 99.11 & 98.74 & 98.94 & 98.87 & 99.01 \\
\bottomrule
\end{tabular}
}
\end{center}
\end{table}

\begin{table}
\caption{
Average Recourse Cost ($\avcost$) of GCEs generated by \texttt{GLANCE} with $s=4$ on the Default Credit dataset. We vary the number of initial clusters from 10 to 100 (step size 10) and the number of diverse actions per cluster from 5 to 50 (step size 5). Average Recourse Cost is reported as the mean over 5-fold cross-validation.
}
\label{fig:app-impact-default-cost}
\begin{center}
\resizebox{\textwidth}{!}{
\begin{tabular}{lccccccccccc}
\toprule
\multirow{2}{*}{Model} & \multirow{2}{*}{Diverse Actions} & \multicolumn{10}{c}{Number of Initial Clusters}\\ \cmidrule(l){3-12}
 && 10 & 20 & 30 & 40 & 50 & 60 & 70 & 80 & 90 & 100 \\
\midrule
\multirow{10}{*}{DNN} & 5 & 4.85 & 3.15 & 1.45 & 1.58 & 1.40 & 2.27 & 1.13 & 2.65 & 1.41 & 1.32 \\
 & 10 & 2.91 & 2.53 & 1.98 & 1.41 & 1.59 & 1.39 & 1.33 & 1.51 & 1.03 & 1.28 \\
 & 15 & 4.66 & 1.90 & 1.19 & 1.25 & 1.21 & 1.01 & 1.03 & 1.11 & 1.02 & 1.03 \\
 & 20 & 2.76 & 1.71 & 1.27 & 1.39 & 1.02 & 1.00 & 1.01 & 1.08 & 1.02 & 1.01 \\
 & 25 & 3.91 & 1.96 & 1.27 & 1.40 & 1.11 & 1.01 & 1.29 & 1.25 & 1.02 & 1.00 \\
 & 30 & 2.16 & 1.90 & 1.27 & 1.40 & 1.01 & 1.01 & 1.21 & 1.28 & 1.01 & 1.01 \\
 & 35 & 2.16 & 1.62 & 1.27 & 1.39 & 1.00 & 1.01 & 1.29 & 1.07 & 1.00 & 1.00 \\
 & 40 & 1.82 & 1.47 & 1.21 & 1.23 & 1.00 & 1.01 & 1.02 & 1.19 & 1.01 & 1.01 \\
 & 45 & 1.78 & 1.54 & 1.21 & 1.07 & 1.01 & 1.00 & 1.02 & 1.07 & 1.00 & 1.01 \\
 & 50 & 1.82 & 1.34 & 1.02 & 1.03 & 1.01 & 1.00 & 1.22 & 1.08 & 1.01 & 1.01 \\
\midrule
\multirow{10}{*}{LR}& 5 & 1.93 & 1.43 & 1.33 & 1.47 & 1.45 & 1.48 & 1.30 & 1.27 & 1.32 & 1.51 \\
 & 10 & 1.91 & 1.13 & 1.18 & 1.34 & 1.43 & 1.29 & 1.26 & 1.09 & 1.02 & 1.38 \\
 & 15 & 1.59 & 1.13 & 1.17 & 1.27 & 1.23 & 1.34 & 1.22 & 1.09 & 1.23 & 1.17 \\
 & 20 & 1.22 & 1.01 & 1.27 & 1.35 & 1.22 & 1.24 & 1.20 & 1.11 & 0.99 & 1.08 \\
 & 25 & 1.20 & 1.13 & 1.19 & 1.10 & 1.11 & 1.32 & 1.19 & 1.05 & 1.17 & 1.10 \\
 & 30 & 1.20 & 1.00 & 1.27 & 1.13 & 1.22 & 1.20 & 1.09 & 1.19 & 1.21 & 1.18 \\
 & 35 & 1.11 & 1.00 & 1.16 & 1.12 & 1.15 & 1.17 & 1.05 & 0.99 & 1.07 & 1.23 \\
 & 40 & 1.11 & 1.00 & 1.20 & 1.16 & 1.10 & 1.11 & 1.11 & 1.00 & 0.99 & 0.99 \\
 & 45 & 1.11 & 1.00 & 1.11 & 1.07 & 1.08 & 1.06 & 1.18 & 0.97 & 1.21 & 0.98 \\
 & 50 & 1.11 & 1.00 & 1.05 & 1.07 & 1.07 & 1.15 & 1.06 & 1.09 & 1.07 & 0.96 \\
\midrule
\multirow{10}{*}{XGB} & 5 & 3.44 & 3.43 & 2.83 & 2.76 & 3.30 & 2.99 & 2.67 & 2.67 & 2.65 & 3.39 \\
 & 10 & 2.97 & 2.81 & 2.91 & 3.88 & 5.08 & 4.05 & 2.97 & 2.89 & 2.99 & 3.47 \\
 & 15 & 3.13 & 2.47 & 2.83 & 2.81 & 3.92 & 3.89 & 3.56 & 3.30 & 3.26 & 4.06 \\
 & 20 & 2.98 & 3.35 & 3.81 & 3.10 & 4.08 & 4.38 & 3.69 & 3.37 & 3.17 & 4.04 \\
 & 25 & 2.95 & 3.15 & 3.81 & 3.03 & 2.63 & 3.29 & 4.26 & 3.75 & 3.75 & 3.84 \\
 & 30 & 3.42 & 3.82 & 3.97 & 3.16 & 4.59 & 2.97 & 4.48 & 3.23 & 3.57 & 4.45 \\
 & 35 & 2.65 & 3.01 & 4.18 & 3.06 & 5.12 & 4.24 & 3.27 & 3.43 & 4.10 & 3.75 \\
 & 40 & 2.43 & 3.09 & 3.39 & 3.98 & 3.54 & 4.57 & 4.12 & 3.52 & 3.74 & 4.59 \\
 & 45 & 2.57 & 3.83 & 3.39 & 4.03 & 2.84 & 4.84 & 5.66 & 4.11 & 3.47 & 5.14 \\
 & 50 & 2.50 & 3.08 & 2.89 & 3.79 & 3.58 & 3.94 & 4.99 & 4.39 & 4.45 & 3.77 \\
\bottomrule
\end{tabular}
}
\end{center}
\end{table}

\begin{table}
\caption{
% HELOC Effectiveness Mean
Effectiveness ($\eff$) of GCEs generated by \texttt{GLANCE} with $s=4$ on the HELOC dataset. We vary the number of initial clusters from 10 to 100 (step size 10) and the number of diverse actions per cluster from 5 to 50 (step size 5). Effectiveness is reported as the mean over 5-fold cross-validation.
}
\label{fig:app-impact-heloc-eff}
\begin{center}
\resizebox{\textwidth}{!}{
\begin{tabular}{lccccccccccc}
\toprule
\multirow{2}{*}{Model} & \multirow{2}{*}{Diverse Actions} & \multicolumn{10}{c}{Number of Initial Clusters}\\ \cmidrule(l){3-12}
 && 10 & 20 & 30 & 40 & 50 & 60 & 70 & 80 & 90 & 100 \\
\midrule
\multirow{10}{*}{DNN} & 5 & 99.21 & 99.67 & 99.61 & 99.34 & 99.53 & 99.46 & 99.88 & 99.69 & 99.88 & 99.61 \\
 & 10 & 99.41 & 99.66 & 99.22 & 99.28 & 99.43 & 99.92 & 99.92 & 99.86 & 99.92 & 99.92 \\
 & 15 & 99.36 & 99.08 & 99.21 & 99.27 & 99.39 & 99.94 & 99.92 & 99.73 & 99.92 & 99.94 \\
 & 20 & 99.72 & 99.08 & 99.14 & 99.67 & 99.92 & 99.89 & 99.98 & 99.92 & 99.97 & 99.98 \\
 & 25 & 99.74 & 99.08 & 99.14 & 99.39 & 99.98 & 99.92 & 99.98 & 99.96 & 99.98 & 99.98 \\
 & 30 & 99.78 & 99.08 & 99.23 & 99.98 & 99.98 & 100.00 & 99.95 & 99.98 & 99.96 & 99.98 \\
 & 35 & 99.78 & 99.10 & 99.27 & 99.98 & 100.00 & 100.00 & 99.98 & 99.98 & 99.96 & 99.98 \\
 & 40 & 99.76 & 98.89 & 99.60 & 100.00 & 100.00 & 99.98 & 99.94 & 99.98 & 99.98 & 100.00 \\
 & 45 & 99.76 & 99.12 & 99.65 & 100.00 & 99.98 & 99.98 & 100.00 & 99.98 & 99.98 & 99.96 \\
 & 50 & 99.76 & 99.12 & 99.65 & 100.00 & 99.98 & 100.00 & 99.94 & 99.98 & 99.98 & 99.96 \\
\midrule
\multirow{10}{*}{LR}& 5 & 100.00 & 100.00 & 100.00 & 100.00 & 100.00 & 100.00 & 100.00 & 100.00 & 100.00 & 100.00 \\
 & 10 & 100.00 & 100.00 & 100.00 & 100.00 & 100.00 & 100.00 & 100.00 & 100.00 & 100.00 & 100.00 \\
 & 15 & 100.00 & 100.00 & 100.00 & 100.00 & 100.00 & 100.00 & 100.00 & 100.00 & 100.00 & 100.00 \\
 & 20 & 100.00 & 100.00 & 100.00 & 100.00 & 100.00 & 100.00 & 100.00 & 100.00 & 100.00 & 100.00 \\
 & 25 & 100.00 & 100.00 & 100.00 & 100.00 & 100.00 & 100.00 & 100.00 & 100.00 & 100.00 & 100.00 \\
 & 30 & 100.00 & 100.00 & 100.00 & 100.00 & 100.00 & 100.00 & 100.00 & 100.00 & 100.00 & 100.00 \\
 & 35 & 100.00 & 100.00 & 100.00 & 100.00 & 100.00 & 100.00 & 100.00 & 100.00 & 100.00 & 100.00 \\
 & 40 & 100.00 & 100.00 & 100.00 & 100.00 & 100.00 & 100.00 & 100.00 & 100.00 & 100.00 & 100.00 \\
 & 45 & 100.00 & 100.00 & 100.00 & 100.00 & 100.00 & 100.00 & 100.00 & 100.00 & 100.00 & 100.00 \\
 & 50 & 100.00 & 100.00 & 100.00 & 100.00 & 100.00 & 100.00 & 100.00 & 100.00 & 100.00 & 100.00 \\
\midrule
\multirow{10}{*}{XGB} & 5 & 87.66 & 91.56 & 94.14 & 94.27 & 95.33 & 96.28 & 97.12 & 97.34 & 98.02 & 98.19 \\
 & 10 & 90.45 & 95.79 & 96.01 & 97.95 & 97.86 & 98.57 & 98.39 & 98.73 & 98.64 & 99.38 \\
 & 15 & 93.13 & 96.53 & 96.93 & 97.94 & 98.05 & 98.40 & 98.69 & 98.82 & 98.92 & 99.32 \\
 & 20 & 94.71 & 97.87 & 96.89 & 98.26 & 98.34 & 98.49 & 99.13 & 99.00 & 98.94 & 99.42 \\
 & 25 & 95.23 & 97.80 & 97.55 & 98.71 & 98.98 & 98.94 & 99.01 & 99.07 & 98.99 & 99.58 \\
 & 30 & 96.14 & 97.76 & 98.65 & 98.84 & 98.62 & 99.00 & 99.07 & 98.98 & 99.38 & 99.38 \\
 & 35 & 95.96 & 98.17 & 98.92 & 99.01 & 99.02 & 99.15 & 99.07 & 98.83 & 99.39 & 99.53 \\
 & 40 & 96.51 & 98.63 & 99.25 & 99.34 & 99.40 & 99.20 & 99.58 & 99.17 & 99.57 & 99.72 \\
 & 45 & 97.43 & 98.97 & 99.27 & 99.37 & 99.39 & 99.13 & 99.54 & 99.54 & 99.58 & 99.85 \\
 & 50 & 97.95 & 98.89 & 99.30 & 99.24 & 99.49 & 99.26 & 99.66 & 99.43 & 99.62 & 99.91 \\
\bottomrule
\end{tabular}
}
\end{center}
\end{table}

\begin{table}
\caption{
% HELOC Average Cost Mean
Average Recourse Cost ($\avcost$) of GCEs generated by \texttt{GLANCE} with $s=4$ on the HELOC dataset. We vary the number of initial clusters from 10 to 100 (step size 10) and the number of diverse actions per cluster from 5 to 50 (step size 5). Average Recourse Cost is reported as the mean over 5-fold cross-validation.
}
\label{fig:app-impact-heloc-cost}
\begin{center}
\resizebox{\textwidth}{!}{
\begin{tabular}{lccccccccccc}
\toprule
 \multirow{2}{*}{Model} & \multirow{2}{*}{Diverse Actions} & \multicolumn{10}{c}{Number of Initial Clusters}\\ \cmidrule(l){3-12}
 && 10 & 20 & 30 & 40 & 50 & 60 & 70 & 80 & 90 & 100 \\
\midrule
\multirow{10}{*}{DNN} & 5 & 9.03 & 7.44 & 10.79 & 9.14 & 11.58 & 10.49 & 10.20 & 11.21 & 9.67 & 10.55 \\
 & 10 & 10.97 & 10.12 & 10.58 & 9.85 & 9.97 & 10.77 & 10.95 & 11.35 & 10.37 & 11.25 \\
 & 15 & 10.54 & 11.09 & 10.96 & 10.35 & 10.36 & 10.23 & 10.57 & 11.17 & 9.83 & 9.43 \\
 & 20 & 10.85 & 11.27 & 11.35 & 10.38 & 11.87 & 12.11 & 10.98 & 11.87 & 10.19 & 10.50 \\
 & 25 & 10.80 & 11.30 & 11.86 & 10.13 & 10.76 & 11.54 & 11.51 & 11.23 & 10.09 & 11.75 \\
 & 30 & 11.70 & 11.08 & 10.87 & 12.77 & 10.92 & 11.51 & 10.96 & 10.80 & 10.34 & 9.54 \\
 & 35 & 11.70 & 11.06 & 10.86 & 11.70 & 11.97 & 12.22 & 10.54 & 11.08 & 10.11 & 10.82 \\
 & 40 & 12.03 & 11.06 & 11.01 & 11.29 & 11.56 & 12.22 & 10.87 & 11.48 & 10.12 & 10.81 \\
 & 45 & 12.03 & 11.25 & 11.63 & 12.47 & 11.59 & 12.65 & 11.90 & 9.85 & 10.21 & 8.84 \\
 & 50 & 12.03 & 11.20 & 11.62 & 12.17 & 11.59 & 11.64 & 11.67 & 10.70 & 11.39 & 11.52 \\
\midrule
\multirow{10}{*}{LR}& 5 & 2.67 & 2.40 & 1.92 & 1.77 & 1.78 & 1.92 & 2.00 & 1.72 & 1.66 & 1.68 \\
 & 10 & 1.91 & 2.04 & 1.88 & 1.63 & 1.79 & 1.66 & 1.66 & 1.50 & 1.57 & 1.52 \\
 & 15 & 2.09 & 1.73 & 1.65 & 1.77 & 1.48 & 1.41 & 1.56 & 1.49 & 1.46 & 1.35 \\
 & 20 & 1.80 & 1.56 & 1.59 & 1.60 & 1.54 & 1.52 & 1.59 & 1.53 & 1.47 & 1.25 \\
 & 25 & 1.80 & 1.65 & 1.62 & 1.51 & 1.51 & 1.52 & 1.50 & 1.48 & 1.43 & 1.46 \\
 & 30 & 1.69 & 1.62 & 1.54 & 1.51 & 1.57 & 1.38 & 1.52 & 1.52 & 1.43 & 1.32 \\
 & 35 & 1.69 & 1.62 & 1.57 & 1.41 & 1.49 & 1.24 & 1.43 & 1.40 & 1.49 & 1.34 \\
 & 40 & 1.66 & 1.53 & 1.41 & 1.38 & 1.52 & 1.28 & 1.34 & 1.39 & 1.40 & 1.35 \\
 & 45 & 1.64 & 1.53 & 1.50 & 1.40 & 1.46 & 1.30 & 1.43 & 1.39 & 1.37 & 1.31 \\
 & 50 & 1.64 & 1.47 & 1.44 & 1.31 & 1.46 & 1.27 & 1.37 & 1.43 & 1.23 & 1.35 \\
\midrule
\multirow{10}{*}{XGB} & 5 & 12.22 & 14.05 & 15.44 & 14.77 & 15.91 & 16.25 & 17.14 & 17.13 & 17.87 & 18.28 \\
 & 10 & 13.83 & 16.14 & 16.51 & 18.75 & 17.54 & 17.57 & 20.25 & 20.62 & 21.32 & 25.72 \\
 & 15 & 15.33 & 18.93 & 17.05 & 17.40 & 18.07 & 21.27 & 19.51 & 21.48 & 27.93 & 18.61 \\
 & 20 & 14.92 & 18.59 & 16.79 & 20.42 & 21.98 & 23.13 & 22.44 & 20.12 & 22.97 & 23.07 \\
 & 25 & 16.03 & 20.48 & 17.85 & 19.41 & 22.98 & 22.43 & 21.69 & 25.36 & 27.18 & 22.68 \\
 & 30 & 15.74 & 20.30 & 18.35 & 21.33 & 23.88 & 24.85 & 23.86 & 23.11 & 24.39 & 24.77 \\
 & 35 & 17.16 & 22.75 & 20.80 & 21.62 & 23.70 & 27.08 & 24.47 & 20.27 & 25.74 & 24.13 \\
 & 40 & 16.58 & 25.14 & 21.67 & 23.77 & 24.82 & 29.50 & 26.66 & 23.40 & 26.58 & 29.42 \\
 & 45 & 18.55 & 25.83 & 22.86 & 25.32 & 25.87 & 26.71 & 25.22 & 25.62 & 22.86 & 27.09 \\
 & 50 & 18.75 & 23.21 & 26.56 & 24.85 & 28.62 & 27.21 & 24.82 & 28.62 & 24.27 & 26.10 \\
\bottomrule
\end{tabular}
}
\end{center}
\end{table}

\section{Addressing Potential Edge Cases in Recourse Algorithms}
\label{app:edge}
While the proposed algorithms are designed to provide robust and actionable recourse, certain edge cases may pose challenges:

Highly nonlinear decision boundaries might cause the model's decision boundary to vary significantly across small regions, leading to oversimplified recourse suggestions. We address this issue by using proximity-aware clustering and local counterfactual action generation to better approximate the decision boundary, ensuring more accurate and actionable recommendations.

Sparse data regions or outlier points might result in unrealistic, infeasible, or overly tailored recourse actions. To mitigate these challenges, we generate many clusters and diverse counterfactuals, ensuring that feasible solutions are identified without being disproportionately influenced by outliers. This approach enhances robustness and broad applicability.

Ambiguity in selecting the ``optimal'' action can arise when multiple actions have similar costs and effectiveness, potentially leading to overly complex recourse suggestions. We resolve this by employing predefined criteria that prioritize actionable and interpretable solutions while maintaining a balance between cost and effectiveness.

\section{User Study}
\label{app:user}
Following the best practices outlined in \cite{pmlr-v202-ley23a,chowdhury2022equi,warren2023explaining}, we conducted an online user study, with the following goals.
\begin{enumerate}
    \item Assess how participants weigh trade-offs between effectiveness, average recourse cost, and size.
    \item Validate our evaluation metrics, i.e., solution practicality and robustness (investigate how variance in recourse cost and effectiveness affects the participants' decisions). 
    \item Assess how participants rank \texttt{GLANCE} relative to baselines in non-dominated solution scenarios.
\end{enumerate}

We recruited 55 participants from six countries, comprising  (a) PhD students and (b) researchers from various machine learning domains. 

\subsection{Part 1: Algorithm Ranking Task}

In the first part of the study, participants were asked to rank Global Counterfactual Explanation (GCEs) produced by three methods: (a) \texttt{GLANCE} with $s=3$, (b) \texttt{GLOBE-CE} with $s=3$, and (c) \texttt{GLOBE-CE} under its default configuration (resulting in action set sizes ranging from 58 to 526). Each question presented visualizations similar to Figure \ref{fig:actions_example}a, showing the effectiveness, average recourse cost, size for each GCEs together with the actions that consist the CGEs.
Algorithm names were anonymized and randomly permuted to prevent identification bias. Participants were asked to select one of six possible total orderings (i.e., complete rankings) over the three alternatives presented per example.

\subsubsection{Design}

Each participant completed three ranking tasks based on different datasets and models. For each task, the three algorithms displayed were evaluated on the same test set and visualized identically. The key objective was to assess how participants perceive trade-offs among effectiveness, average recourse cost, and size (number of actions), and how these factors influence their preferences. The default configuration of \texttt{GLOBE-CE} outperformed its constrained variant with $s=3$ in terms of both average recourse cost and effectiveness. However, it did not always outperform \texttt{GLANCE} across both effectiveness and cost, highlighting possible trade-offs between the size, effectiveness, and average recourse cost. 

To evaluate global preferences across participants, we aggregated individual rankings using the Borda count method \cite{fishburn1976borda}. Each ranking position was assigned a numerical score (3 for first, 2 for second, 1 for third), and the scores were summed across participants. 

\subsubsection{Key Findings}

\begin{itemize}
    \item The resulting aggregated ranking was:
    $$
    \texttt{GLANCE}\mid s=3 \succ\texttt{GLOBE-CE} \succ \texttt{GLOBE-CE}\mid s=3,
    $$
    with total Borda scores of 473, 277, and 241, respectively, and where $a \succ b$ means algorithm $a$ is preferred over $b$.
    
    \item Although \texttt{GLANCE} received the highest Borda score, participants expressed mixed preferences between the two versions of \texttt{GLOBE-CE}. The default configuration of \texttt{GLOBE-CE} achieved better effectiveness and cost performance than its constrained version ($s=3$), but this came at the expense of much larger action sets (up to 526 actions), which negatively influenced participant preference. 
    
    Several participants were willing to trade off higher cost and lower effectiveness in favor of smaller and more interpretable action sets, indicating that the explanation size significantly influenced perceived utility, even when effectiveness and cost were not comparable.

    \item Qualitative feedback indicated that the number of actions played a critical role in participants' decisions, even when the cost and effectiveness of larger solutions were more favorable. Many participants reported prioritizing interpretability, often trading off marginal gains in cost or effectiveness.
\end{itemize}

\subsubsection{Statistical Analysis}
To assess the significance of preference differences across the evaluated methods, we conducted a Friedman test followed by post-hoc pairwise comparisons. The Friedman test revealed a highly significant difference across the methods ($x^2 = 190.05, p < 10^{-40}$), indicating that at least one method outperforms the others. 

We then applied the Nemenyi post-hoc test to compare the methods pairwise. The results show that \texttt{GLANCE} with $s=3$ significantly outperforms both \texttt{GLOBE-CE} under default parameters  ($p <0.0001$)
and \texttt{GLOBE-CE} with $s=3$  ($p <0.0001$). However, the performance difference between the two \texttt{GLOBE-CE} variants was not statistically significant under the Nemenyi test
($p = 0.1035$).

These results demonstrate that \texttt{GLANCE} is significantly preferred across the benchmark tasks, with consistent and statistically validated preference over the competitors.

\subsection{Part 2: Pairwise Algorithm Comparison and Trade-off Reasoning}
In the second part, participants were shown visualizations resembling the subplots in Figure~\ref{fig:cost-eff-plots-4acts}, which depict the performance metrics presented in Table~\ref{tab:results_4} for dataset/model combinations. Each figure depicted two anonymized algorithms with randomly permuted labels, colors, and markers to avoid bias. Participants were asked to select which algorithm they preferred and provide their main reasoning. 

To support interpretation and ensure consistency in how participants viewed trade-offs, we provided a list of predefined justifications for each question. Some justifications follow, but are not limited to :
\begin{itemize}
    \item Algorithm 1 because it has higher effectiveness
    \item Algorithm 1 because it has higher effectiveness and comparable cost
    \item Algorithm 2 because it has a smaller cost
\item Algorithm 2 because it has a smaller cost and comparable effectiveness
\item Algorithm 2 because it has a smaller variance in cost. 

\end{itemize}

The structured reasons allowed us to assess how participants evaluate effectiveness, average recourse cost, and variance in either when comparing solutions. 

Multiple comparisons were constructed to evaluate:
\begin{enumerate}
    \item How participants respond to trade-offs between effectiveness and average recourse cost, assuming low variance.
    \item Whether participants prioritize robustness (i.e., low variance).
    \item Whether dominance is respected or overridden by subjective considerations.
\end{enumerate}

\subsubsection{Design}

Each participant answered five pairwise comparison questions:
\begin{enumerate}
    \item One involved an impractical baseline solution (e.g., DNN/Adult : \texttt{Fast Ares} vs \texttt{GLANCE}).
    \item One involved a case where \texttt{GLANCE} was formally dominated, demonstrating lower effectiveness and higher cost  (DNN/HELOC: \texttt{dGLOBE-CE} vs. \texttt{GLANCE}).
    \item Three involved non-dominated algorithm pairs, including \texttt{GLANCE} and various baselines (e.g., DNN/German Credit: \texttt{CET} vs \texttt{GLANCE}).
\end{enumerate}
Participants chose one preferred algorithm per question and selected a justification from the list. Free-text feedback was also collected to better understand reasoning.

\subsubsection{Key Findings}

\begin{enumerate}
    \item In the comparison with an impractical solution, 100\% of the participants preferred \texttt{GLANCE}. Of these, 77.8\% stated that their decision stemmed from the low effectiveness of the baseline, not just \texttt{GLANCE}'s strength. The remaining 22.2\% stated that they simply prioritized effectiveness over cost. These results validate the practicality criterion used in our experimental evaluation, as the majority of the participants deemed these solutions impractical and therefore rejected them, despite favorable cost metrics. 
    \item In the dominated comparison (DNN/HELOC), where \texttt{dGLOBE-CE} outperformed \texttt{GLANCE} in both cost and effectiveness, 74.5\% of participants still preferred \texttt{GLANCE}, citing its lower variance. This suggests that robustness considerations can override formal dominance in human evaluation. The 74.5\% preference for \texttt{GLANCE} (vs. 25.5\% for baseline method) was statistically significant (binomial test, $p < 0.01$).
    \item In the remaining three non-dominated comparisons, an average of 71.5\% of participants preferred \texttt{GLANCE} over baseline methods. Across all non-dominated comparisons, 14.5\% of participants on average selected the “smaller variance” option as their main justification for their decisions, and 27.8\% of participants explicitly stated in free text feedback that robustness was critical in their evaluations.
\end{enumerate}
 \subsubsection{Statistical Analysis} 
To statistically validate the superiority of \texttt{GLANCE} in pairwise comparisons, we conduct one-sided binomial tests for the three cases we previously described.
\begin{enumerate}
    \item \texttt{GLANCE} vs.\ impractical solutions.  Out of 55 test cases, \texttt{GLANCE} was judged to outperform the baseline in all cases. Under the null hypothesis of no difference (i.e., a 50\% chance of either method being preferred), the probability of observing at least 55 wins is $p = 2.776 \times 10^{-17}$. This result confirms that GLANCE is significantly preferred over the competing method across the benchmark ($p<0.01$).

    \item \texttt{GLANCE} dominated by \texttt{dGLOBE-CE} for DNN/HELOC. Out of 55 test cases, \texttt{GLANCE} was judged to outperform the baseline in 41 cases. Under the null hypothesis of no difference (i.e., a 50\% chance of either method being preferred), the probability of observing at least 41 wins is $p=0.0001776$. This result confirms that GLANCE is significantly preferred over the competing method across the benchmark ($p<0.01$).

    \item  \texttt{GLANCE} vs.\ competitors in non-dominated cases. Out of 165 test cases, \texttt{GLANCE} was judged to outperform the baseline in all cases. Under the null hypothesis of no difference (i.e., a 50\% chance of either method being preferred), the probability of observing at least 118 wins is $p = 1.564 \times 10^{-8}$. This result confirms that GLANCE is significantly preferred over the competing method across the benchmark ($p<0.01$).
\end{enumerate}

\subsubsection{Participants' Preferences and Trade-offs}
Even though 58.2\% of the participants reported prioritizing effectiveness overall, they provided justifications that indicated that trade-offs were carefully considered. Participants were more consistent in evaluating what counts as comparable effectiveness than in judging comparable cost. Multiple participants commented on the challenge of interpreting cost without domain knowledge and difficulties in deciding if a cost difference was indeed significant. This highlights the need for domain-specific cost metrics for recourse evaluation, which can only be done by domain experts, and justifies our decision on \texttt{GLANCE} framework flexibility on using various distance metrics.

\subsection{User Study Limitations and Future Work}
This study was conducted with explainability-aware participants, including researchers and engineers familiar with machine learning terminology. Future work could replicate this user study, designed for non-expert participants, to evaluate whether preferences generalize across populations. The current results could then serve as a control baseline.

\end{document}